\documentclass  [
paper    = a4,
BCOR     = 10mm,
twoside,
fontsize = 12pt,
fleqn,
toc      = bibnumbered,
toc      = listofnumbered,
numbers  = noendperiod,
headings = normal,
listof   = leveldown,
version  = 3.03
]                                       {scrreprt}

\usepackage     [utf8]                  {inputenc}
\usepackage     [T1]                    {fontenc}
\usepackage                             {color}
\usepackage                             {amsmath}
\usepackage                             {graphicx}
\usepackage     [english]               {babel}
\usepackage                             {natbib}
\usepackage                             {hyperref}
\usepackage                             {subfig}
\usepackage                             {float}
\usepackage     [normalem]              {ulem}
\usepackage                             {booktabs}
\usepackage                             {tabularx}
\usepackage                             {nccmath}
\usepackage                             {amssymb}
\usepackage                             {algorithm}
\usepackage                             {rotating}
\usepackage                             {algpseudocode}
\useunder                               {\uline}{\ul}{}

\definecolor{darkblue}{rgb}{0.0,0.0,0.4}
\definecolor{darkgreen}{rgb}{0.0,0.4,0.0}
\hypersetup{
colorlinks,
linkcolor=black,
citecolor=darkgreen,
urlcolor=darkblue
}

\begin{document}

\thispagestyle{empty}
\begin{center}
  \renewcommand{\baselinestretch}{2.00}
  \Large\sffamily
  Department of Mathematics and Computer Science\\
  \large University of Heidelberg
  \par\vfill\normalfont
  Master thesis\\
  in Scientific Computing\\
  submitted by\\
  Shourya Verma\\
  born in Kanpur, India\\
  2023
\end{center}
\newpage

\thispagestyle{empty}
\begin{center}
  \renewcommand{\baselinestretch}{2.00}
  \Large\bfseries\sffamily
    Self-Trained Panoptic Segmentation
  \par
  \vfill
  \large\normalfont
  This master thesis has been carried out by Shourya Verma\\
  at the\\
  European Molecular Biology Laboratory, Heidelberg\\
  under the supervision of\\
  Dr Anna Kreshuk and Prof Fred Hamprecht
\end{center}\par
\vspace{5\baselineskip}

\renewcommand{\baselinestretch}{1.00}\normalsize 
\thispagestyle{empty}
\normalfont
\small
\begin{center}
  \begin{minipage}[c][0.48\textheight][b]{0.9\textwidth}   
   
I extend my heartfelt gratitude to my supervisors, Doctor Anna Kreshuk and Professor Fred Hamprecht, for their guidance throughout my Master's journey. Prof Hamprecht's gracious offer to engage in a student research project within his esteemed lab at Heidelberg University, and Dr Kreshuk's kind opportunity to pursue my thesis within her distinguished group at the European Molecular Biology Laboratory enriched my academic experience immeasurably. Their encouragement allowed me to delve into exciting cross-disciplinary research projects.

Furthermore, I wish to express my appreciation to Doctor Adrian Wolny, for his mentorship during our joint project. His insightful contributions greatly inspired me, and the wealth of knowledge I learned from him has been instrumental in shaping my understanding.
   
  \end{minipage}
\end{center}
\thispagestyle{empty}
\normalfont
\small
\begin{center}
  \begin{minipage}[c][0.48\textheight][b]{0.9\textwidth}
    \textbf{
      Self-Trained Panoptic Segmentation:
    }\par
    \vspace{\baselineskip}
    Panoptic segmentation is an important computer vision task which combines semantic and instance segmentation. It plays a crucial role in domains of medical image analysis, self-driving vehicles, and robotics by providing a comprehensive understanding of visual environments. Traditionally, deep learning panoptic segmentation models have relied on dense and accurately annotated training data, which is expensive and time consuming to obtain. Recent advancements in self-supervised learning approaches have shown great potential in leveraging synthetic and unlabelled data to generate pseudo-labels using self-training to improve the performance of instance and semantic segmentation models. The three available methods for self-supervised panoptic segmentation use proposal-based transformer architectures which are computationally expensive, complicated and engineered for specific tasks. The aim of this work is to develop a framework to perform embedding-based self-supervised panoptic segmentation using self-training in a synthetic-to-real domain adaptation problem setting.
  \end{minipage}\par
  \vfill
  \begin{minipage}[c][0.48\textheight][b]{0.9\textwidth}
    \small
    \textbf{
      Selbst-Trainierte Panoptische Segmentierung:
    }\par
    \vspace{\baselineskip}
    Die panoptische Segmentierung ist eine wichtige Computer-Vision-Aufgabe, die semantische und Instanzsegmentierung kombiniert. Es spielt eine entscheidende Rolle in den Bereichen medizinische Bildanalyse, selbstfahrende Fahrzeuge und Robotik, indem es ein umfassendes Verständnis visueller Umgebungen ermöglicht. Traditionell basieren panoptische Deep-Learning-Segmentierungsmodelle auf dichten und genau kommentierten Trainingsdaten, deren Beschaffung teuer und zeitaufwändig ist. Jüngste Fortschritte bei selbstüberwachten Lernansätzen haben gezeigt, dass die Nutzung synthetischer und unbeschrifteter Daten zur Generierung von Pseudobezeichnungen durch Selbsttraining ein großes Potenzial bietet, um die Leistung von Instanz- und semantischen Segmentierungsmodellen zu verbessern. Die drei verfügbaren Methoden für die selbstüberwachte panoptische Segmentierung verwenden vorschlagsbasierte Transformatorarchitekturen, die rechenintensiv, kompliziert und für bestimmte Aufgaben konzipiert sind. Das Ziel dieser Arbeit ist die Entwicklung eines Frameworks zur Durchführung einer einbettungsbasierten, selbstüberwachten panoptischen Segmentierung unter Verwendung von Selbsttraining in einer Problemumgebung zur Anpassung von synthetischen an reale Domänen.
  \end{minipage}
\end{center}

\tableofcontents

\part{Introduction}

\chapter{Motivation}

Panoptic segmentation is a computer vision task which combines semantic and instance segmentation techniques. It segments images into pixel wise regions for all object instances while also providing labels for all semantic categories. The semantic segmentation task performs pixel-wise classification that assigns a label or category to each pixel in an image, which partitions the image into meaningful regions based on the objects present. Instance segmentation extends this by distinguishing individual instances of objects within the same class by assigning a unique identifier to each distinct object instance. Panoptic segmentation combines these two tasks by providing a complete and unified understanding of an image, including both semantic categories and object instances in a single coherent output. It's applications are in fields where human-level detection of environment is needed, like surveillance and security, traffic control systems, autonomous driving, and medical image analysis etc. In the context of self-driving vehicles task, panoptic segmentation can enhance perception capabilities of autonomous systems by enabling them to understand the surrounding environment in greater detail allowing for more informed decision-making during driving. 

Deep learning based approaches used to tackle panoptic segmentation problems often require large amounts of annotated data and substantial computational resources for training and inference. Compiling and labelling these datasets involves extensive human effort and the use of specialized software and tools which is costly and time consuming. Recent advancements in self-supervised learning have allowed models to be trained without requiring dense manual annotation of data. Self-supervised learning aims to learn meaningful representations from synthetic and unlabelled data through self-training that iteratively generates pseudo-labels from the input data and use them to train a segmentation model. This self-training method not only reduces data annotation overhead but also facilitates domain adaptation by transferring knowledge from a source domain with labelled data to a target domain without labelled data. 

Despite the immense potential, the field of self-supervised panoptic segmentation has been fairly unexplored with less than half a dozen papers available to be cited. All previous approaches employ region proposal based transformer object detection models which generate weak bounding box proposals with associated category labels using task specific pre-defined algorithms. In contrast, this project explores embedding based methods to produce pixel-level embeddings allowing the network to efficiently capture the semantic context of each pixel while also preserving instance-level details. This thesis implements self-trained instance and semantic segmentation models and merge them to compute panoptic segmentation masks. It also investigates if the segmentation performance can be further improved by guided iterative refinement process using a multi-branch approach which leverages complementary information between semantic and instance masks. This refinement loop allows initial predicted semantic masks to guide the instance network producing more confident instance masks which in-turn improve incomplete semantic masks.

\chapter{Main Contribution}

This thesis focuses on development of self-trained models for panoptic segmentation task and addresses key problems like: unavailability of densely annotated data, domain shifts in data and modality, generalization to unseen labels, low scalability and transferability, and computationally inefficient training methods. The proposed method extends two existing techniques for semantic and instance segmentation and merges them for panoptic segmentation task using a multi-branch architecture which allows flexibility in model selection and category-centric training for different tasks. Predictions made from both semantic and instance models improve each other's training process allowing them to outperform the vanilla implementations. This method can be used to solve other panoptic segmentation tasks for different natural image datasets in biology, medicine.

\part{Background}

\chapter{Image Segmentation}
\begin{figure}[h]
  \centering
  \subfloat[City Scene]{\includegraphics[width=0.495\textwidth]{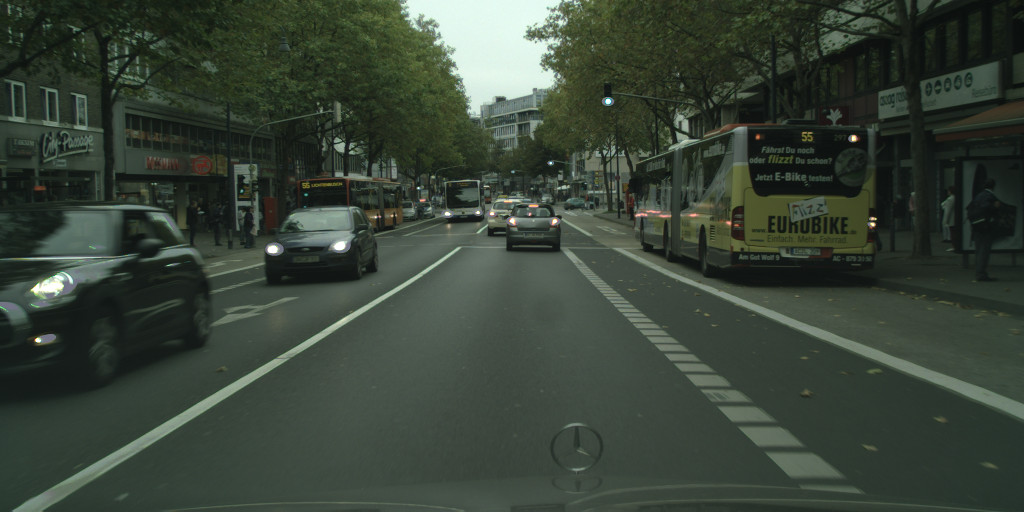}\label{fig:scene}}
  \hfill
  \subfloat[Segmentation]{\includegraphics[width=0.495\textwidth]{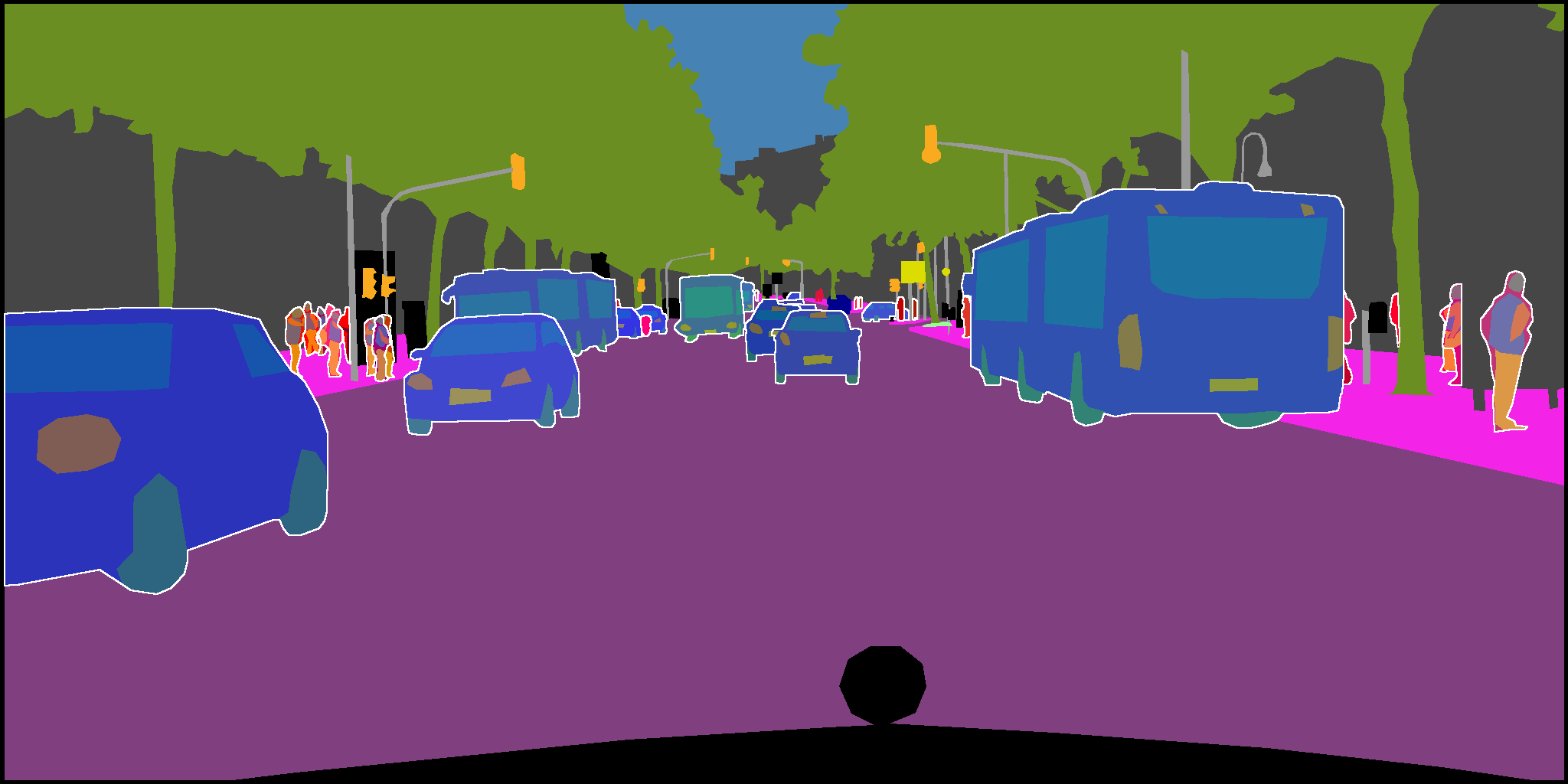}\label{fig:segmentation}}
  \caption{City scene and segmentation}
  \label{fig:image1}
\end{figure}

Image segmentation partitions different areas of an image into segments where each pixel is mapped to an object \cite{minaee2021image}. The portions of the image which are isolated from the rest due to a specific property are called segmentation masks. Since an image is a collection of pixels, the segmentation masks are groups of pixels with similar attributes and hence can be used to measure different properties of the subject of segmentation. Figure \ref{fig:image1} shows image of a scene and its segmentation from a city. Image segmentation can be of three primary categories namely semantic, instance and panoptic segmentation.

\begin{figure}[h]
\centering
\subfloat[City Scene]{\includegraphics[trim={5 5 5 5},clip, width=0.499\textwidth]{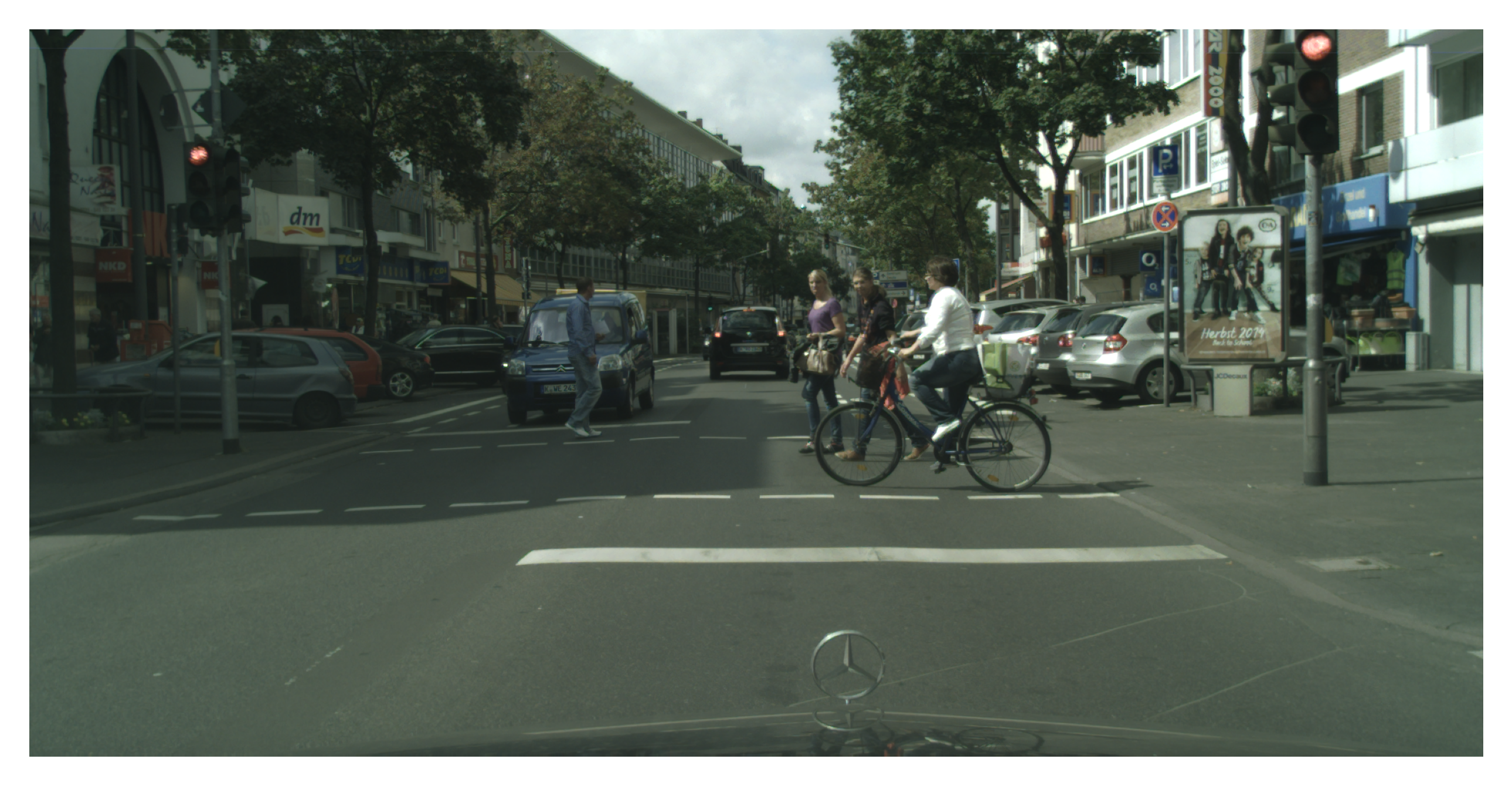}\label{fig:scene0}}
\hfill
\subfloat[Semantic Segmentation (stuff)]{\includegraphics[trim={5 5 5 5},clip, width=0.499\textwidth]{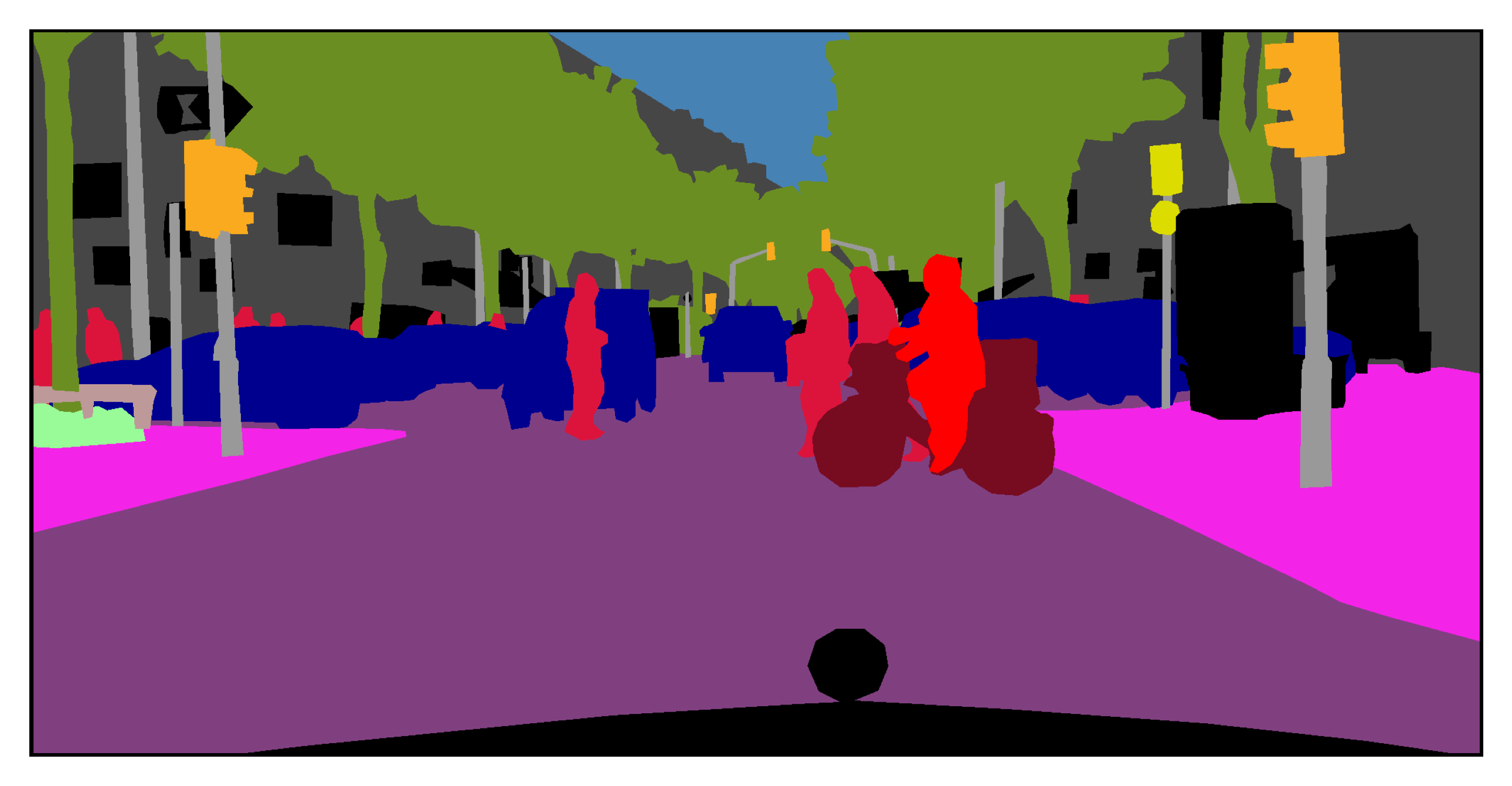}\label{fig:scene1}}
\hfill
\subfloat[Instance Segmentation (things)]{\includegraphics[trim={5 5 5 5},clip, width=0.499\textwidth]{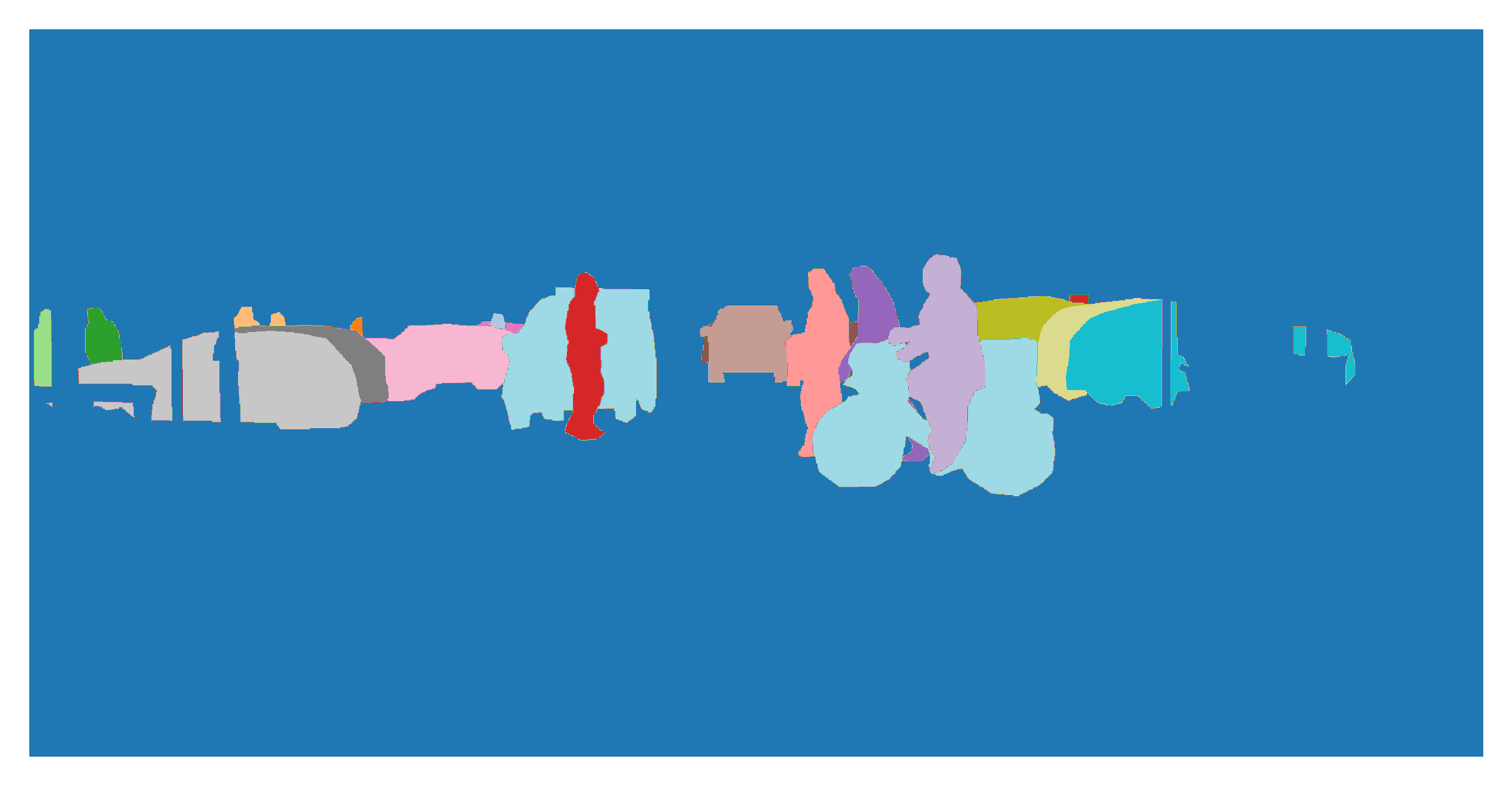}\label{fig:scene2}}
\hfill
\subfloat[Panoptic Segmentation]{\includegraphics[trim={5 5 5 5},clip, width=0.499\textwidth]{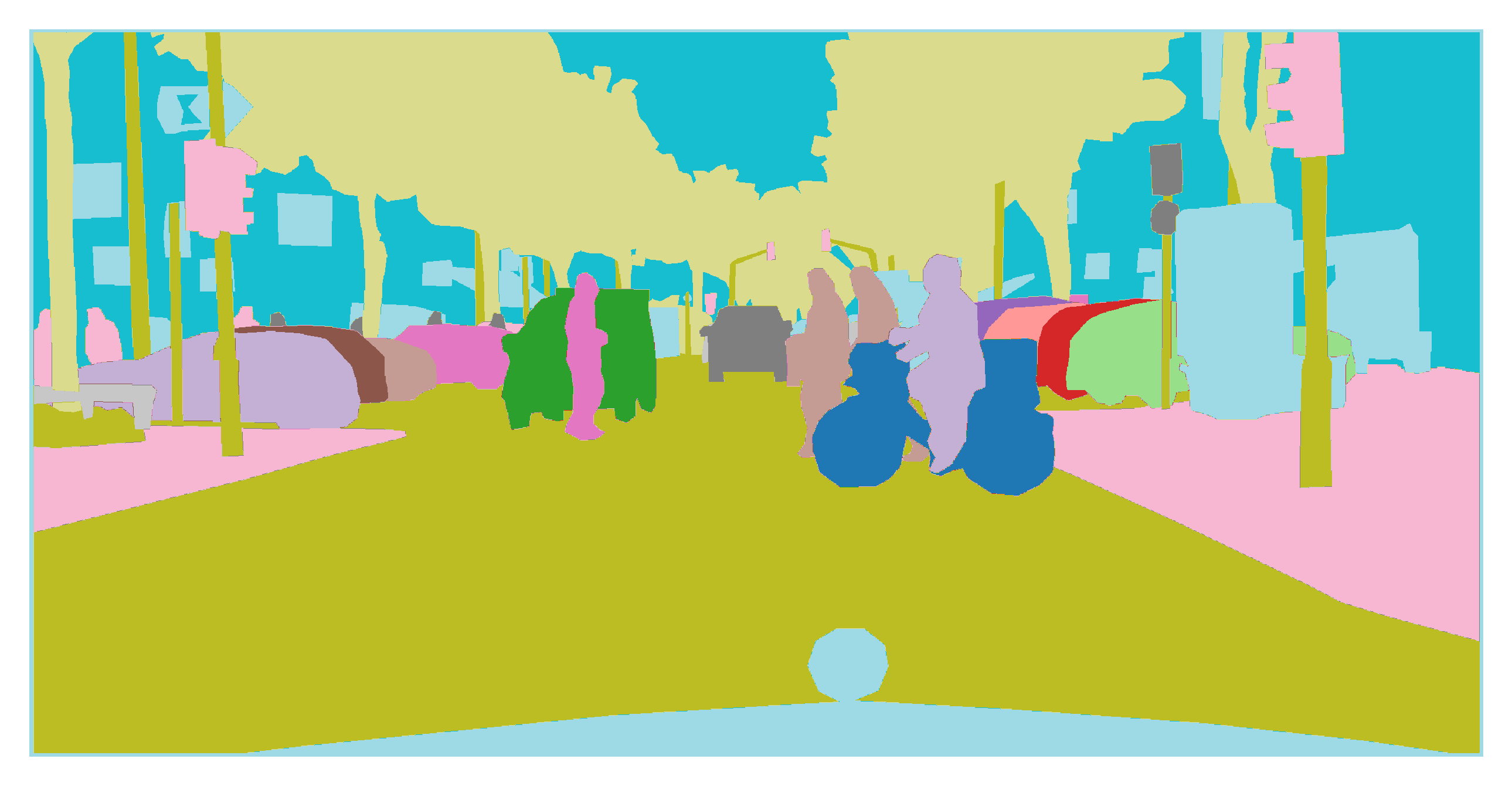}\label{fig:scene3}}
\caption{Image of city scene with semantic, instance and panoptic segmentations}
\label{fig:image2}
\end{figure}

Semantic segmentation is where each pixel is labelled with corresponding class, indicating the region to which that pixel belongs to \cite{guo2018review}. The meaningful labelling assignment allows for the understanding the semantic context of an image. Figure \ref{fig:scene1} shows a scene where each pixel is labelled according to its semantic category as \{\textit{cars, trees, road, persons etc.}\}. Semantic segmentation does not distinguish between various instances of the same class, i.e. all instances of the same class will have the same label. The goal is to provide an understanding of the scene by identifying different object categories and their spatial arrangement. Deep learning architectures like U-Net and DeepLab which use convolutional neural networks are popular choices for this task.

Instance segmentation is where the image is not only segmented into different object classes but also distinguishing individual instances of those objects within the same class \cite{hafiz2020survey}. Hence it provides pixel-level labels for each object instance present in the image. Each pixel is labelled corresponding to the specific instance of the object it belongs to which allows for delineation between multiple instances of the same class. Figure \ref{fig:scene2} shows a scene where each pixel is labelled as individual instance of \{\textit{car, person, bicycle, etc.}\}. It is useful in scenes where multiple instances of the same class overlap or are closely located like a crowded street with multiple cars. It is a more complicated problem compared to semantic segmentation since it requires object classification and localization between every instance of multiple classes. Deep learning architecture like Mask Region Based Convolutional Neural Networks \cite{he2017mask} is a popular choice for this task.

Panoptic segmentation combines semantic and instance segmentation providing a more complete and extensive understanding of an image \cite{kirillov2019panoptic}. Every pixel in an image is assigned both a semantic category label and an instance label which produces masks with information about what objects are present and where each instance of those objects is located. Figure \ref{fig:scene3} shows a scene where each pixel is labelled for all object instances called "things" while also providing labels for all semantic categories called "stuff". Deep learning architectures like Panoptic-DeepLab and OneFormer are popular methods that solve this task.

Panoptic segmentation has applications in various fields like biology, medicine and self-driving \cite{li2022survey}. In bio-science it can be used for cell segmentation and analysis in microscopic images and helps researches study cell morphology by identifying cellular structures and interactions. It can also aid in medical image analysis \cite{zhang2018panoptic} by segmenting different tissues and organs in magnetic resonance imaging and computerised tomography scan images, and assist in diagnosing diseases, tracking treatment progress and planning surgical interventions. In autonomous driving panoptic segmentation can provide a detailed understand of the surrounding environment in the field of view by identifying different elements in a scene like sidewalks, roads, vehicles, humans etc. which helps in informed driving decisions. In summary, panoptic segmentation provides detailed pixel-level parsing of images, simultaneously capturing semantics and instance-level distinctions. This comprehensive understanding of visual environments has widespread benefits for scientific research and safety-critical applications.

\chapter{Deep Learning}

\section{Neural Networks For Image Segmentation}
Deep learning focuses on training artificial neural networks and enables automatic extraction of intricate features from given data \cite{lecun2015deep}. It uses an input layer that takes raw data followed by hidden layers for feature extraction. Neurons within a layer are connected to ones in subsequent layer which enable propagation of information. Various activation functions introduce non-linearity into the model, important for learning intricate patterns. Training deep neural network models involves minimizing a predefined loss function that quantifies dissimilarity between predicted outputs and ground-truth values.

\begin{figure}[h]
\centering
\includegraphics[width=0.55\textwidth]{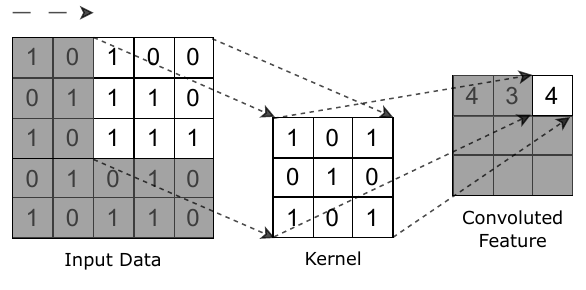}
\caption{Convolution on input image using kernel}
\label{fig:cnn}
\end{figure}

\textbf{Convolutional Neural Network} (CNN) \cite{lecun1995convolutional} is a type of artificial neural network designed specifically for processing grid-like data, such as images and videos. It's widely used in various applications, including image recognition, image segmentation, and image generation. CNN employs "convolutional layers" that automatically learn and extract features from input data shown in figure \ref{fig:cnn}. These layers apply filters or kernels to input images, detecting patterns like edges, textures, and shapes. The learned features become progressively more abstract as they move deeper into the network. This hierarchical representation allows the network to capture complex patterns and relationships within the data. Training a CNN involves adjusting the parameters of the network by using an optimizer to update weights based on gradients calculated by backpropagation.

DeepLab \cite{chen2017deeplab} is a family of architectures designed for semantic image segmentation. One key feature of DeepLab is its use of atrous (dilated) convolutions, which allow the network to increase the receptive field of each neuron without significantly increasing the computational cost. This enables capturing context over large image regions while maintaining efficiency. One variant of DeepLab, known as DeepLabv3+, employs an encoder-decoder architecture with atrous convolutions in the encoder to capture context and maintain fine details. It also uses atrous spatial pyramid pooling to capture multi-scale information, which is explained in detail in the methodology section. DeepLab architectures have demonstrated impressive results in various segmentation challenges, such as segmenting objects in natural and medical images. Further details about DeepLab are explained in section \ref{semantic_model}

U-Net \cite{ronneberger2015u} is a CNN architecture introduced for biomedical image segmentation tasks. It's particularly well-suited for cases where there is a limited amount of labelled data. The architecture resembles a "U" shape, with an encoder on one side and a decoder on other side. The encoder captures features from the input image, reducing the spatial dimensions. The decoder then upsamples and fuses these features to generate a segmentation map that matches the original image's size. Skip connections, which connect corresponding layers from the encoder to the decoder, help retain fine details during upsampling. U-Net's architecture enables it to capture both local and global context, making it effective for tasks like cell and organ segmentation in medical images. Further details about U-Net are explained in section \ref{instance_model}

Transformer architecture models like DETR \cite{carion2020end} and MaskFormer \cite{cheng2021per} use object detection with region proposal based on queries to create bounding boxes around object instances to perform segmentation. Region proposal based segmentation involves first generating candidate segments in an image using object detection or segmentation algorithms. They generate a set of candidate regions for input images which are likely to contain objects using techniques such as sub-segmentation, greedy search, grouping etc. These proposals represent regions of interest that could contain objects and are individually classified using a classifier. This approach of combining object detection with classification involves implementing complex pipelines which use pre-defined bounding box algorithms for specific tasks and require multiple steps. These steps introduce significant complexity which can limit practical use, however it can provide accurate results by focusing on specific object-like regions in the image. 

Embedding-based segmentation, on the other hand, uses end-to-end fully convolutional network to directly embed each pixel or region of an image into a high-dimensional vector space. These embeddings are then clustered to group pixels of similar characteristics into the same semantic category. This approach eliminates the need for explicit region proposal generation and works directly with pixel-level information \cite{8803021} \cite{wolny2022sparse}. Embedding-based segmentation can be more computationally efficient and easier to implement than proposal-based approaches since it directly operates at the pixel level and often requires fewer separate steps. However, it might struggle with accurately distinguishing small or closely grouped objects.

\section{Self-Training and Pseudo-label Generation} \label{ssl}
Supervised learning is a conventional approach where a deep learning model is trained using labelled data which consists of input examples paired with corresponding target outputs \cite{caruana2006empirical}. The objective is to teach the model to map inputs to the correct outputs by learning the underlying patterns in the data. The labels serve as ground-truth that guide the model's learning process and requires a significant amount of labelled data for training.

Self-supervised learning is an alternative paradigm where the models are trained using inherent structures or context within the data itself, without requiring explicit labelled data. This is particularly valuable when labelled data is sparse or expensive to obtain \cite{goodfellow2016deep}, \cite{hastie2009elements}. Domain adaptation involves adapting a model's learned features and representations from the source dataset to the target dataset, which may have different distributions or characteristics. This is particularly important when there is a lack of labelled data in the target domain. Instead of starting the training of a deep learning model from scratch using the limited target domain data, domain adaptation techniques aim to transfer the knowledge gained from the source domain to improve the model's performance on the target domain. Pre-training a model on a source domain and self-training it on a target domain helps adapt the model's learned features to the target domain while retaining the knowledge from the source domain. Common techniques involve using the input image to create multiple augmented views, and training the model to predict consistency between the different views. Other approaches involve using cues like image rotation prediction, clustering, or generation/prediction tasks on image patches as pretext tasks.

\begin{figure}[h]
  \centering
  \subfloat[Knowledge Distillation]{\includegraphics[width=0.625\textwidth]{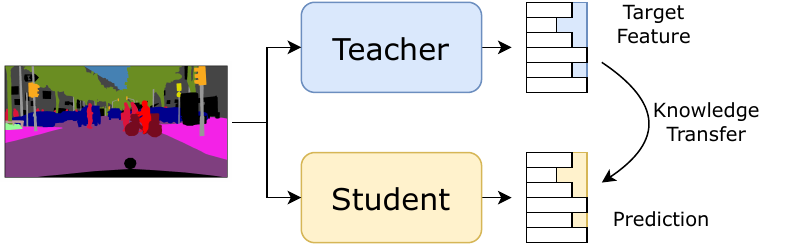}\label{fig:KD}}
  \hfill
  \subfloat[Co-evolving Pseudo-labels]{\includegraphics[width=0.375\textwidth]{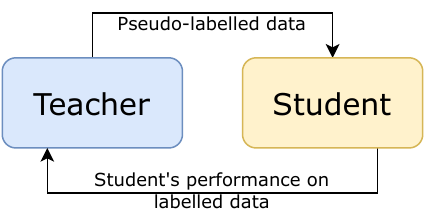}}\label{fig:PL}
  \caption{Overview of self-training framework}
  \label{fig:self-train}
\end{figure}

Self-training using pseudo-label generation and knowledge distillation is a combination of two techniques aimed at improving the performance of deep learning models for tasks like segmentation. Firstly, a baseline model is trained using set of labelled data in the source domain. Then during self-training, the model predicts soft-labels for unlabelled data in the target domain, which generates preliminary pseudo-labels based on the highest confidence scores in its predictions. These pseudo-labels are incorporated into the target domain training dataset for iterative model refinement. This process works together to produce co-evolving pseudo-labels using a mean teacher framework introduced by \cite{tarvainen2017mean}. The mean teacher framework introduces two models: the primary "student" model and a secondary "teacher" model. Knowledge distillation is employed to transfer knowledge from a teacher model to a student model which are parameterized identically \cite{ericsson2022self}, \cite{chen2017learning}. The approach is to use a teacher model's confident predictions as pseudo-labels to train a student model. A large pre-trained teacher model generates soft probabilistic labels (probability distribution over all possible classes), and the student model is trained to reproduce these soft targets, learning a compact representation. The main element is the introduction of a consistency loss which quantifies the divergence between student and teacher model predictions on both labelled and pseudo-labelled data. The teacher model is updated by maintaining an exponential moving average of the student weights, as shown by \cite{french2017self}. During training, the objective is to minimize a combined loss, consisting of the standard cross-entropy segmentation loss on labelled data and the consistency loss on both labelled and pseudo-labelled data. This framework ensures that pseudo-labels align with the teacher model's predictions, minimizing error propagation and increasing training stability. Figure \ref{fig:self-train} shows the framework of knowledge distillation and co-evolving pseudo-labels self-train methods.

\section{Discriminative Contrastive Learning}

Contrastive learning is a self-supervised learning approach where the model learns to differentiate between positive (similar) and negative (dissimilar) pairs of data samples \cite{jaiswal2020survey} \cite{liu2021self}. Contrastive learning maximizes agreement between various augmented views of input data batches through a contrastive loss, where the idea is that semantically similar samples should have similar latent representations. The loss pulls positive pairs close while pushing negative pairs apart in the embedding space. Data augmentations like cropping, flipping etc. create different views and the contrastive prediction task allows model to learn useful features without needing semantically meaningful labels. Large batches ensure sufficient negative samples and the resulting embedded representations excel at generalization tasks. Through training, meaningful representations are learned as it encourages the model to reduce the distance between similar samples in the feature space while simultaneously increasing the separation between dissimilar ones.

\begin{figure}[h]
\centering
\subfloat[Instance-wise pixels]{\includegraphics[trim={50 40 70 30},clip, width=0.49\textwidth]{images/scene2.png}\label{fig:scene22}}
\hspace{10pt}
\subfloat[Pixel embeddings in 2D feature space]{\includegraphics[trim={0 0 0 0},clip, scale=0.7]{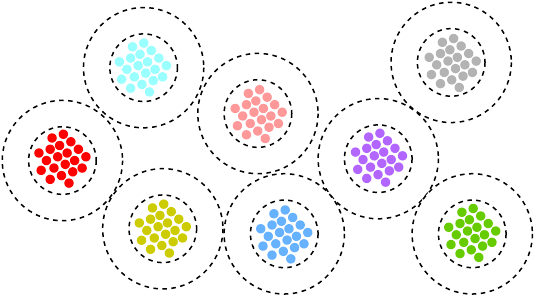}\label{fig:scene33}}
\caption{Overview of discriminative contrastive learning}
\label{fig:contrast}
\end{figure}

\cite{wolny2022sparse} and \cite{de2017semantic} implement a discriminative loss with "push" and "pull" forces. The push force encourages embeddings of similar pixels within the same class to be closer in the feature space, enhancing intra-class compactness, and the pull force enforces greater separation between embeddings of different classes, promoting inter-class discrimination. Further implementation details about the pull and push discriminative contrastive loss are explained in section \ref{instance_model}. These forces are embedded into a unified loss function, optimizing both compactness and distinctiveness of embeddings $\mathcal{L} =L_{\text {pull }}+ L_{\text {push }}$. The resulting framework offers improved segmentation performance by encouraging more effective feature representations. \cite{wolny2022sparse} uses clustering created by introducing this loss to extract soft masks for given instance. This is done by selecting anchor points randomly on an image, then computing distance map in the embedding space starting from the anchor point and extending to all image pixels. This process involves applying a Gaussian kernel function to softly determine pixels within the neighbourhood to produce pseudo-labels. Figure \ref{fig:contrast} illustrates the main idea behind discriminative contrastive learning where pixels belonging to the same class are near each other and can be easily clustered, since the segmentation network maps each pixel to a point in feature space. 

\chapter{Related Work}

\section{Semantic and Instance Segmentation}

Domain adaptive semantic segmentation often relies on techniques like fine-tuning pre-trained models on target domain data or adversarial training to align feature distributions across domains \cite{kang2020_plca} \cite{Luo_2019_ICCV} \cite{hoyer2022daformer} \cite{lai2022decouplenet}. These methods, while effective in some cases, struggled with addressing domain shifts adequately. Domain adaptive instance segmentation uses methods like few-shot instance segmentation, feature-level adaptation and semi-supervised techniques, combining labelled source data with limited target data \cite{hsu2021darcnn} \cite{liu2020pdam} \cite{chen2020semi} \cite{bellver2020mask}. These methods struggle with generalizing to these unseen object shapes and require coarse annotation of all the objects in a few images. This project utilizes and builds on methods for domain adaptive self-supervised semantic segmentation by \cite{araslanov2021self} and weakly-supervised instance segmentation by \cite{wolny2022sparse}, to develop framework for self-trained panoptic segmentation.

\cite{araslanov2021self} proposed a method in domain adaptation for semantic segmentation which efficient and accurate by developing a lightweight self-supervised framework. Their methodology streamlines the training process by adopting a one-round training approach, employing co-evolving pseudo labels. This approach combines aspects of the noisy mean teacher \cite{xie2020self} \cite{tarvainen2017mean} concept with consistency regularization and self-ensembling \cite{izmailov2018averaging}. They opt for photometric invariance, scale equivariance, and flip equivariance \cite{wang2020self}, as opposed to more resource-intensive sampling methods, to derive accurate pseudo-supervision. They deviate from a previous study by \cite{subhani2020learning}, by finding that scale alone isn't a reliable indicator of label quality. Therefore, they aggregate predictions from various scalings and flips, similar to uncertainty estimation during inference augmentation, but applied during training.

Their framework encompasses a segmentation network designed for adapting to a target domain, coupled with a slowly evolving copy updated through a momentum network. In the context of self-supervised scene adaptation, both networks receive random crop batches and horizontal flips from a target domain image. By averaging predictions from the momentum network for each pixel and considering inverse spatial transformations, they establish a pseudo-ground-truth. This is determined by picking confident pixels based on dynamically adaptable thresholds derived from dynamic statistics. Following this, the segmentation network adjusts its parameters through stochastic gradient descent referencing pseudo-labels, and their momentum network provides reliable targets for self-supervised training of the segmentation network.

\cite{wolny2022sparse} introduced an instance segmentation method that relies on non-spatial embeddings, which leverages the structure of the learned embedding space to extract individual instances in a differentiable manner. Their work tackles dense annotation by introducing a weak supervision for instance segmentation, by providing mask annotations for a small batch of instances in images, leaving other pixels unlabelled, a concept known as "positive unlabelled" setting \cite{liu2003building} \cite{lejeune2021positive}. Instance segmentation is well-suited for positive unlabelled supervision. Sampling a few objects in each image rather than densely annotating a few images introduces the network to a more diverse training set. This enhances model generalization which is beneficial for datasets with varied sub-domains in the raw data distribution, enabling comprehensive sampling without added annotation time.

In the context of weak positive unlabelled supervision, their approach suggests adding an instance-level consistency loss \cite{he2020momentum} \cite{tarvainen2017mean} to stabilize training from sparse object masks in the unlabelled areas of images. Unlike other weakly supervised segmentation methods, their simple unlabelled consistency loss doesn't require estimating class priors or propagating pseudo-labels. Additionally, their approach facilitates efficient domain adaptation using a small number of object masks in the target domain as supervision, offering an alternative to starting from scratch.

\section{Self-Supervised Panoptic Segmentation}

The field of domain adaptive self-supervised panoptic segmentation is extremely new and remains fairly unexplored. Based on available information and current knowledge, only three studies tackle this problem, two of which are from the same set of authors. These methods are from researchers at Nanyang Technological University and Wenzhou University: \textbf{Cross-View Regularization for Domain Adaptive Panoptic Segmentation} (CVRN) by \cite{huang2021cross}; and \textbf{Unified Domain Adaptive Panoptic Segmentation Transformer via Hierarchical Mask Calibration} (UniDAformer) by \cite{zhang2023unidaformer}. The third method \textbf{Enhanced Domain-Adaptive Panoptic Segmentation} (EDAPS) created by \cite{saha2023edaps} at ETH Zurich is currently the state-of-the-art method for domain adaptive self-supervised panoptic segmentation. Instead of treating semantic segmentation and instance segmentation as independent training tasks, CVRN \cite{huang2021cross} introduces an inter-task regularizer that encourages these two tasks to collaborate and mutually reinforce each other. This regularizer formulates semantic consistency across various styles, using variations in illumination, weather conditions, contrast, and more. It treats different styles as distinct views of the same image, facilitating domain adaptation within individual images. The two inter-task and inter-style regularization components involve predicting pseudo-labels for target-domain samples, using self-training techniques commonly employed in various domain adaptive computer vision tasks. UniDAformer \cite{zhang2023unidaformer} undertakes a dynamic self-training process in real-time and addresses the challenge of inaccurate predictions by introducing a technique called hierarchical mask calibration. This transformer based method focuses on refining incorrect predictions at different levels, including regions, superpixels, and pixels. This approach treats both object-based instances and contextual semantic predictions as masks, uniformly improving each predicted pseudo mask in a coarse-to-refined manner. It has three distinctive attributes: comprehensive panoptic adaptation by uniformly treating both objects and context as masks, mitigating inaccurate predictions through an iterative and progressive calibration process, and end-to-end training characterized by fewer parameters, and a simpler training and inference pipeline. UniDAformer provides no open-source code-base for their experiments is available so the results cannot be replicated. EDAPS \cite{saha2023edaps} takes a step further and utilizes feature stacking and context-aware fusion for semantic decoder along with region proposals and feature alignment for the instance decoder. They share a transformer backbone and encoders for both tasks since it allows the model to  jointly adapt semantic and instance features from the source to the target domain. They also inherit rare-class sampling feature from UniDAformer.

\begin{figure}[h]
\centering
\includegraphics[width=0.60\textwidth]{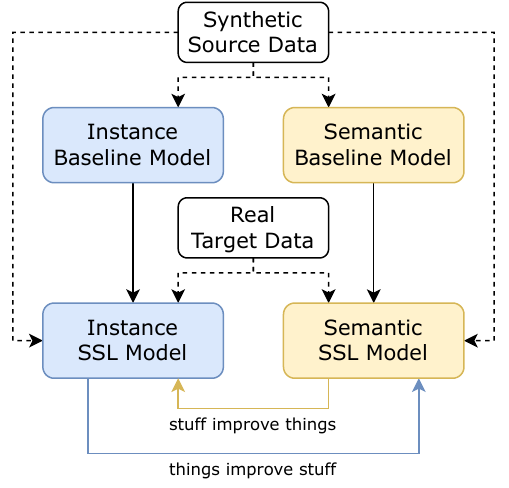}
\caption{Overview of self-supervised panoptic segmentation framework}
\label{fig:self-train_framework}
\end{figure}

Our proposed method implements a multi-branch architecture using separate networks for semantic and instance segmentation, unlike UniDAformer and EDAPS where a unified architecture aims to process the entire image in a single network. Although the unified approach enables simplicity and efficiently extracts masks uniformly, the multi-branch method allows each branch to be optimized for its specific task, potentially leading to better performance in semantic and instance segmentation separately. It's also easier to experiment with different architectures and loss functions for each branch, and allows flexibility in models to be individually fine-tuned depending on new tasks and changes in dataset. UniDAformer and EDAPS employ a transformer based segmentation model which are computationally expensive, particularly for high-resolution images and may require a large amount of training data to generalize effectively. Our method uses computationally efficient embedding based convolutional networks like U-Net and DeepLab, designed for segmentation  which capture local information well, although they may struggle with capturing long-range dependencies across the entire image. Figure \ref{fig:self-train_framework} illustrates the proposed framework for domain adaptive self-supervised panoptic segmentation using a multi-branch architecture.

\part{Methodology}

\chapter{Synthetic and Real Datasets}

\begin{figure}[h]
\centering
\subfloat[Cityscapes]{\includegraphics[trim={5 5 5 5},clip, width=0.33\textwidth]{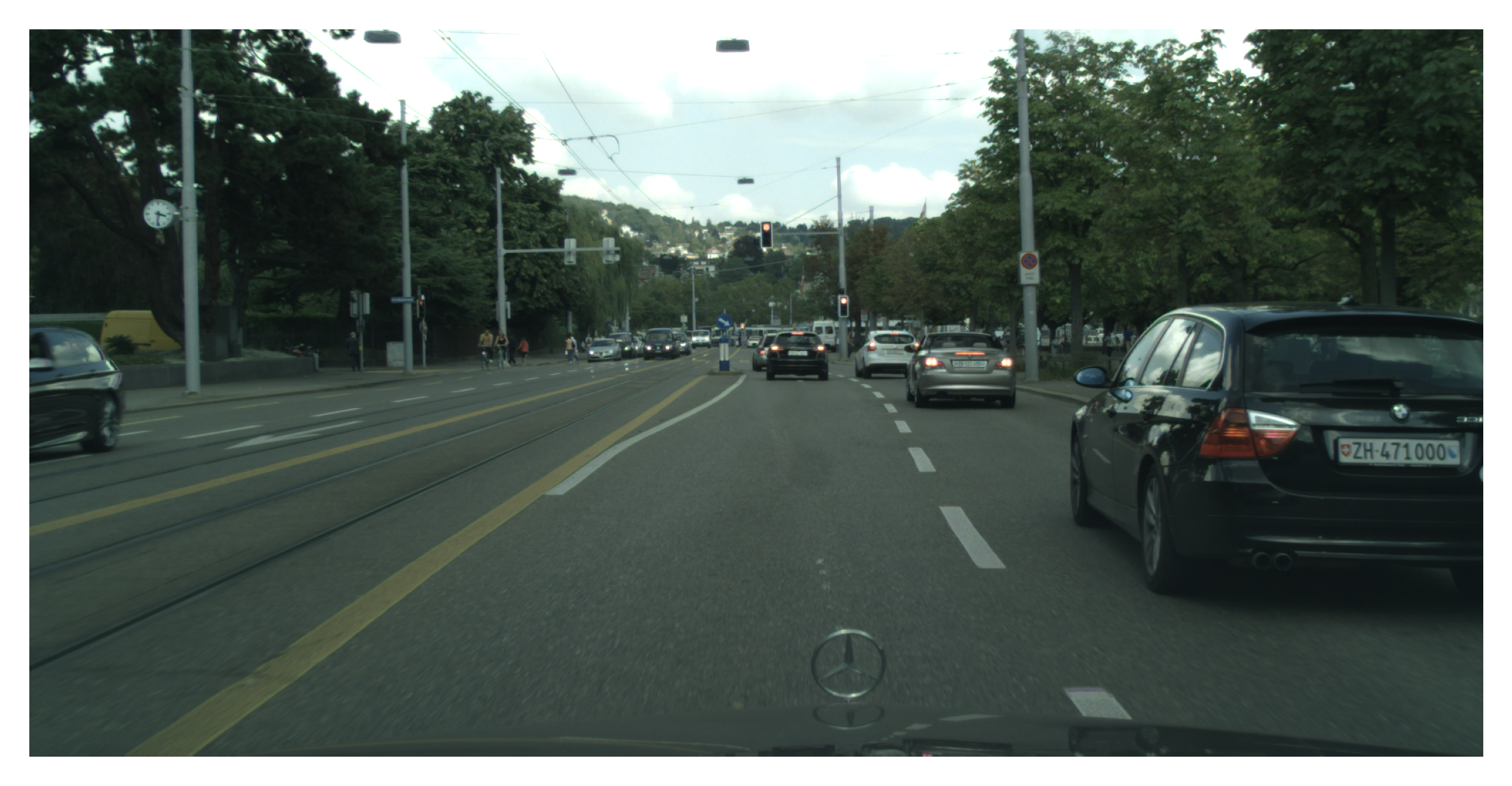}\label{fig:city}}
\hfill
\subfloat[Synthia]{\includegraphics[trim={5 38 5 5},clip, width=0.33\textwidth]{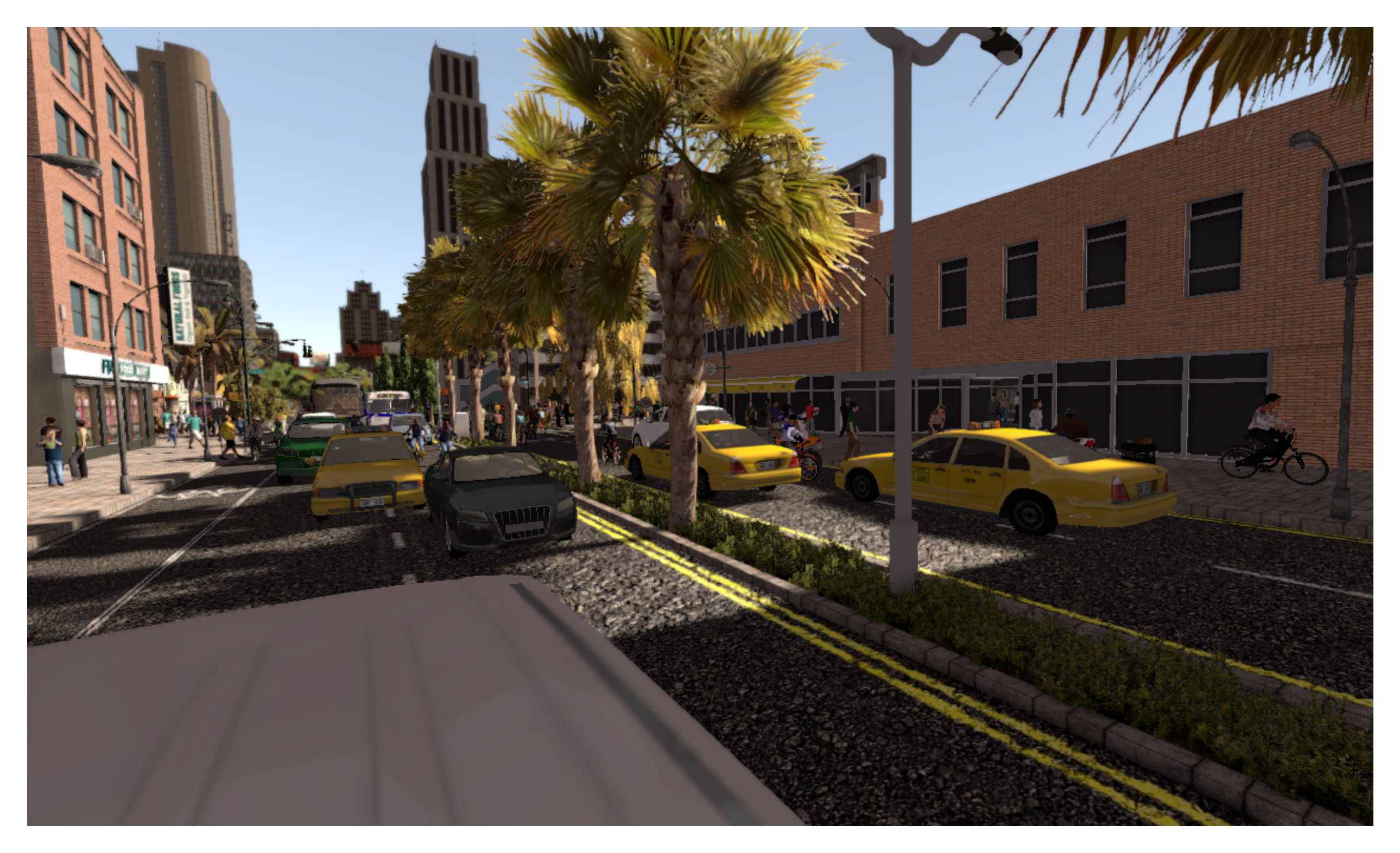}\label{fig:synth}}
\hfill
\subfloat[GTA-V]{\includegraphics[trim={5 22 5 5},clip, width=0.33\textwidth]{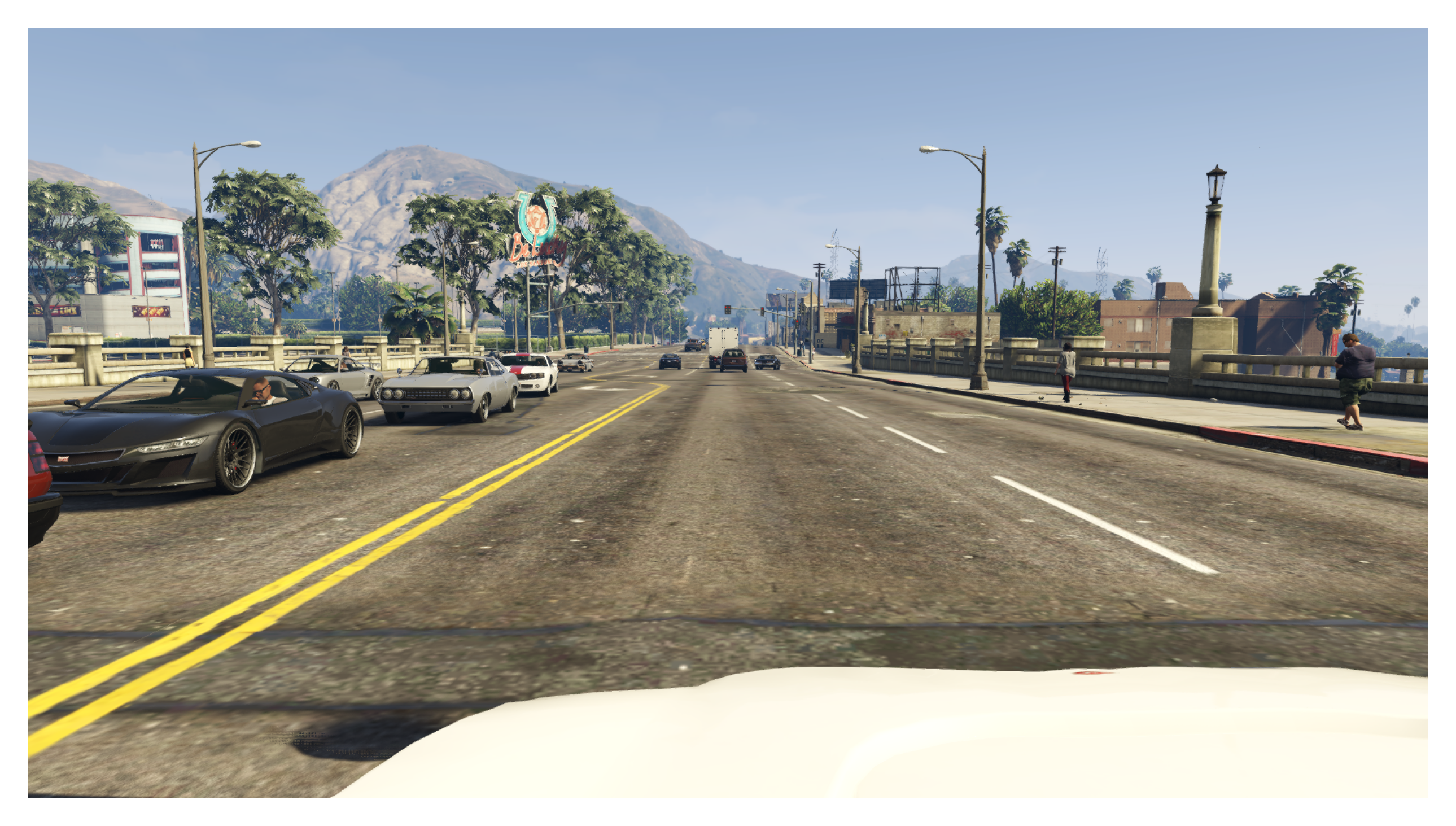}\label{fig:gta}}
\caption{Comparison of synthetic and real datasets}
\label{fig:dataset}
\end{figure}

\begin{figure}[h]
\centering
\includegraphics[width=0.99\textwidth]{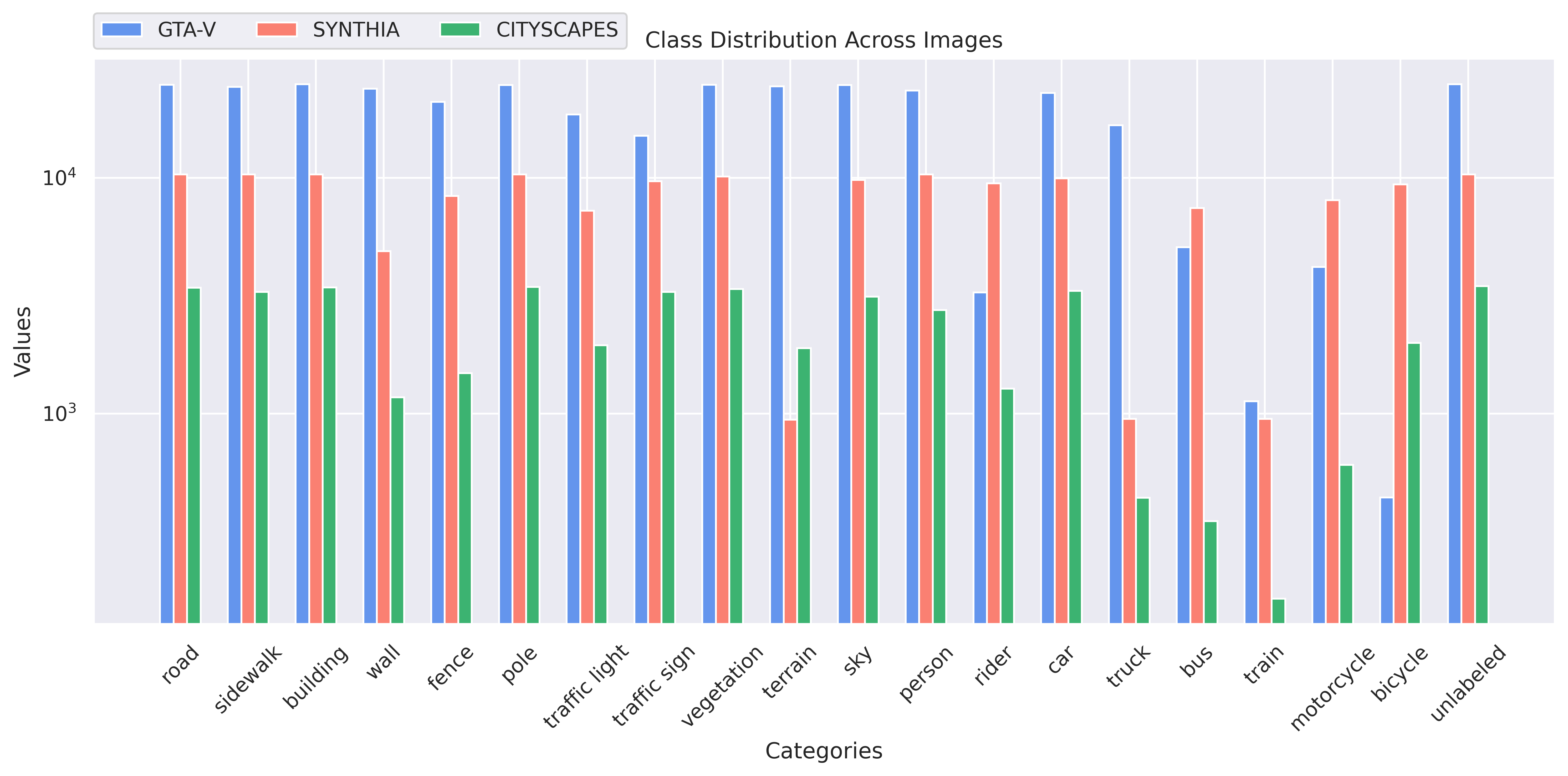}
\caption{Distribution of classes across images for all datasets}
\label{fig:masks}
\end{figure}

The baseline semantic and instance models are trained on synthetic datasets used as source domain, containing large amount of fully annotated high quality data. The models are then iteratively fine-tuned on the target domain's real dataset which contained limited annotated data. Synthetic data can be generated to resemble the target domain, but it might not capture the exact characteristics of the real world. Two synthetic datasets Synthia \cite{ros2016Synthia} and GTA-V \cite{richter2016playing} are used for pre-training the source domain model and one real world dataset Cityscapes \cite{cordts2016Cityscapes} is used for self-training the target domain model. Figure \ref{fig:dataset} shows sample images from the three datasets. The list of classes consistent with all the datasets include: \{\textit{unlabelled, sky, building, road, sidewalk, fence, vegetation, pole, car, traffic sign, person, bicycle, motorcycle, traffic light, terrain, truck, train, rider, bus, wall}\}. Figure \ref{fig:masks} shows how many images in the datasets contain the listed classes and it is clear that GTA-V and Synthia datasets are huge compared to Cityscapes.

The Synthia dataset \cite{ros2016Synthia}, also known as SYNTHetic collection of Imagery and Annotations, has been developed with the primary objective of assisting in solving challenges related to semantic and instance segmentation and scene comprehension within the context of driving scenarios. This dataset is a collection of photo-realistic, finely detailed pixel-level semantic annotation frames that have been synthesized from a virtual urban environment. It contains 9,000 random images of frame resolution 1280×760 pixels with labels compatible with the real world Cityscapes dataset. Note that even though the train and truck classes are present in the Synthia dataset the semantic annotations for them are missing, i.e. the instance ground-truth annotations include truck and train class, but they are not labelled in the semantic part. The GTA-V dataset \cite{richter2016playing} contains 24966 synthetic images with fine and accurate semantic annotation. These images are rendered through the open-world video game GTA-V and are entirely from the car perspective in the streets of American-style virtual cities. Each frame has a resolution of 1914×1052 pixels and there are 19 semantic classes which are compatible with the ones of real world Cityscapes dataset.

The Cityscapes dataset \cite{cordts2016Cityscapes} is collection of real world images with pixel level and instance level semantic labelling. It is comprised of a large, diverse set of stereo video sequences recorded in streets from 50 different cities and 2975 training and 500 validation images of resolution 2048×1024 pixels have high quality pixel-level annotations. It is also accompanied by comprehensive evaluation metrics and benchmarking tools, aiding researchers in comparing and assessing the performance of different algorithms. 

Data augmentations are applied to the source and target datasets to improve the semantic and instance model's ability to generalize across different domains. This included a range of transformations like rotation, scaling, cropping, and noise, applied to training data which augment dataset diversity and exposing the model to various scenarios. Augmentations act as regularization, introducing noise to prevent overfitting and encouraging the model to focus on essential features which improves generalization to unseen data and enhances robustness to real-world variations \cite{shorten2019survey} \cite{zhao2017pyramid}. Augmentations reduce the effort and cost of creating labelled datasets by generating additional training samples from existing ones for tasks like object detection and segmentation. In transfer learning and domain adaptation, augmentations align source and target domains by simulating domain-specific variations, which is important when applying pre-trained models to novel domains. Some augmentations applied to the the source and target dataset included: image resizing with nearest-neighbor interpolation to simulate variations in the field of view, flips to handle mirrored cases, color jittering and random grayscale conversion along with gaussian blur to simulate varying levels of appearance and focus.

\chapter{Network Architectures}

\section{Semantic Model} \label{semantic_model}
The goal of the semantic model is to predict semantic segmentation masks for real images without receiving any ground-truth annotation for the objects present in the images at training time. To achieve this, the semantic model assumes access to synthetic data which has fully and finely annotated ground-truth label masks. This synthetic set of images is perceptually different from from natural images and hence the model has to generalize to the discrepancy across domain distributions. To achieve this, semantic model utilizes the well known data augmentation transformation techniques from supervised learning methods which include photo-metric jitter, flipping and multi-scale cropping. It also assumes that the semantic output masks from the input images are invariant under photo-metric transform and equivariant under spatial similarity transformation as shown by \cite{araslanov2021self}. The Cityscapes dataset contains low image-level classes like "truck" and "bus", and low pixel-level classes like "traffic light" and "pole". To handle these special cases the semantic model encourages smaller thresholds for determining their pseudo-labels and increase their participation to the gradient with a focal loss.

\begin{figure}[h]
\centering
\includegraphics[width=0.80\textwidth]{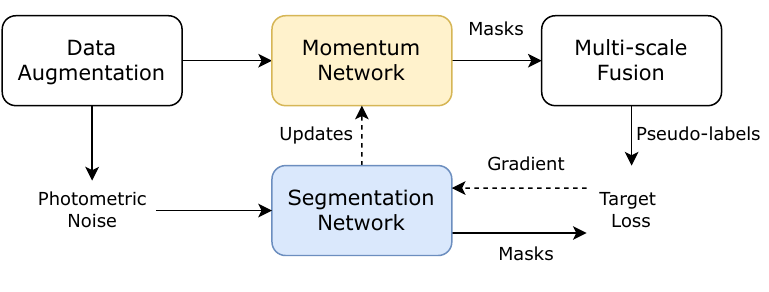}
\caption{Semantic segmentation self-training framework}
\label{fig:sem_steps}
\end{figure}

\begin{figure}[h]
\centering
\includegraphics[width=0.90\textwidth]{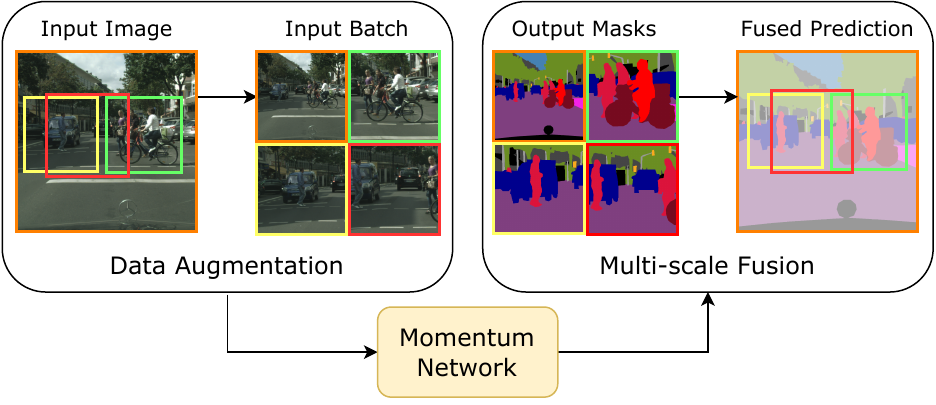}
\caption{Input image and output masks of momentum network}
\label{fig:data_fuse}
\end{figure}

The overview for semantic model's self-training framework is illustrated in figure \ref{fig:sem_steps}. The baseline semantic model is pre-trained on synthetic source data using supervised learning with cross-entropy loss, i.e. using labelled data where each pixel in an input image is associated with a corresponding label that represents the object or class it belongs to. Here cross-entropy loss is defined as: 
\begin{equation}
L_{CE} = - \sum_{i=1}^{n} t_i log(p_i),
\end{equation}
where $t_i$ is the ground-truth label and $p_i$ is the softmax probability for class $i$. Then for the self-training part, the segmentation network is jointly trained on the source and target data with random crops and flips for each sample image along with random photometric noise. In parallel, the same image batch is fed to the momentum network but without the photometric noise. The outputs produced by this network are used further to generate pseudo-labels and these pseudo-labels are in turn used to train the segmentation network. The momentum network is an exponentially moving average of the segmentation network which greatly improves the training stability. For each softmax prediction of the momentum network, a prior estimate $\chi_c$ is computed using the probability that pixels in sample image belongs to a specific class with respect to the masked prediction. The input batch samples a number of randomly flipped multi-scale crops for a given target sample and then re-projects and averages the segmentation maps produced by the momentum network to create a refined prediction for the original sample. This refined prediction is used to select pixels for self-supervision using an adaptive threshold which selects a lower confidence threshold for rare classes and a higher threshold for the more frequent classes. The joint training of the segmentation network is done with with stochastic gradient descent (learning rate $=2.5 \times 10^{-4}$, momentum $=0.9$ and weight decay $=5 \times 10^{-4}$) using the cross-entropy loss for the source and a focal loss for the target data. The focal loss incorporates a focal multiplier \cite{lin2017focal} to improve the contribution of the rare classes in the gradient signal and the moving class prior $\chi_c$ regulates the focal term. An exponential moving average is stored each training iteration $t$ with a momentum $\gamma_\chi \in[0,1]$:

\begin{equation}
\chi_c^{t+1}=\gamma_\chi \chi_c^t+\left(1-\gamma_\chi\right) \left(\frac{1}{ab}\sum_{i,j}{m_{c, n,i,j}}\right),
\end{equation}
where $m_{c, n,:,:}$ is the mask prediction for class $c$ of shape $a \times b$. The focal loss is then formulated as:

\begin{equation}
L_{focal}(\bar{m}, m \mid \phi)=-m_{c^*, n}\left(1-\chi_{c^*}\right)^\lambda \log \left(\bar{m}_{c^*, n}\right),
\label{eq:focal}
\end{equation}
where $\bar{m}$ and $m$ are the predictions of the segmentation and momentum network respectively with parameters $\phi$, pseudo-label $c^*$, and hyperparameter $\lambda$ of the focal term. High values of $\lambda($ i.e. $>1)$ increase the relative importance on the rare classes, while setting $\lambda=0$ disables the focal term. The loss is also regularized with confidence value of the segmentation $\bar{m}_{c^*, n}$ and momentum $m_{c^*, n}$ networks. The parameters of the momentum network $\psi$ are periodically updated as:

\begin{equation}
\psi_{t+1}=\gamma_\psi \psi_t+\left(1-\gamma_\psi\right) \phi,
\end{equation}
where $\phi$ are the parameters of the semantic model and $\gamma_\psi$ regulates the speed of updates. Giving low $\gamma_\psi$ values result in faster, but unstable training, while high $\gamma_\psi$ leads to a untimely and substandard convergence, thus $\gamma_\psi$ is kept moderate and the momentum network is updated only every $T$ iterations. For the momentum network $\lambda=3$, $\gamma_\psi = 0.99$ and $T = 100$ are fixed in all the experiments. Figure \ref{fig:data_fuse} helps visualize the input and output of the momentum network where the output masks are re-projected from back to the original image canvas of size a×b. In the output image each pixel in the overlapping areas average and fuse the predictions to produce the pseudo-label targets.

The implementation of the semantic model framework is in PyTorch, adopting the DeepLabv2 segmentation architecture using ResNet-101 backbone initialized from models pre-trained on ImageNet. The semantic model is also extended to explore the newer DeepLabv3 and DeepLabv3+ segmentation architectures. ImageNet \cite{deng2009imagenet} is a large-scale dataset and benchmark consisting of over a million labelled images spread across thousands of categories. 

ResNet-101 \cite{he2016deep} uses CNN architecture introduced to address vanishing gradients during training. Traditional neural networks suffer from degradation, where the accuracy of the model saturates or even degrades as the network depth increases and ResNet-101 tackles this issue by introducing residual blocks, which enable the learning of residual functions instead of direct mappings. These residual blocks use skip connections that bypass various number of layers allowing gradients to flow in a simple manner during backpropagation and mitigates the vanishing gradient problem. The network starts with a single CNN layer followed by max pooling layer to reduce the spatial dimensions of the input. The first stage of the network contains 3 residual blocks each containing 3 CNN layers, followed by the second stage which contains 4 residual blocks containing 4 CNN layers each. The third stage contains 23 residual block each containing 3 CNN layers and the final stage contains 3 residual blocks of 3 CNN layers each.

\begin{figure}[h]
\centering
\includegraphics[width=0.70\textwidth]{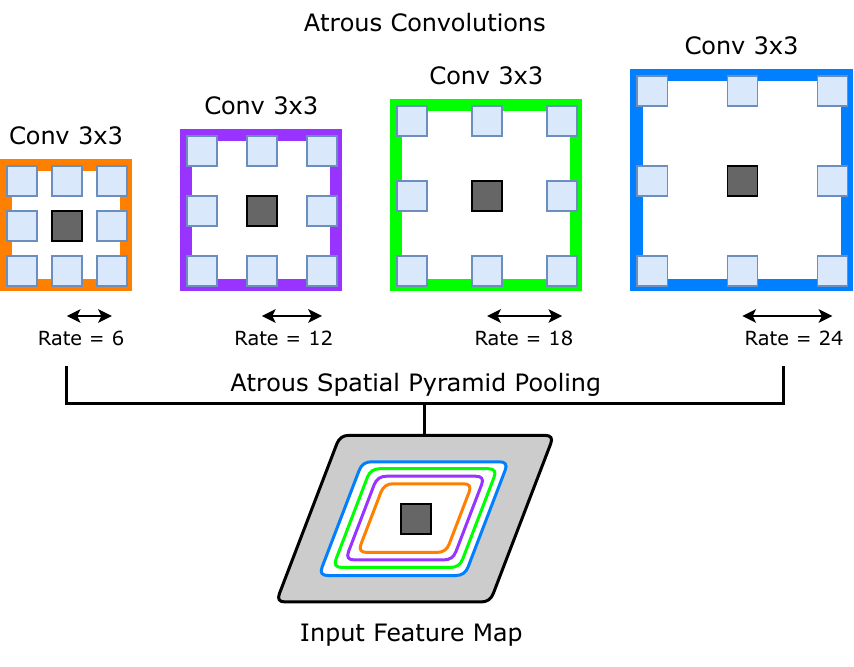}
\caption{Atrous convolutions and atrous spatial pyramid pooling}
\label{fig:aspp}
\end{figure}

DeepLabv2 \cite{chen2017deeplab} segmentation architecture uses \textbf{Atrous Spatial Pyramid Pooling} (ASPP) module which employs multiple parallel dilated convolutions with different dilation rates to capture features at various scales. This allows it to retain multi-scale contextual information and generate high resolution feature maps. The ASPP module consists of four parallel atrous convolution layers with different dilation rates of 6, 12, 18, and 24. The output of each branch is concatenated to form a combined input feature map covering features extracted at multiple scales. Figure \ref{fig:aspp} shows visualization of atrous convolution with different rates and how the ASPP module concatenates these convolutions to produce input feature maps. DeepLabV2 uses fully convolutional network based on the ResNet-101 architecture as the encoder module since the encoder-decoder structure recovers object representation effectively. The atrous convolution is applied to the last two blocks of ResNet layers in order to maintain feature map resolution during encoding. The decoder module upsamples the output of the ASPP module using bi-linear interpolation to recover sharp object boundaries in the segmentation result. The criss-cross atrous convolution enhances connectivity between distant pixels and helps the model learn more meaningful features for segmentation. This enables better handling of objects at different sizes in the image and improved semantic segmentation accuracy.

DeepLabv3 \cite{chen2017rethinking} refined the ASPP module by adding batch normalization and global average pooling to improve segmentation performance. The main new component in Deeplabv3 is the incorporation of depth-wise separable convolution in the encoder module. Depthwise separable convolution splits a standard convolution into two layers: a depthwise convolution to filter each input channel separately; and a pointwise convolution to combine the outputs. This factorized convolution operation reduces computation substantially while maintaining similar representational capacity. Orignally, Deeplabv3 employed an improved version of Xception \cite{chollet2017xception} as its encoder network, however ResNet-101 is used in this case for consistency of architecture. Deeplabv3 modifies ResNet-101 by applying atrous convolution with upsampled filters to the last few layers, maintaining feature map resolution for denser encoding. The decoder module bi-linearly upsamples and concatenates the low resolution encoder output with the image pyramid pooling output which gives sharper segmentation masks along object boundaries. 

DeepLabv3+ \cite{chen2018encoder} extended DeepLabv3 by incorporating a feature pyramid network inspired decoder which uses feature maps from different levels of the backbone network to capture both high-level semantics and finer boundary details more effectively. This design enables the model to handle objects of varying sizes more accurately. The decoder employs a 1x1 convolution layer to project the low resolution encoder features to a higher dimensional space. These features are then bi-linearly upsampled by a factor of 4 and concatenated with the corresponding high resolution encoder features. Additional 3x3 convolutions are applied on the concatenated features to propagate higher resolution features from the encoder to the decoder output. Orignally, DeepLabv3+ used MobileNetV2 \cite{sandler2018mobilenetv2} network for the encoder but again it is replaced with ResNet-101 for architecture consistency. 

\newpage

\section{Instance Model} \label{instance_model}

\begin{figure}[h]
\centering
\includegraphics[width=0.80\textwidth]{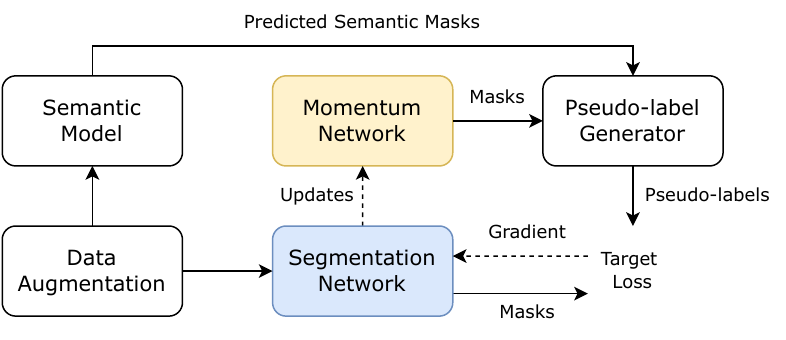}
\caption{Instance segmentation self-training framework}
\label{fig:inst_steps}
\end{figure}

The goal of the instance model is to predict instance segmentation masks for real images without receiving any ground-truth annotation for the objects present in the images at training time. Instance model is implemented specifically for object level classes \{\textit{car, bus, truck, train, person, rider, bicycle, motorcycle}\}. This implemented self-trained instance model is an updated version of the semi-supervised SPOCO model by \cite{wolny2022sparse}. Figure \ref{fig:inst_steps} shows the overview of the instance model's self-training framework. Similar to the semantic model, the instance model consists of two networks with shared weights and parameters: a segmentation network and a momentum network. They take RGB image batches as input and output a dense embedding vector for each pixel. During training, the weights of the momentum network are updated online as an exponential moving average of the segmentation network weights. This momentum update allows the second network to accumulate more stable representations over time, which is useful for generating pseudo-labels for self-training. The input data for the instance model goes through similar augmentation transformations and normalizations as the semantic model.

For self-training, the frozen momentum network is used to predict dense embeddings on unlabelled images and a pseudo-label generator creates soft targets. The idea of freezing the momentum network for some iterations and replacing it with the current network is inspired from \cite{mnih2013playing}. Additionally, the previously trained semantic segmentation model provides high-level guidance through predicted category masks on the unlabelled data. The set of anchor points are placed within these embeddings of predicted semantic masks to remove unwanted classes from affecting the model. Different clustering algorithms are applied on these embeddings to cluster them into groups representing object instances. These clustering algorithms are explained in detail in section \ref{clustering}. The generated pseudo-labels serve as ground truth instance masks for computing the contrastive loss during self-training. A brief introduction to contrastive learning is given in section \ref{ssl}.

\begin{figure}[h]
\centering
\subfloat[Input Image]{\includegraphics[trim={50 40 70 30},clip,width=0.495\textwidth]{images/scene0.png}\label{fig:anchor0}}
\hfill
\subfloat[Anchor Points]{\includegraphics[trim={50 45 70 30},clip,width=0.496\textwidth]{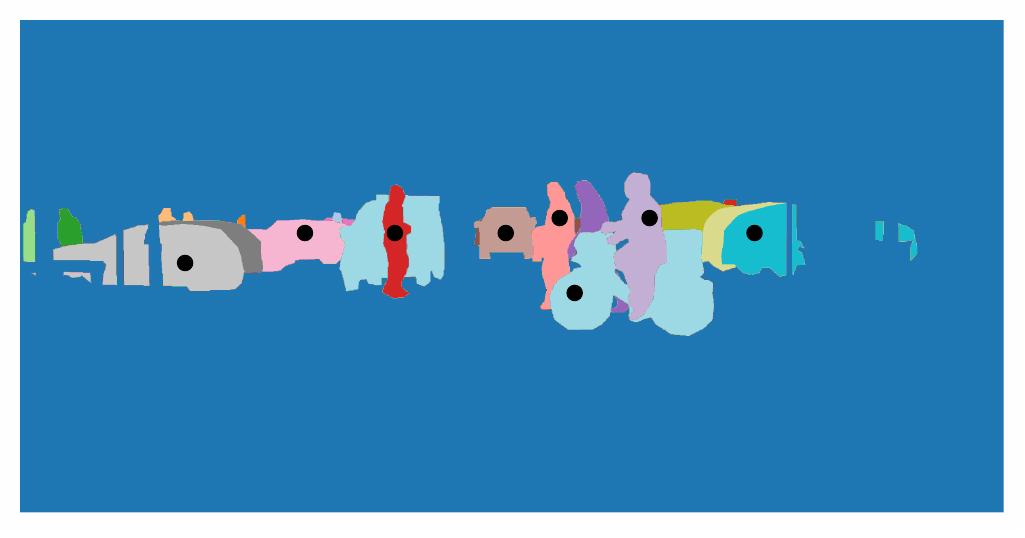}\label{fig:anchor1}}
\hfill
\subfloat[Push and Pull Forces]{\includegraphics[width=0.50\textwidth]{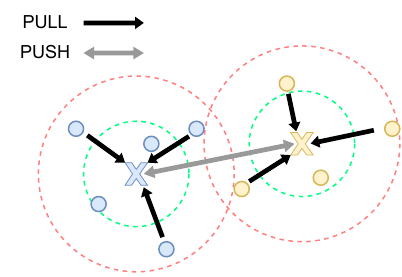}\label{fig:force}}
\caption{Overview of sampling anchor points on input image and how push and pull forces act on embedding clusters of selected instance pixels.}
\label{fig:anchor}
\end{figure}

The training loss is a weighted combination of two components: a contrastive loss and a consistency loss.  The contrastive loss is the main loss that enables instance discrimination in the embedding space by using the pull force and the push force terms: 

\begin{equation}
L_{\text {pull }} =\frac{1}{C} \sum_{k=1}^C \frac{1}{N_k} \sum_{i=1}^{N_k}\left[\left\|\boldsymbol{\mu}_k-\boldsymbol{e}_i\right\|_2-\delta_v\right]_{+}^2,
\label{eq:pull}
\end{equation}

\begin{equation}
L_{\text {push }} =\frac{1}{C(C-1)} \sum_{\substack{k=1 \\ k \neq l}}^C \sum_{\substack{l=1}}^C\left[2 \delta_d-\left\|\boldsymbol{\mu}_k-\boldsymbol{\mu}_l\right\|_2\right]_{+}^2,
\label{eq:push}
\end{equation}
where $C$ is the number of objects in the image, $N_k$ the size of object $k$, $\delta_v$ and $\delta_d$ are margin hyper-parameters, and $[x]_{+}=\max (0, x)$ is the rectifier function. The pull force brings the object’s pixel embeddings closer to their mean embedding $\mu_k$, while the push force pushes the objects away, by increasing the distance between mean embeddings. Both terms are hinged which means embeddings within the area of $\delta_v$ of the mean embedding $\mu_k$ are not pulled to it. Similarly, mean embeddings which are further away than $2\delta_d$ are no longer pushed. Figure \ref{fig:anchor} shows the basic idea behind selecting random anchor points to obtain anchor embeddings and how the pull force pulls embeddings (blue and yellow dots) towards their corresponding cluster centers and the push force pushes cluster centers away from each other. These forces are hinged to distances denoted by circles around the cluster center. Note that during the actual process, multiple anchor points are distributed on each potential object.

The contrastive loss works by sampling a set of anchor points in the embedding space by taking the mean vector of randomly selected points in an image. For each anchor, a distance map is computed by taking the L2 norm between the anchor and all points in the embedding space. This distance map is converted into a soft mask using a Gaussian kernel, with the distance spread controlled by a hyper-parameters $\delta_v$ and $\delta_d$. These soft masks activate on the instances closest to the corresponding anchor. The contrastive loss encourages agreement between the soft masks and the ground truth instance segmentation masks. The Dice loss \cite{sudre2017generalised} \cite{milletari2016v} is used for the objective to minimize the discrepancy between the predicted soft masks $S_k$ and the the corresponding ground-truth masks $I_k$. The object level loss is given by:

\begin{equation}
L_{o b j}=\frac{1}{C} \sum_{k=1}^C D\left(S_k, I_k\right),
\label{eq:obj}
\end{equation}
for Dice loss $D$:

\begin{equation}
D\left(S_k, I_k\right)=1-\frac{2 \sum_i^N p_i q_i}{\sum_i^N p_i^2+\sum_i^N q_i^2},
\label{eq:dice}
\end{equation}
where $p_i$ and $q_i$ represent pairs of pixel values of predicted mask $S_k$ and ground-truth mask $I_k$. Another "unlabelled push" loss term is used to push clusters of unlabelled region $(U)$ which can contain both background and unlabelled instances:

\begin{equation}
L_{U_{push}}=\frac{1}{C} \sum_{k=1}^C \frac{1}{N_U} \sum_{i=1}^{N_U}\left[\delta_d-\left\|\mu_k-e_i\right\|\right]_{+}^2,
\label{eq:unlabel_push}
\end{equation}
where $C$ is the number of labelled clusters and $N_U$ is the number of pixels in the unlabelled region $U$. 

The consistency loss enforces rotation equivariance in the embedding space, which works by rotating the input image randomly between 0-180 degrees and feeding it through the momentum network to obtain rotated embeddings. The resulting vector fields from the segmentation $(f)$ and momentum $(g)$ network come from the same input geometry, which result in consistent instance segmentation after clustering. Anchor points are randomly sampled and projected into the $f$ and $g$ embedding spaces, to obtain anchor embeddings and compute two soft masks $S^f$ and $S^g$. Similarly to equation \ref{eq:obj} the embedding consistency is obtained by maximising the overlap of the two masks, using the Dice loss $D$ which encourages the model to produce embeddings that are consistent under rotations. The consistency level loss is given by:

\begin{equation}
L_{con}=\frac{1}{K} \sum_{k=1}^K D\left(S_k^f, S_k^g\right),
\label{eq:consistency}
\end{equation}
where $K$ is the number of sampled anchor points such that the whole region is covered by the union of extracted masks. Combining the losses in equations \ref{eq:pull}, \ref{eq:push}, \ref{eq:obj}, \ref{eq:unlabel_push}, and \ref{eq:consistency} the instance self-train loss is obtained $\left(L_{self-train}\right)$:

\begin{equation}
L_{self-train}=\alpha L_{p u l l}+\beta L_{p u s h}+\lambda L_{o b j}+\gamma L_{U_{push}}+\delta L_{con},
\label{eq:contrastive}
\end{equation}
where $\alpha=\beta=\gamma=\lambda=1$ as used by \cite{wolny2022sparse} and \cite{de2017semantic}, $\delta \in \{0.0,1.0\}$ explored in table \ref{tab:inst_miou}, and the pull and push margin parameters are set to $\delta_v=0.5, \delta_d=1.5$. 

\begin{figure}[h]
\centering
\includegraphics[width=0.90\textwidth]{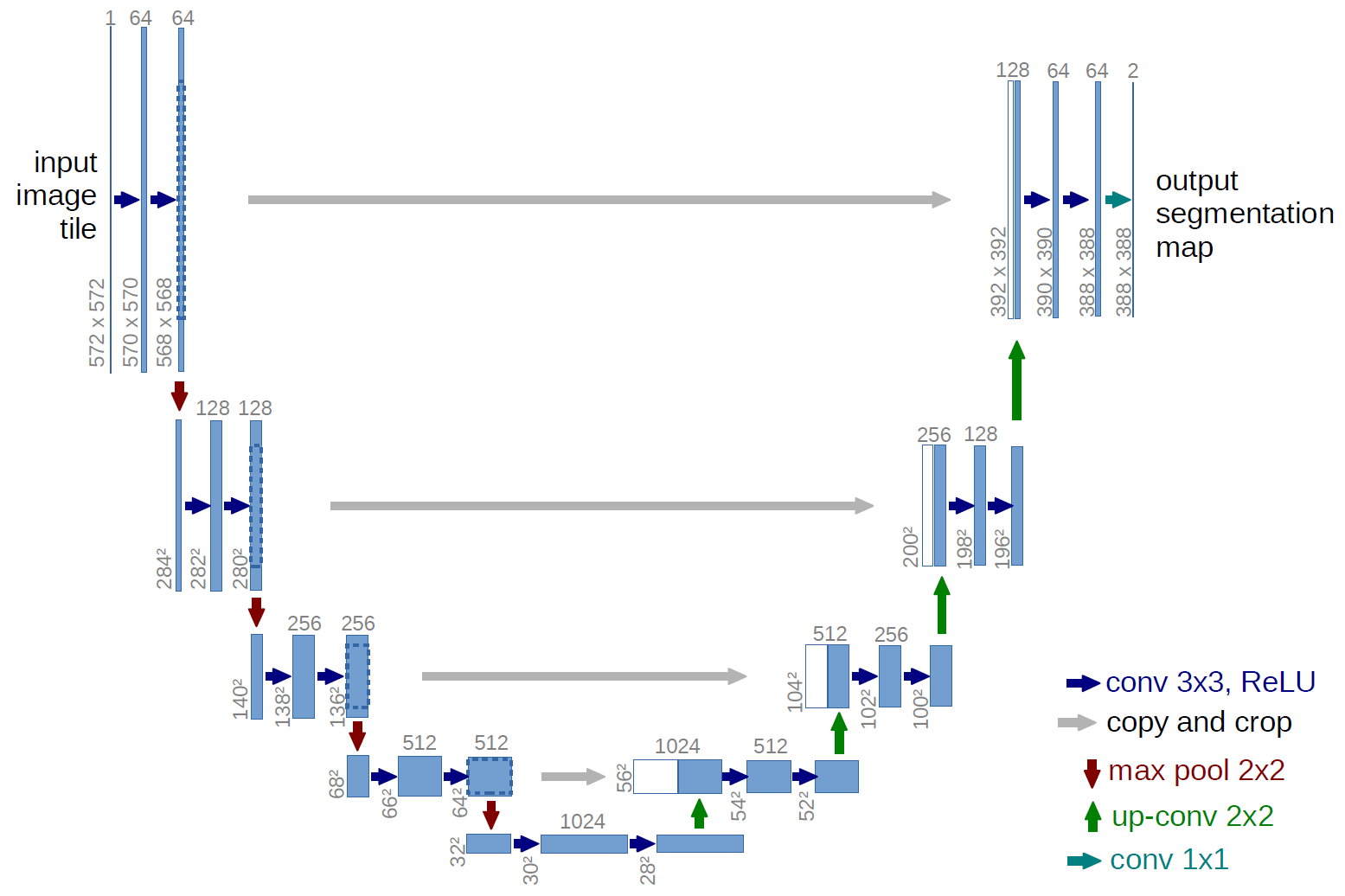}
\caption{U-Net Architecture \cite{ronneberger2015u}}
\label{fig:unet}
\end{figure}

Both the segmentation and momentum networks have an encoder-decoder U-Net \cite{ronneberger2015u} architecture built on top of a ResNet-101 \cite{he2016deep} backbone. The segmentation network is trained end-to-end on the combined loss using stochastic gradient descent. The momentum network and pseudo-label generation combine to achieve a self-trained instance segmentation framework through contrastive learning, consistency regularization, and clustering. U-Net \cite{ronneberger2015u}, illustrated in figure \ref{fig:unet} is designed to address the challenge of pixel-wise image segmentation. The first part of the U-Net architecture serves as an encoder, responsible for capturing hierarchical features from the input image. It consists of several CNN layers followed by pooling operations. These layers progressively reduce the spatial resolution of the input while increasing the depth of feature maps, which helps extract abstract and context-rich representations of the input image. It also includes skip connections, which directly link the corresponding layers from the encoder to the decoder. They enable the decoder to access high-resolution feature maps that contain fine details about the input, which helps in preserving spatial information during the upsampling process. The second part of the U-Net architecture acts as a decoder and involves a series of upsampling operations, often implemented as transposed convolutions or bilinear upsampling, followed by CNN layers. The decoder's role is to reconstruct the segmented output map from the abstract features generated by the encoder. The decoder leads to a final layer that employs a softmax activation function to produce the pixel-wise classification probabilities. In segmentation tasks, this final layer outputs a probability distribution over the different classes for each pixel in the input image. Training is performed through stochastic gradient descent optimizer and backpropagation, optimizing the network's parameters to minimize the loss.

\chapter{Pseudo-label Generation}

The semantic and instance segmentation models generate pseudo-labels during the self-training loop. The models make predictions on unlabelled data to produce pseudo-labels and these are treated as ground-truth labels for training the model.

The self-train semantic model works by using predictions from a trained teacher model to generate targets for the student model. Firstly, the unlabelled input images are passed through the teacher model to obtain logits and softmax probabilities and the model runs in evaluation mode with gradients disabled to avoid updating its parameters. Next, these softmax probabilities are processed to generate pseudo-labels and confidence scores at the pixel-level by extracting the predicted class. The maximum confidence scores are used to only retain high confidence predictions i.e. top pixels and the thresholds are computed dynamically based on the class-wise confidence scores tracked during training. This prevents noisy pseudo-labels for uncertain predictions and pixels below the threshold are assigned the void/ignore label. The thresholded pseudo-labels and confidence scores are obtained from this method. Performing thresholding based on the knowledge of previous predictions allows the model to filter out low confidence pixels, helping prevent confirmation bias. Tracking running confidence scores allows adjusting the thresholds dynamically as the teacher model's predictions evolve. These generated targets are then used to compute the focal loss \ref{eq:focal} between the student model predictions and the generated pseudo-labels. This drives the student to learn from the teacher's knowledge in the self-training loop. Algorithm \ref{alg:1} gives a general overview of the steps to used by the semantic model to generate pseudo-labels.

\begin{algorithm}
\caption{Semantic Model Pseudo-label Generation}
\textbf{Input:} Predicted probabilities: $probs$ \\
\textbf{Output:} Pseudo-labels: $P$
\begin{algorithmic}[1]
\Function{gen\_pseudo\_labels}{$probs$}
\State Calculate maximum pixel confidence: $con,idx \gets probs.max()$
\State Calculate top pixels: $top \gets con,idx$
\State Calculate threshold value for classes: $thresh_c \gets top_c$
\State Initialize pseudo-labels: $P$
\For{pixel \textbf{in} $P$}
\If{$probs_{pixel} < thresh_c$}
\State $top_c = 255$        \#void/ignore
\EndIf
\EndFor
\State \Return Pseudo-labels $P$
\EndFunction
\end{algorithmic}
\label{alg:1}
\end{algorithm}

\newpage

The self-train instance model's pseudo-label generation process starts by looping through each of the classes present in the input image. For each class, a binary mask is created by selecting pixels labelled as that object in the predicted semantic segmentation mask. This focuses the pseudo-label generation on regions of the image belonging to only instance class categories: \{\textit{car, bus, truck, train, person, rider, bicycle, motorcycle}\}. Next, the image embeddings inside this binary mask are clustered using clustering algorithms, explained in section \ref{clustering}. The clustering algorithm initializes multiple random anchors inside the semantic mask and extracts instances around each anchor within a few iterations. This results in a set of instance segmentation embeddings for the given class. After clustering, the generated instance masks are merged if their mean embeddings are close to each other, which removes duplicate detections and combines instances that were over-segmented by the clustering. The merged instance masks become the pseudo-labels and this process is repeated independently for each class. The final class-wise pseudo-labels are fused into full image pseudo-labels. An additional filtering step is applied to remove unstable instances based on threshold value which helps select only the most consistent, high-quality pseudo-labels for training. Algorithm \ref{alg:2} gives a general overview of the steps to used by the instance model to generate pseudo-labels.

\begin{algorithm}
\caption{Instance Model Pseudo-label Generation}
\textbf{Input:} Embeddings: $E$, Semantic Mask: $S$, Instance Size: $minSize$  \\
\textbf{Output:} Pseudo-labels: $P$
\begin{algorithmic}[1]
\Function{gen\_pseudo\_labels}{$E, S, minSize$}
\State Initialize class IDs: $IDs$
\State Initialize pseudo-labels: $P$
\For{$ID$ \textbf{in} $IDs$}
\State $mask \gets (S == ID)$
\If{$mask.sum() > minSize$}
\State $P_{ID} \gets$ Place anchors and cluster embeddings
\State $P_{ID} \gets$ Select stable instances
\EndIf
\State Store $P_{ID}$ in $P$
\EndFor
\State \Return Merged pseudo-labels $P$
\EndFunction
\end{algorithmic}
\label{alg:2}
\end{algorithm}

\chapter{Experiments}
\section{Mixing Synthetic Datasets} \label{mix}

The utilization of synthetic datasets Synthia and GTA-V in training the baseline semantic and instance models introduced certain constraints. The Synthia dataset lacked semantic ground-truth annotations for the truck and train classes, yet contained instance annotations for these categories. This absence of annotations resulted in the semantic self-trained model on the Cityscapes dataset also unable to generate predictions for the truck and train classes due to the lack of a prior distribution to inform the model's learning process. Simultaneously, the GTA-V dataset possessed complete semantic ground-truth annotations to all the classes in Cityscapes dataset but lacked instance ground-truth annotations altogether, making training of the instance baseline model solely on the GTA-V dataset infeasible.

The idea to tackle these limitations is to mix the synthetic datasets in a manner that provided both the semantic and instance baseline models with annotations spanning all classes, while minimizing disparities in prior distributions. This amalgamated synthetic dataset is constructed by incorporating GTA-V images into the Synthia dataset, focusing only on instances where both the truck and train classes are present. This curation ensured the presence of semantic ground-truth masks for the instance classes of trucks and trains within the synthetic dataset. It is balanced with the aim to provide sufficient information for the baseline model's learning process without overrunning it with excess representations of purely semantic objects such as trees, buildings, and vegetation etc. A total of 948 images are selected from the GTA-V dataset which contained both truck and train classes, and images containing only truck or train instances are disregarded. The total number of training samples for the baseline semantic model jumped from 9400 to 10349. Mixing the synthetic datasets improved the performance of the semantic model along with fixing the problem of missing annotations for truck and train classes.

\begin{table}[h]
\centering
\begin{tabular}{@{}l|cccc|ccc@{}}
\toprule
& \multicolumn{2}{c}{\textbf{Araslanov}} & \multicolumn{2}{c|}{\textbf{Reproduced}} & \multicolumn{3}{c}{\textbf{Mixed Synthetic Data}} \\ \midrule
\textbf{DeepLab} 
& \multicolumn{2}{c}{V2} & \multicolumn{2}{c|}{V2} & V2 & V3 & V3+ \\
\textbf{Dataset}
& SYN & GTA & SYN & GTA & MIX & MIX & MIX \\ \midrule
\textbf{Baseline}
& 36.3 & 40.8 & 35.1 & 41.8 & 40.8 & 39.6 & 40.7  \\
\textbf{Self-train} 
& 52.6 & 53.8 & 51.1 & 53.8 & 52.4$\pm$0.2 & 50.2$\pm$0.1 & 52.6$\pm$0.2 \\ \bottomrule
\end{tabular}
\caption{Total MIoU (\%) on Cityscapes validation for baseline and self-train semantic models}
\label{tab:miou}
\end{table}

The total \textbf{Mean Intersection-Over-Union} (MIoU) score results are shown in table \ref{tab:miou}. IoU is calculated using the formula $\mathrm{IoU} = TP / (TP + FP + FN)$ where TP, FP, FN are true positive, false positive, and false negative respectively, and MIoU is defined as:
\begin{equation}
    \mathrm{MIoU} = \frac{1}{C} \sum_c \mathrm{IoU_c},
\end{equation}
where C is the total number of objects in a class. These scores are calculated by the official Cityscapes evaluation tool on the validation set. In table \ref{tab:miou}, "Araslanov" column shows the original scores reported in their paper \cite{araslanov2021self}, and "Reproduced" column shows the reproduced scores of models trained on same parameters with different random seeds. "Mixed Dataset" column shows scores of different DeepLab models with the mixed synthetic dataset curated in this experiment. Note that the results show higher baseline score for model trained on GTA-V but lower for Synthia. Mixing only 984 images from GTA-V to Synthia dataset added three new classes to the dataset and also significantly increased MIoU scores for the baseline model, compared to purely training on Synthia.

\section{Hierarchical and Density Clustering} \label{clustering}

After the instance model converges, segmentations of input image are generated by clustering the embedding output of the model. Two types of clustering algorithms are explored: hierarchical agglomerative clustering and density-based mean-shift clustering. 

Mean-shift \cite{comaniciu2002mean} clustering is a density-based non-parametric clustering technique that doesn't require specifying the number of clusters beforehand. It works by iteratively shifting the center of a kernel density estimate towards regions of higher data point density, which results in the convergence of data points towards cluster centers i.e. modes in the data distribution. Mean-shift clustering is particularly useful for identifying clusters of varying shapes and sizes in the data, and it doesn't assume any predefined cluster structure. For a given candidate geometric centre $x$ for iteration $t$, the candidate point is updated as $ x^{t+1} = x^t + m(x^t) $, where $m$ is the mean-shift vector that is calculated for each geometric centre which points towards an area of the maximum increase in the density of points, and the equation for $m$ depends on a kernel used for density estimation:

\begin{equation}
m(x) = \frac{\sum_{x_j \in N(x)}K(x_j - x)x_j}{\sum_{x_j \in N(x)}K(x_j - x)} - x,
\end{equation}

where kernel $K(x)$ is equal to 1 if $x$ is small enough or $0$ otherwise, and $K(y-x)$ indicates whether $y$ is in the neighborhood of $x$. 

Agglomerative \cite{lukasova1979hierarchical} clustering is a hierarchical clustering technique that starts with each data point as its own cluster and iteratively merges the closest clusters based on a defined distance metric. The process continues until all data points belong to a single cluster or until a certain number of clusters is reached. This method produces a tree-like structure called a dendrogram involving pairwise distance calculations which can help in understanding the hierarchical relationships between data points. At each clustering step, the newly calculated distance between the merged clusters $A \cup B$ and a new cluster $X$ is given by the proportional averaging of the $d_{A,X}$ and $d_{B,X}$ distances:
\begin{equation}
d_{(A \cup B),X} = \frac{\vert A \vert\cdot d_{A,X} + \vert B \vert \cdot d_{B,X}}{\vert A \vert + \vert B \vert}
\end{equation}

Mean-shift+ \cite{wolny2022sparse} clustering algorithm is implemented since regular mean-shift is extremely slow as it runs solely on the CPU. The mean-shift+ function is implemented to run on the GPU, which can help in optimizing the clustering process and making it run faster as it performs clustering and merging  in parallel. After performing the initial clustering, mean-shift+ applies additional filter and merging logic to refine the clusters and labels based on the computed mean embeddings and the $\delta_d$ distance threshold. This refinement process can help in merging clusters that are close to each other, potentially reducing the number of final clusters.

\begin{figure}[h]
\centering
{\includegraphics[trim={110 110 110 110},clip, width=0.32\textwidth]{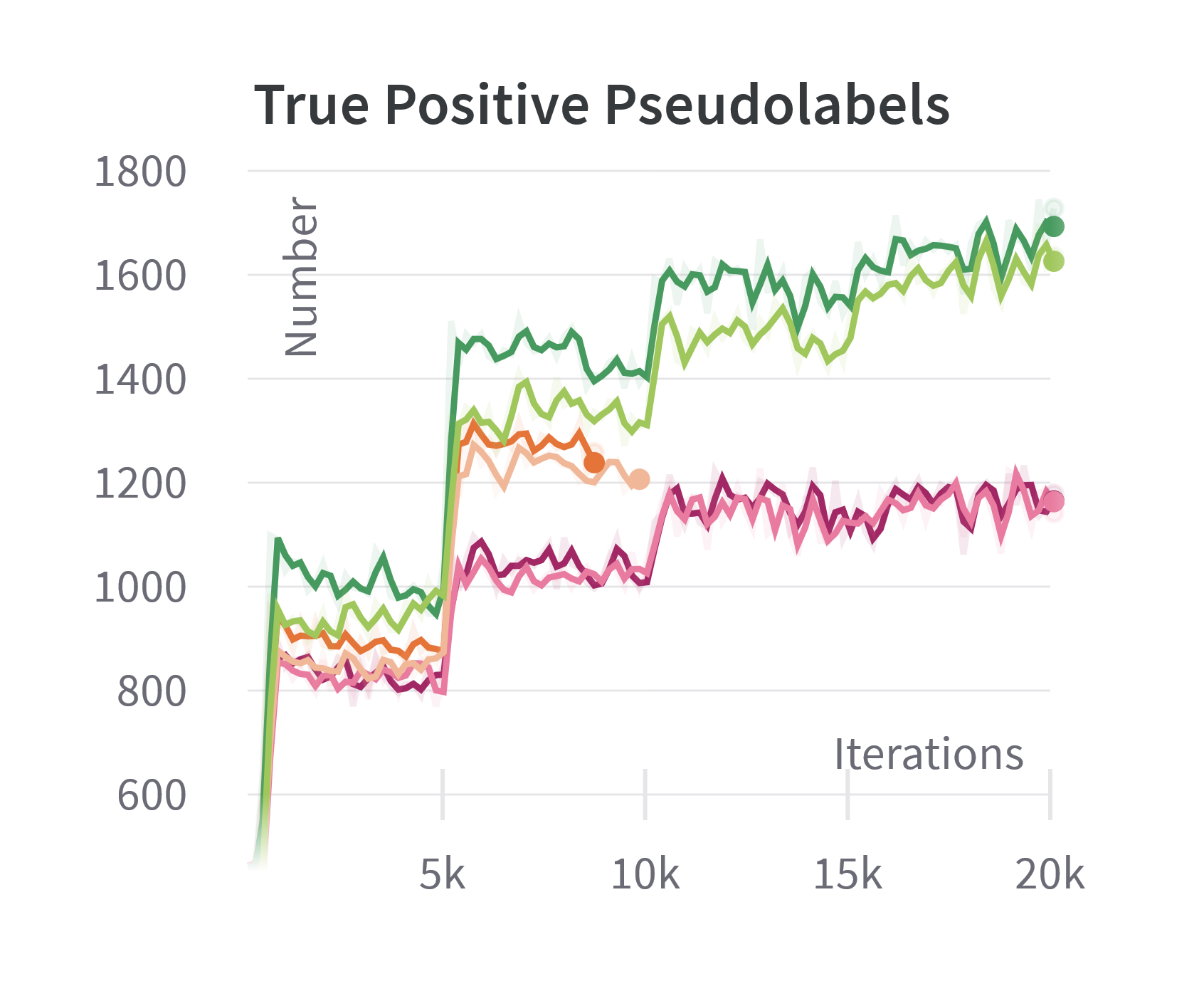}\label{fig:cluster_tpp}}
{\includegraphics[trim={110 110 110 110},clip, width=0.32\textwidth]{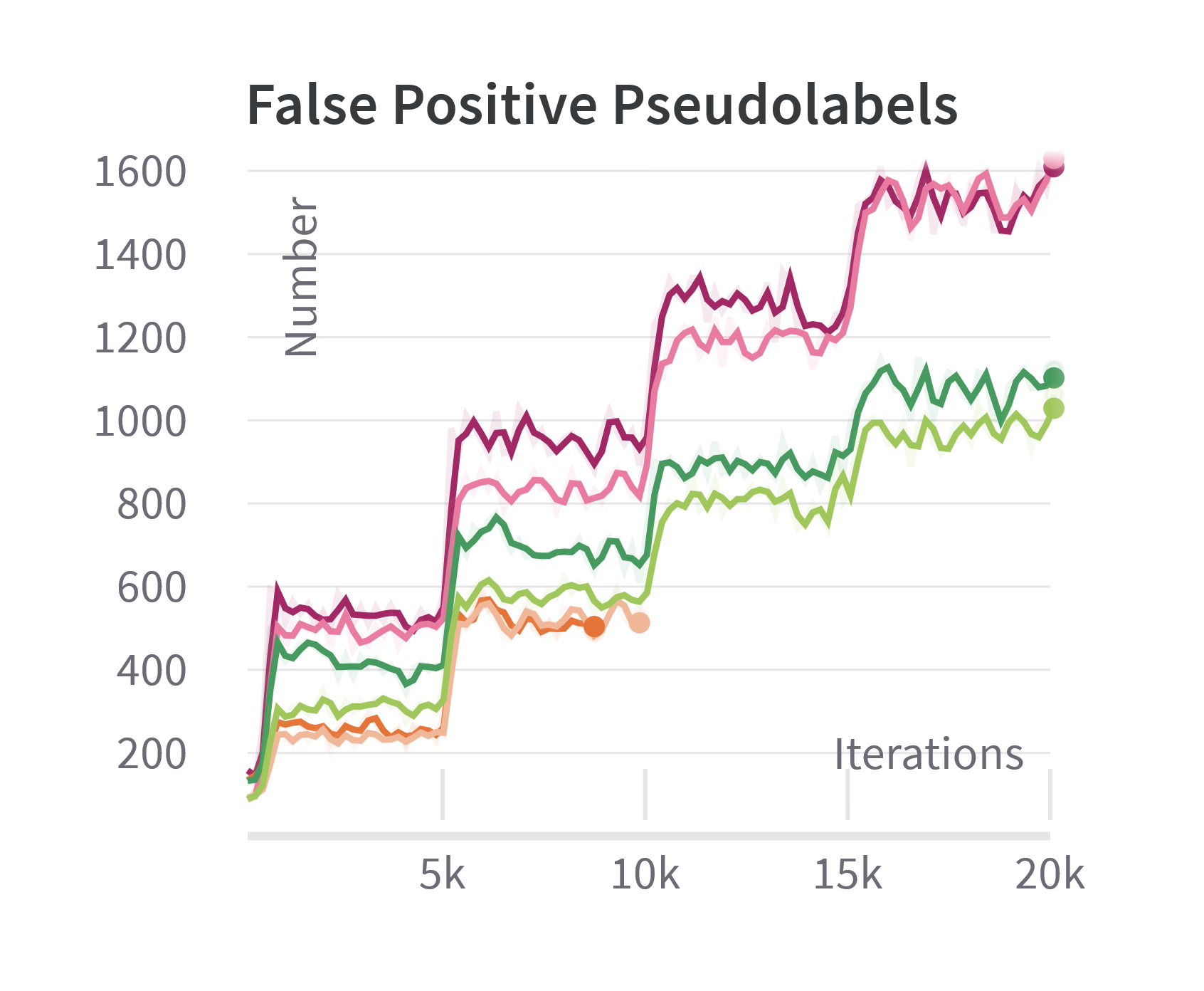}\label{fig:cluster_fpp}}
{\includegraphics[trim={110 110 110 110},clip, width=0.32\textwidth]{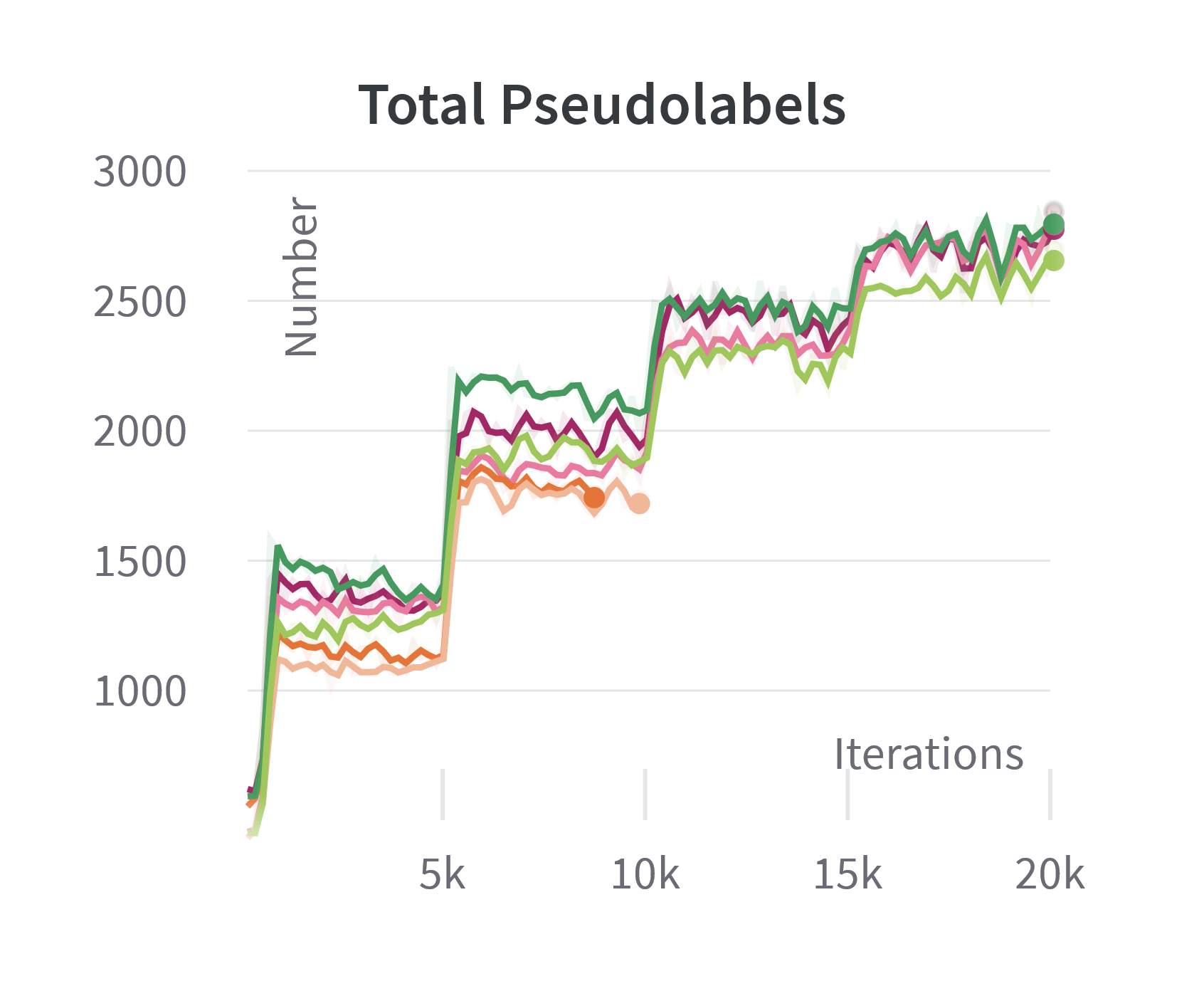}\label{fig:cluster_total}}
{\includegraphics[trim={110 110 110 110},clip, width=0.32\textwidth]{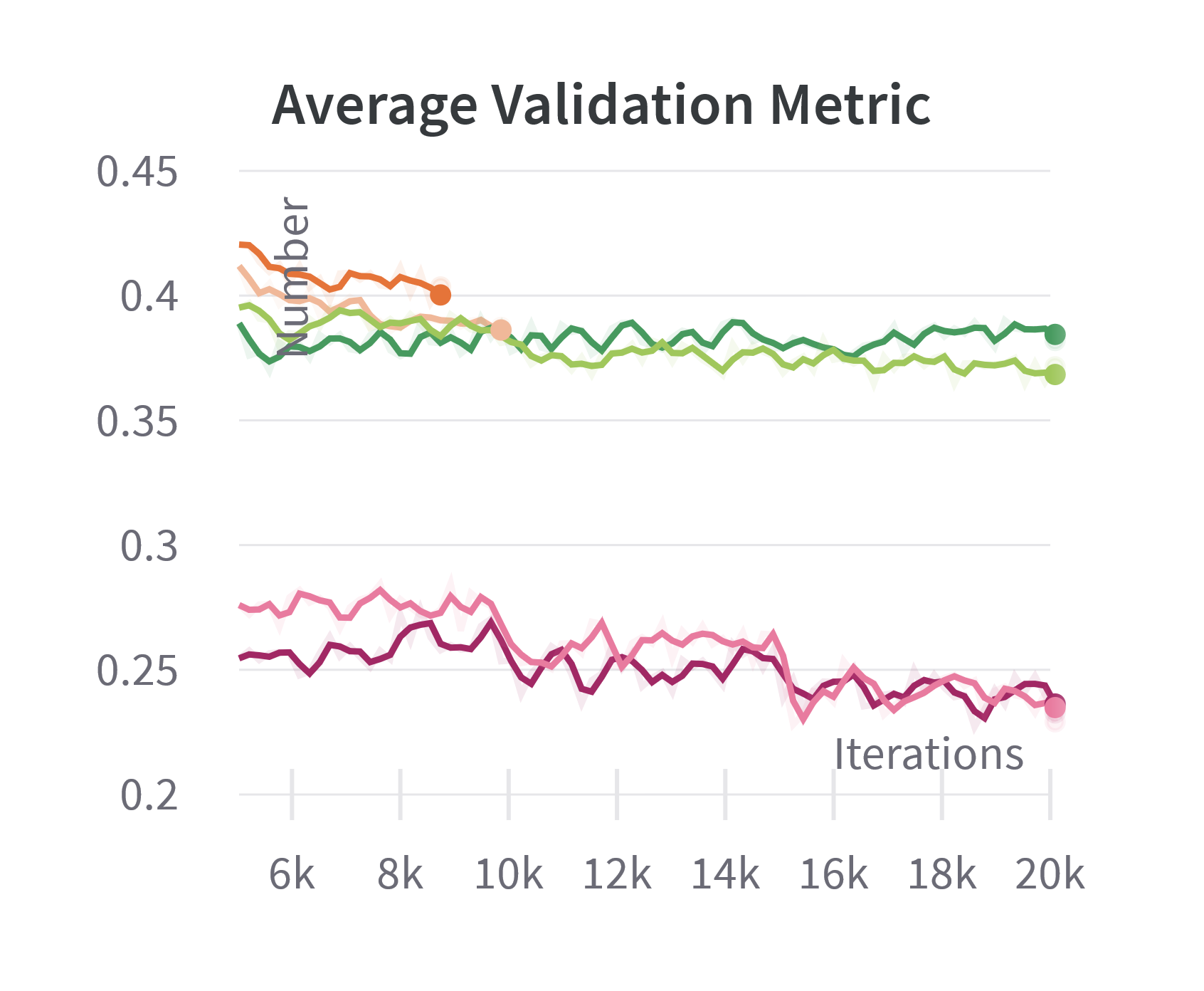}\label{fig:cluster_avm}}
{\includegraphics[trim={110 110 110 110},clip, width=0.32\textwidth]{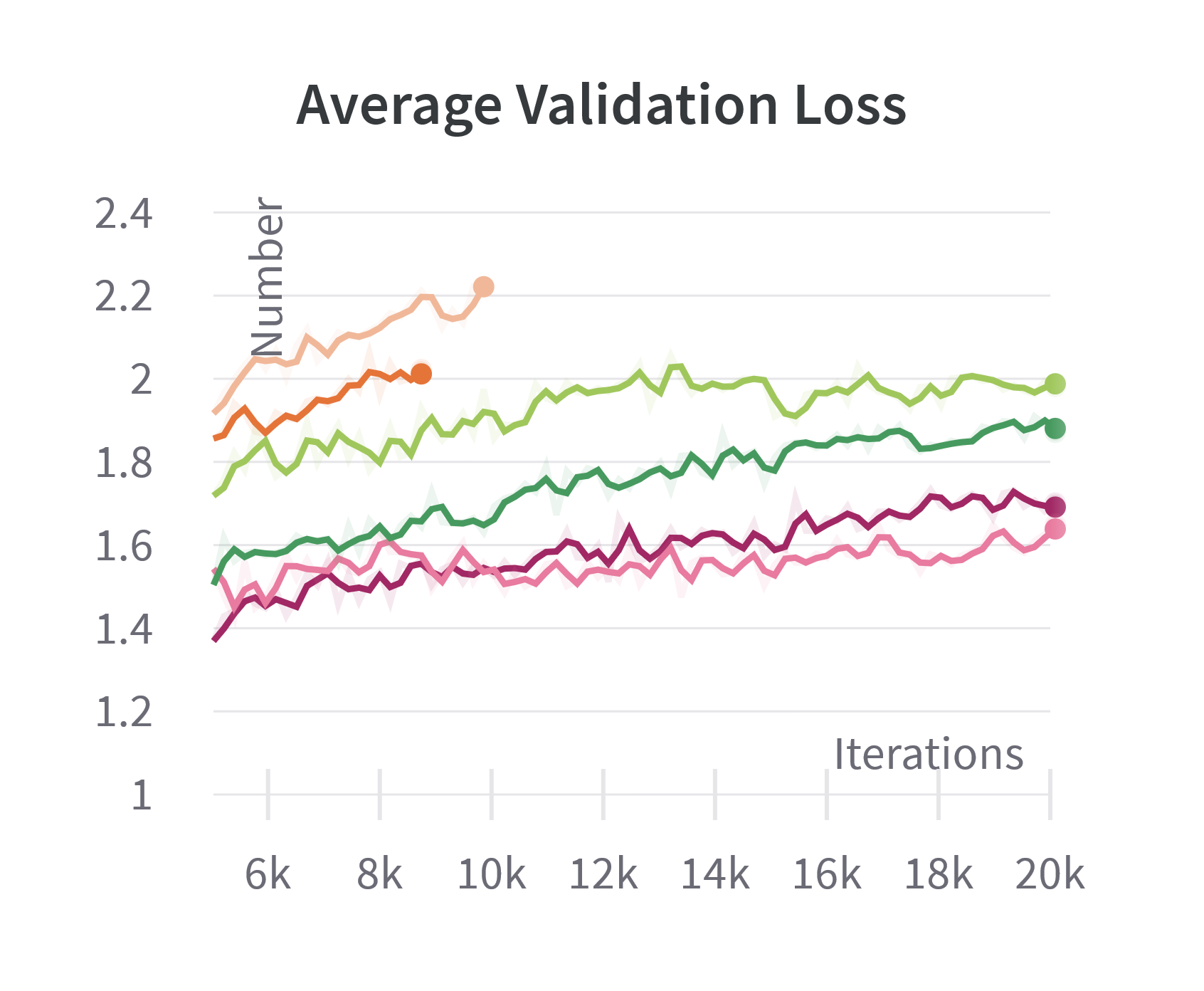}\label{fig:cluster_avl}}
{\includegraphics[trim={110 110 110 110},clip, width=0.32\textwidth]{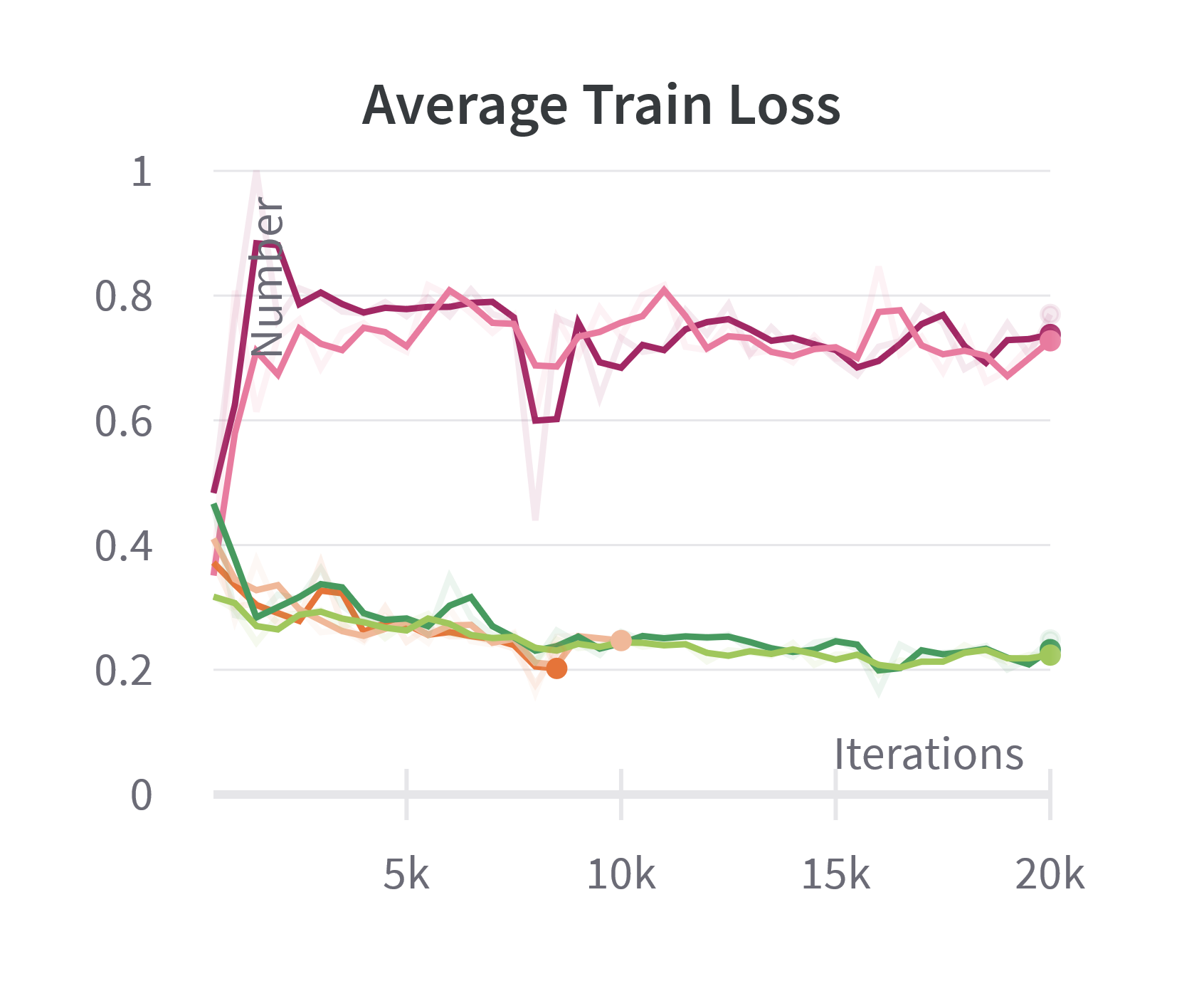}\label{fig:cluster_atl}}
{\includegraphics[trim={200 0 200 1400},clip, width=0.99\textwidth]{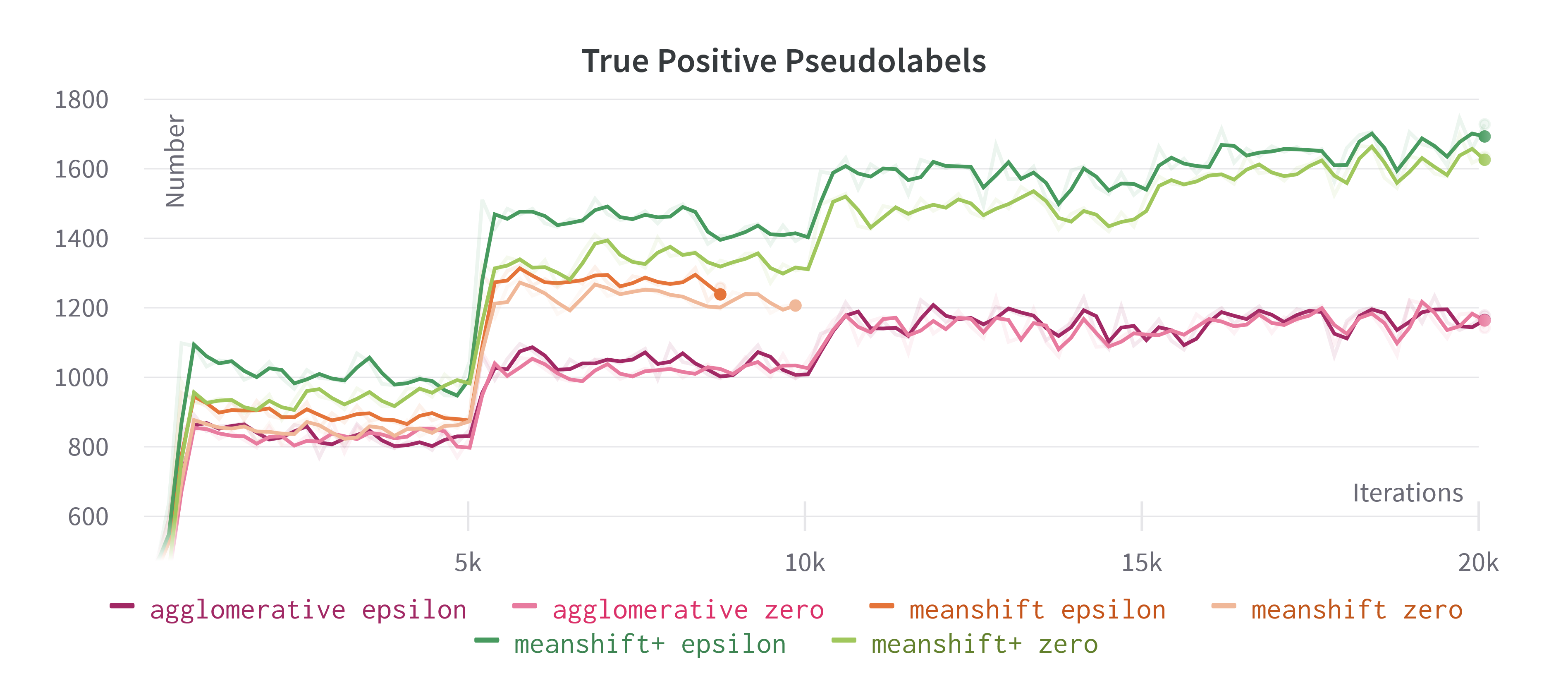}\label{fig:legend}}
\caption{Training metrics of mean-shift, mean-shift+, agglomerative clustering}
\label{fig:cluster}
\end{figure}

The clustering algorithms for pseudo-label generation are compared in an initial experiment during sef-training of the instance segmentation model. By default the $\delta$ distance values of cluster rings in equations \ref{eq:pull} and \ref{eq:push} are fixed during baseline and self-training. However to prevent early oversplitting of the pseudo-labels, a new $\epsilon$ distance value is introduced to further control the size of the push and pull rings by adding $\epsilon$. The results are shown in figure \ref{fig:cluster} where each algorithm is tested without $\epsilon$ i.e. $\epsilon = 0$ "zero", and an exponential decay of $\epsilon \in \{-0.2,0.2\}$ "epsilon".

\begin{figure}[h]
\centering
\captionsetup[subfloat]{labelformat=empty}
\rotatebox[origin=lc]{90}{Mean-shift}\hfill
\subfloat{\includegraphics[trim={0 0 70 0},clip, width=0.323\textwidth]{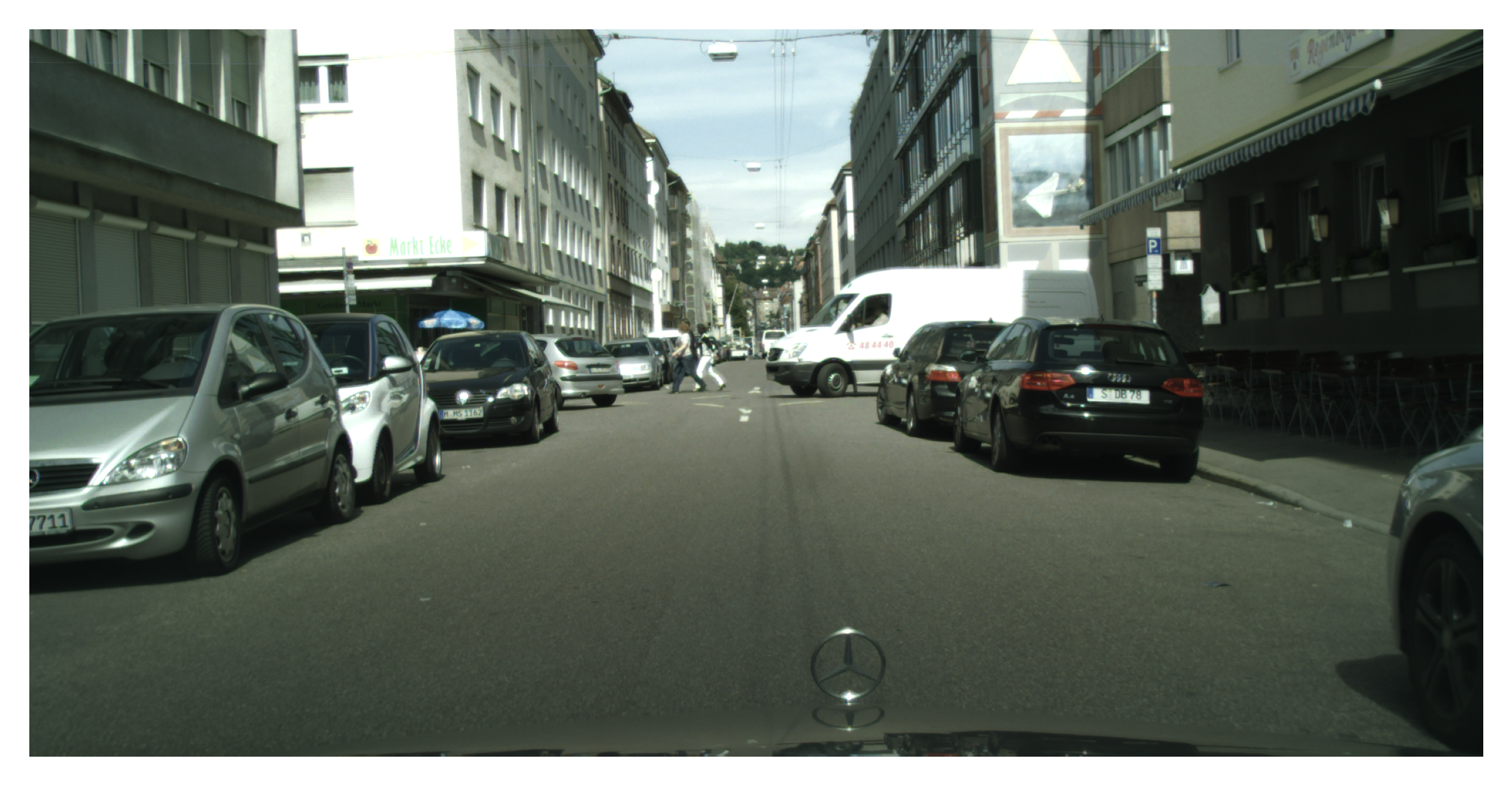}\label{fig:msi}}
\hfill
\subfloat{\includegraphics[trim={0 0 70 0},clip, width=0.323\textwidth]{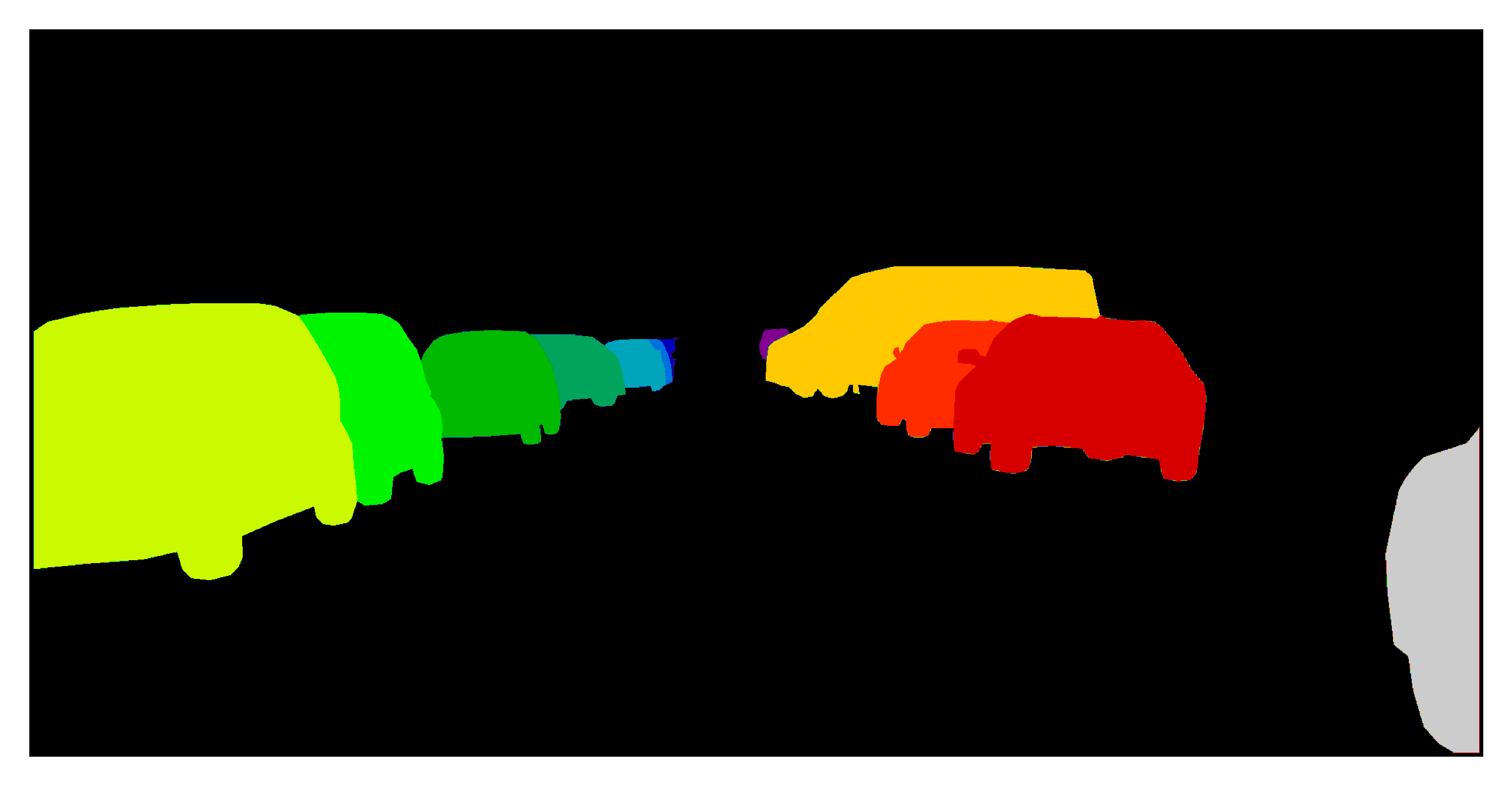}\label{fig:msg}}
\hfill
\subfloat{\includegraphics[trim={0 0 70 0},clip, width=0.323\textwidth]{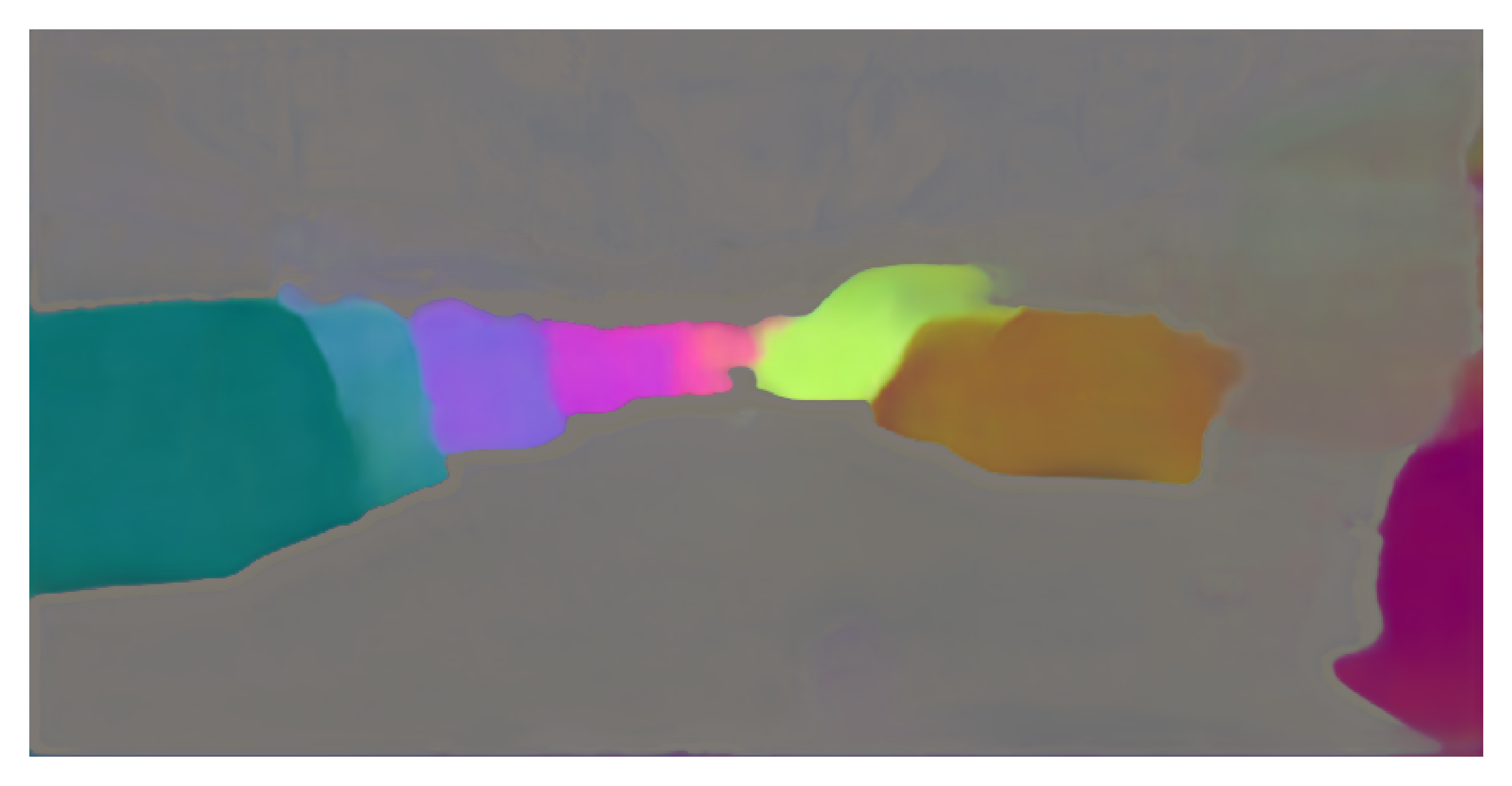}\label{fig:msc}}
\hfill
\rotatebox[origin=lc]{90}{Mean-shift+}\hfill
\subfloat{\includegraphics[trim={0 0 70 0},clip, width=0.32\textwidth]{images/input.png}\label{fig:mspi}}
\hfill
\subfloat{\includegraphics[trim={0 0 70 0},clip, width=0.32\textwidth]{images/gt.png}\label{fig:mspg}}
\hfill
\subfloat{\includegraphics[trim={0 0 70 0},clip, width=0.32\textwidth]{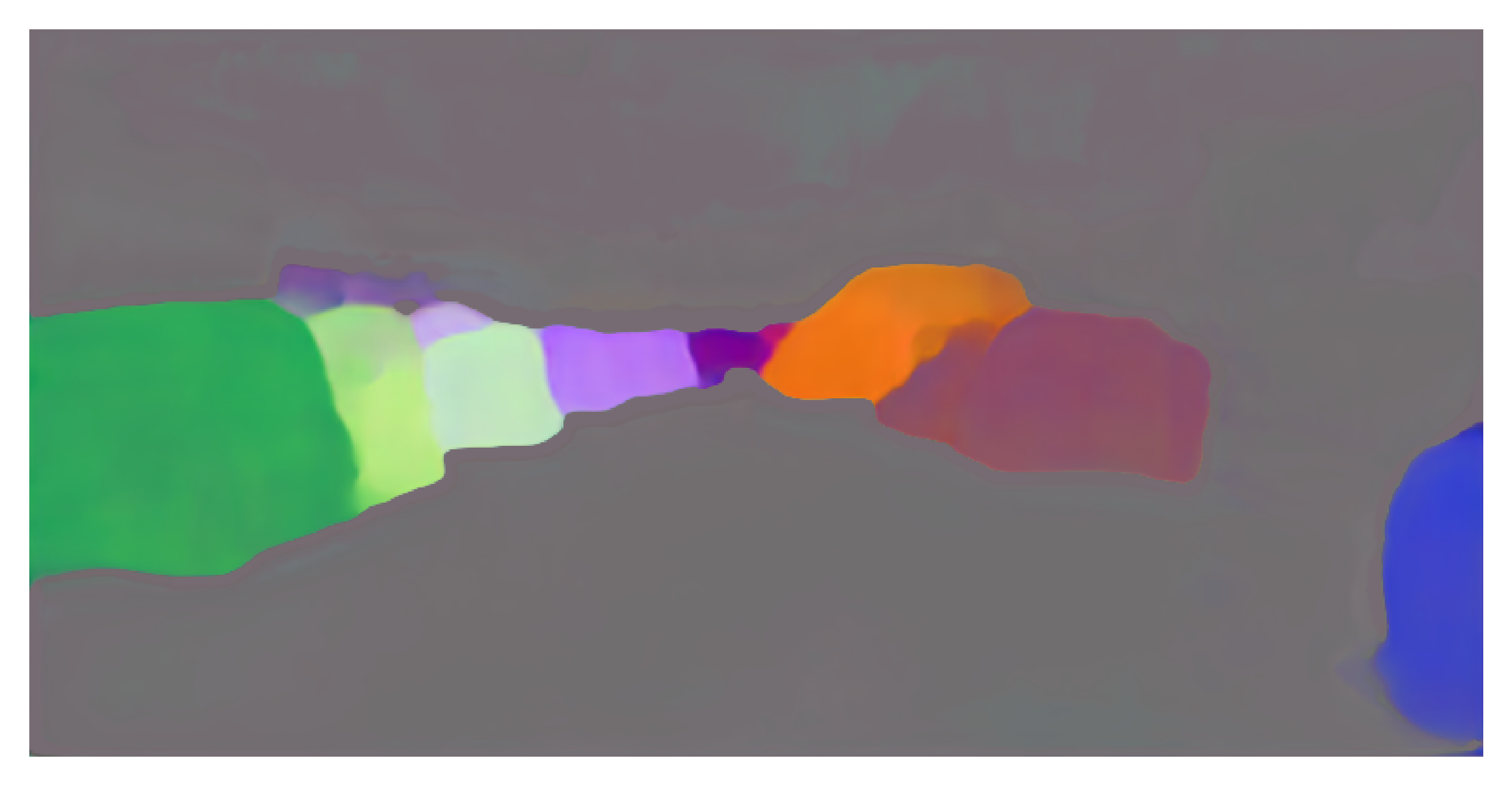}\label{fig:mspc}}
\hfill
\rotatebox[origin=lc]{90}{Agglomerative}\hfill
\subfloat[Input]{\includegraphics[trim={0 0 70 0},clip, width=0.32\textwidth]{images/input.png}\label{fig:aggi}}
\hfill
\subfloat[Ground-truth]{\includegraphics[trim={0 0 70 0},clip, width=0.32\textwidth]{images/gt.png}\label{fig:aggg}}
\hfill
\subfloat[Embedding]{\includegraphics[trim={0 0 70 0},clip, width=0.32\textwidth]{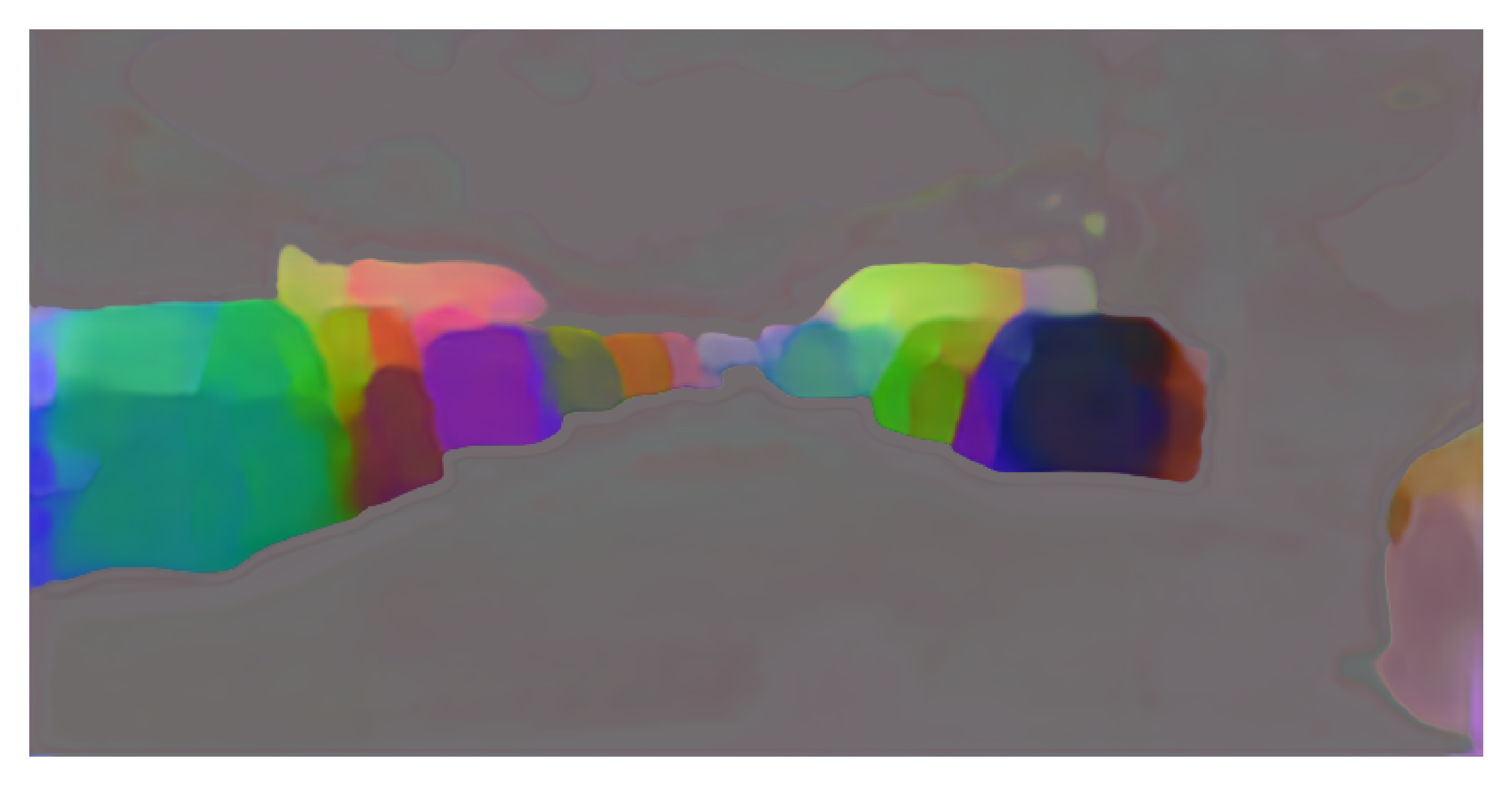}\label{fig:aggc}}
\caption{Embeddings of mean-shift, mean-shift+, agglomerative clustering}
\label{fig:cluster_images}
\end{figure}

The regular mean-shift clustering is extremely slow and took approximately 96 hours to reach 10k iterations while the mean-shift+ and agglomerative clustering reached 10k iterations in around 20 hours, as seen in figure \ref{fig:cluster}. In figure \ref{fig:cluster}, the true positive (TP) and false positive (FP) pseudo-labels, and the average validation metric is calculated using instance segmentation ground-truth, however it does not affect the model in any way. The average validation metric is calculated using \textbf{Average Precision} (AP) which is defined as $AP = TP / (TP + FP)$. It is also observed that number of false positive pseudo-labels are constantly increasing for all methods while number of true positive pseudo-labels start to plateau after 10K iterations. PCA projected cluster embeddings shown in figure \ref{fig:cluster_images} are from the validation set of the final iteration. The embeddings of agglomerative clustering method show oversplitting of the instances caused due to overfitting of the model, which is also why the average training loss metric rapidly rose during first few iterations. Since the performance of mean-shift+ is significantly better compared to agglomerative clustering and the run-time is quicker than mean-shift clustering, it is picked as the default clustering algorithm with exponential decay of $\epsilon \in \{-0.2,0.2\}$. The problem of rapidly rising false positives still persists in mean-shift+ clustering, and the causes of high false positives and increasing validation loss could be due to "catastrophic forgetting". Another reason could be the variance in the shapes and sizes of instance masks between the source and target domains, along with the variance of object masks for different classes within the datasets. The next sections will introduce ideas to tackle that problem.

\section{Category and Class-wise Models}

\begin{figure}[h]
\centering
\includegraphics[width=0.70\textwidth]{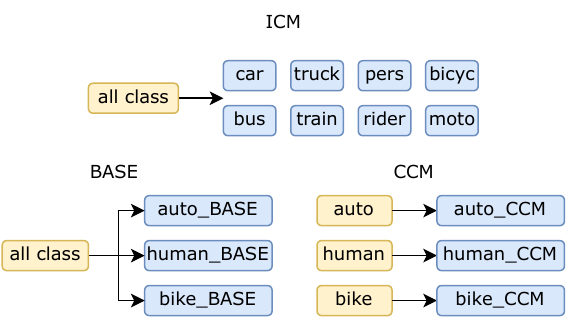}
\caption{Comparison of all ICM, BASE and CCM training workflow}
\label{fig:ccm_base}
\end{figure}

Initially, the instance baseline and self-train models are trained using all available instance objects from the Cityscapes dataset which included classes: \{\textit{car, bus, truck, train, person, rider, bicycle, and motorcycle}\}. This approach led to an instance model that is excessively complex and computationally intensive, demanding significant training time and computational resources. To address this challenge, three different training workflows are explored: \textbf{Individual-Class Models} (ICM) where the baseline model is trained on all classes and self-train models are split into individual eight classes; \textbf{Baseline-Category Models} (BASE) where the baseline model is trained on all classes and self-train models are partitioned into three distinct groups based on broad class categories; and \textbf{Category-Centric Models} (CCM) where both baseline and self-train models are split into three categories. Specifically, auto-model is trained on \{\textit{car, bus, truck,  train}\} classes; human-model on \{\textit{person, rider}\} classes; and bike-model on \{\textit{bicycle, motorcycle}\} classes. These training workflows are better visualized in figure \ref{fig:ccm_base} where yellow are baseline models and blue are self-train models. For comparison, the initial setup with "{all\_classes}" baseline and self-train model is also included.

\begin{figure}[H]
\centering
{\includegraphics[trim={110 110 110 110},clip, width=0.32\textwidth]{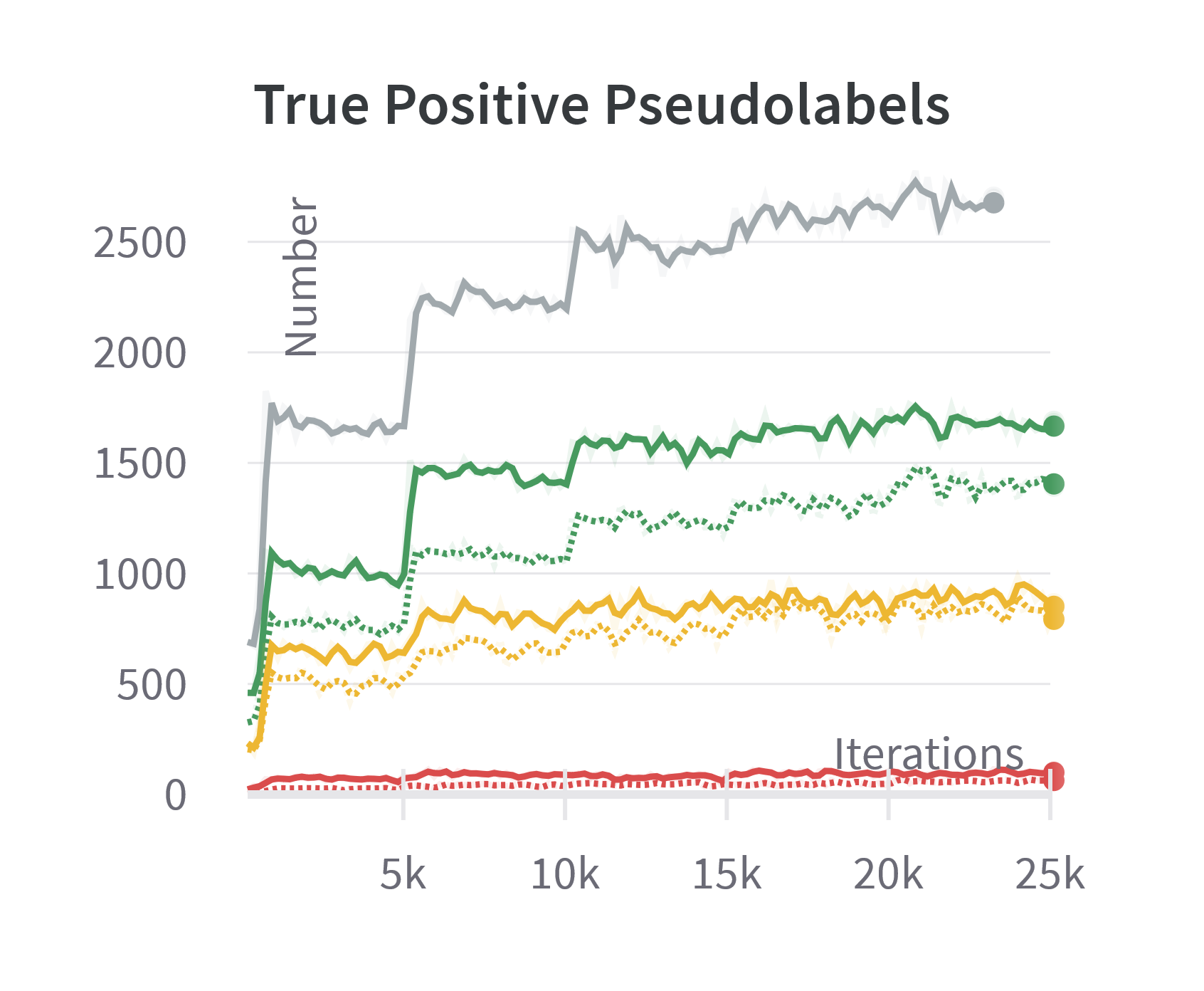}\label{fig:ccm_tpp}}
{\includegraphics[trim={110 110 110 110},clip, width=0.32\textwidth]{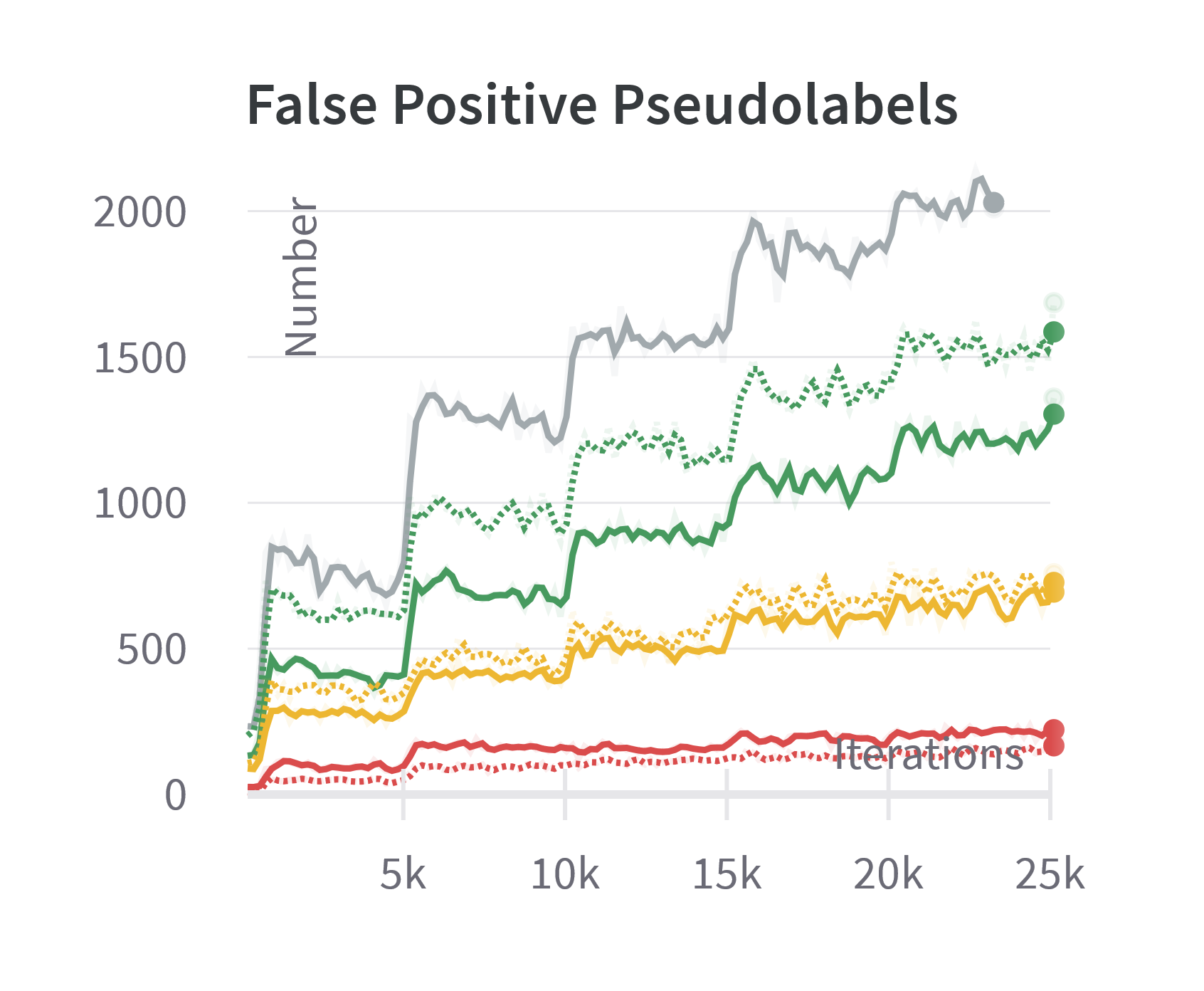}\label{fig:ccm_fpp}}
{\includegraphics[trim={110 110 110 110},clip, width=0.32\textwidth]{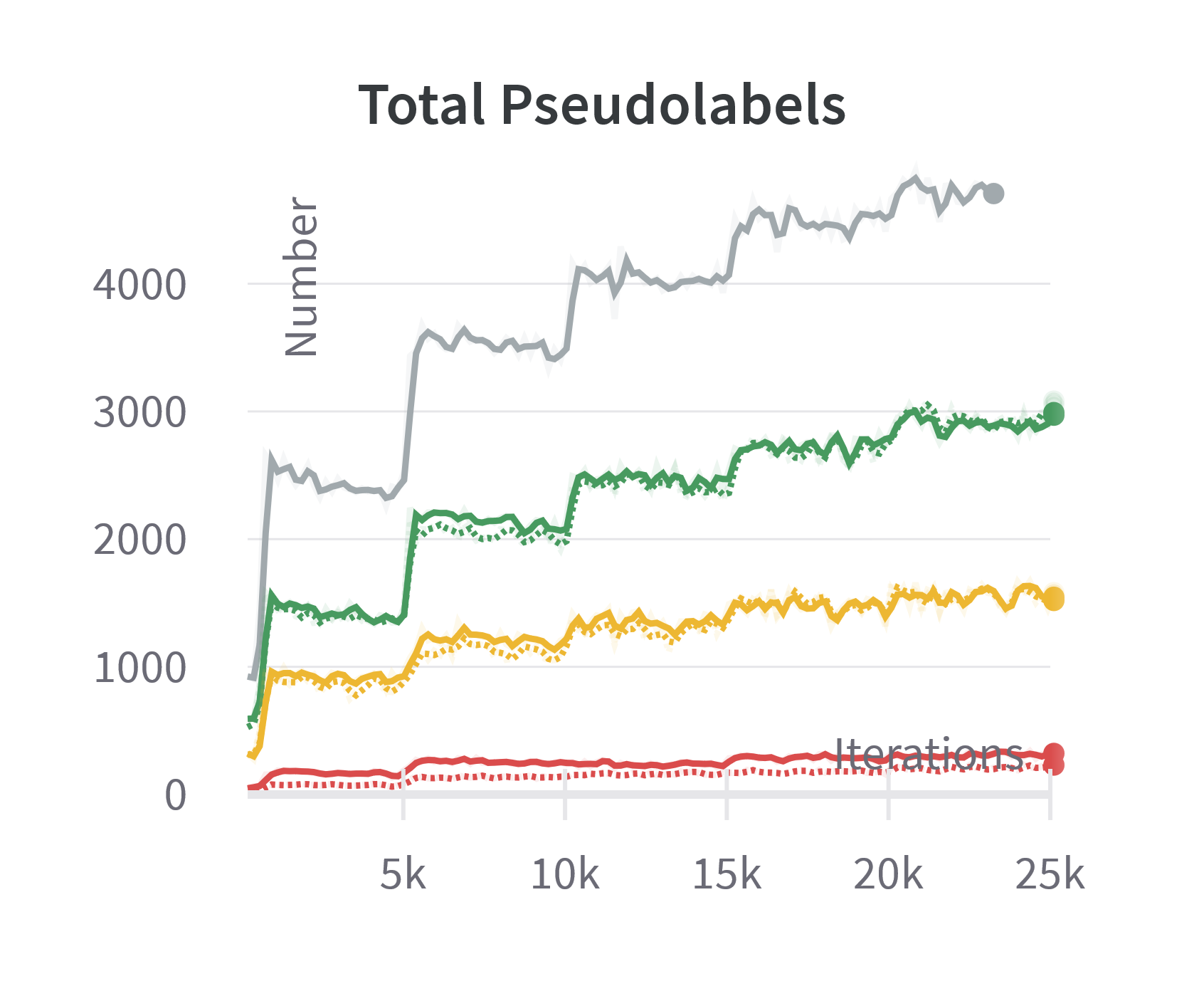}\label{fig:ccm_total}}
{\includegraphics[trim={110 110 110 110},clip, width=0.32\textwidth]{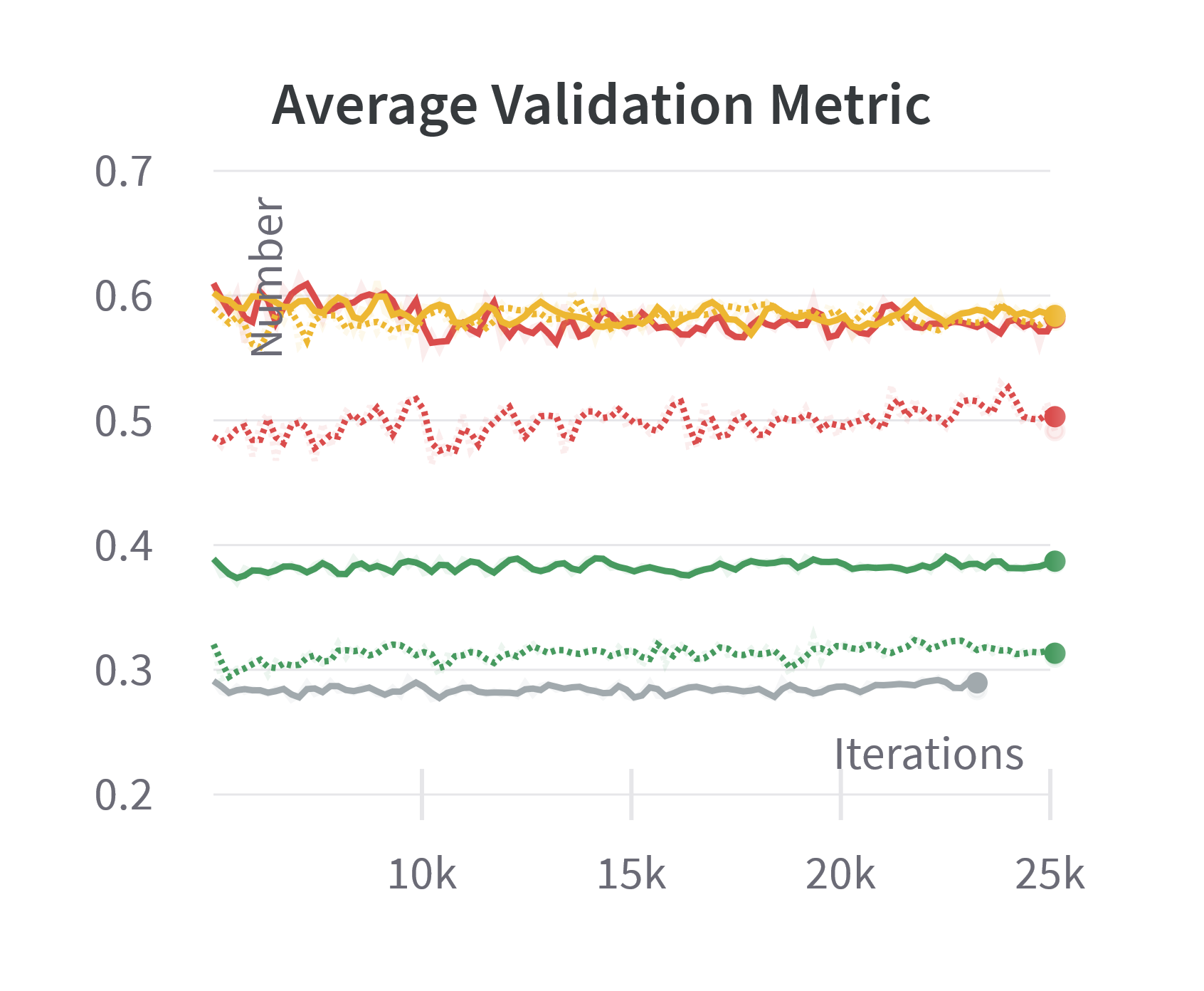}\label{fig:ccm_avm}}
{\includegraphics[trim={110 110 110 110},clip, width=0.32\textwidth]{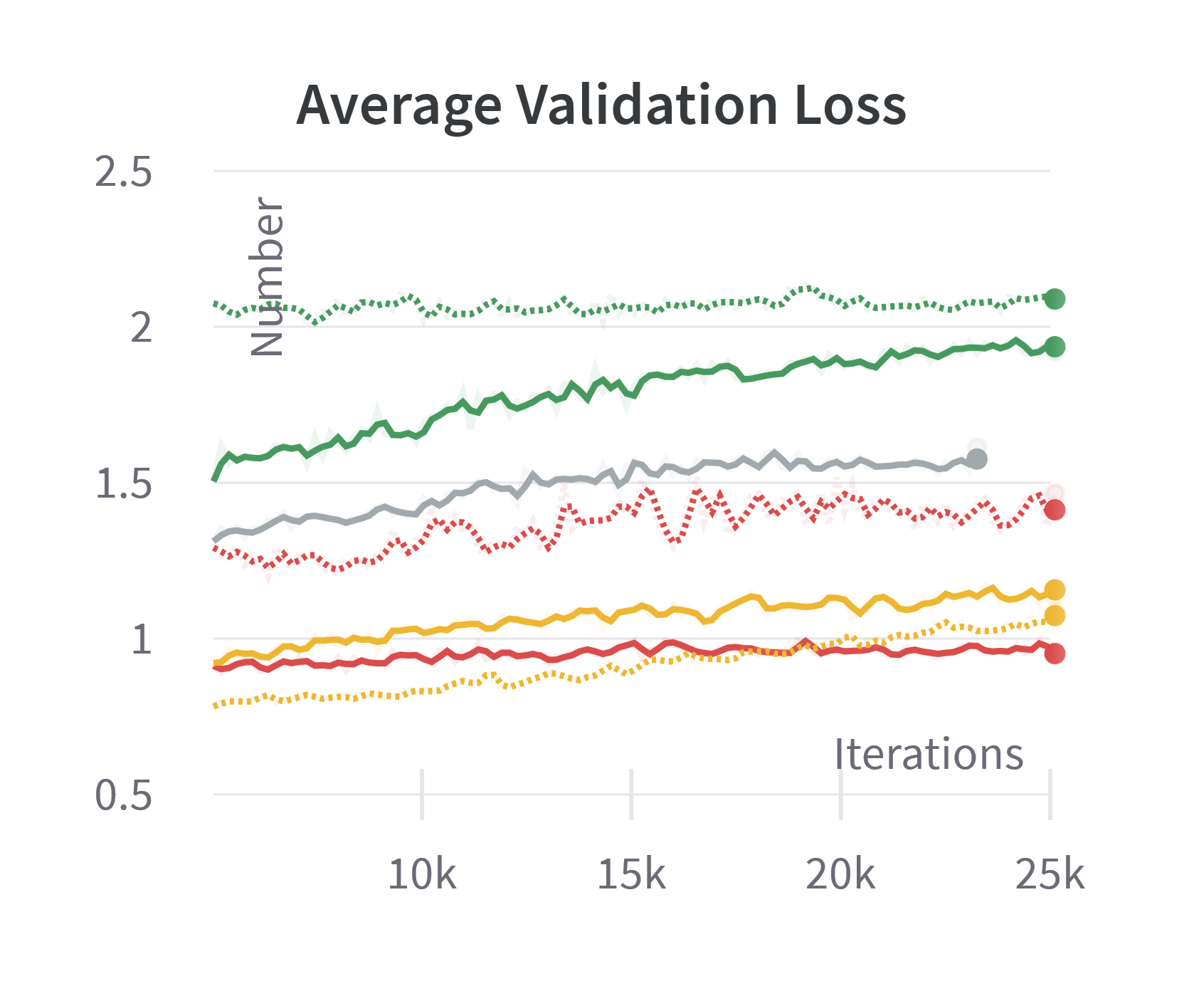}\label{fig:ccm_avl}}
{\includegraphics[trim={110 110 110 110},clip, width=0.32\textwidth]{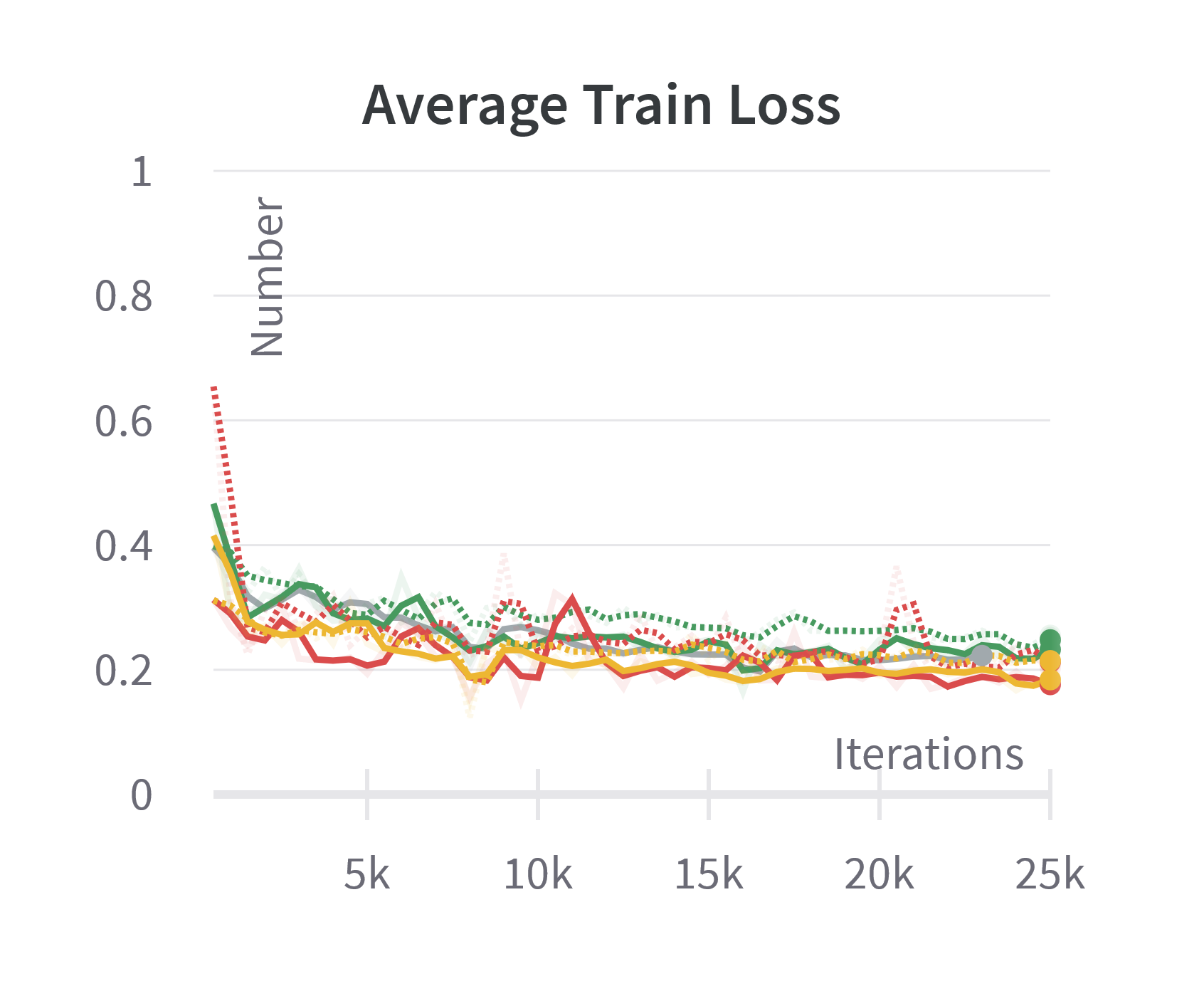}\label{fig:ccm_atl}}
{\includegraphics[trim={200 0 200 1700},clip, width=0.85\textwidth]{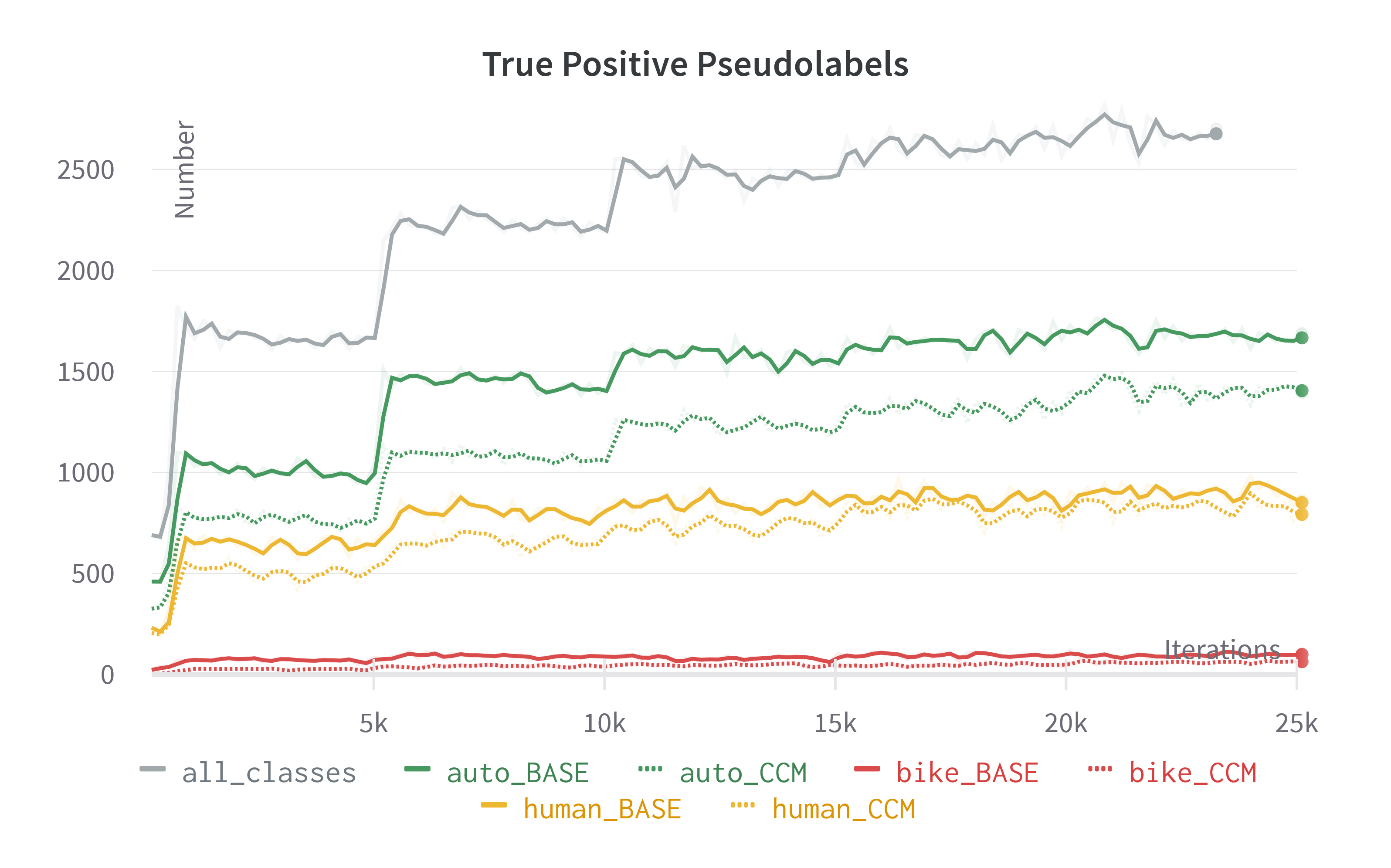}\label{fig:ccm_legend}}
\caption{Training metrics of BASE and CCM methods}
\label{fig:ccm}
\end{figure}

\begin{figure}[h]
\centering
{\includegraphics[trim={110 110 110 110},clip, width=0.32\textwidth]{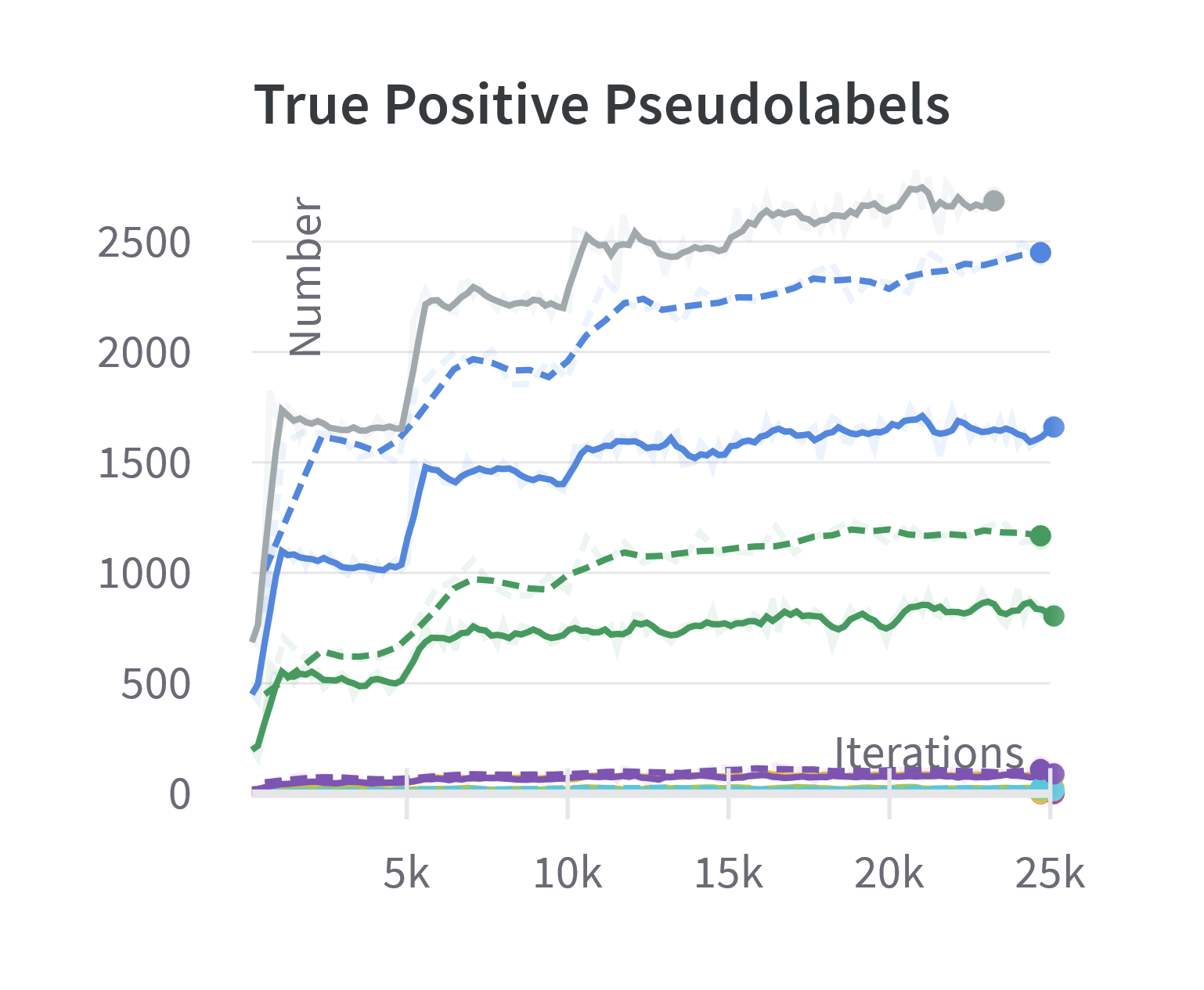}\label{fig:icm_tpp}}
{\includegraphics[trim={110 110 110 110},clip, width=0.32\textwidth]{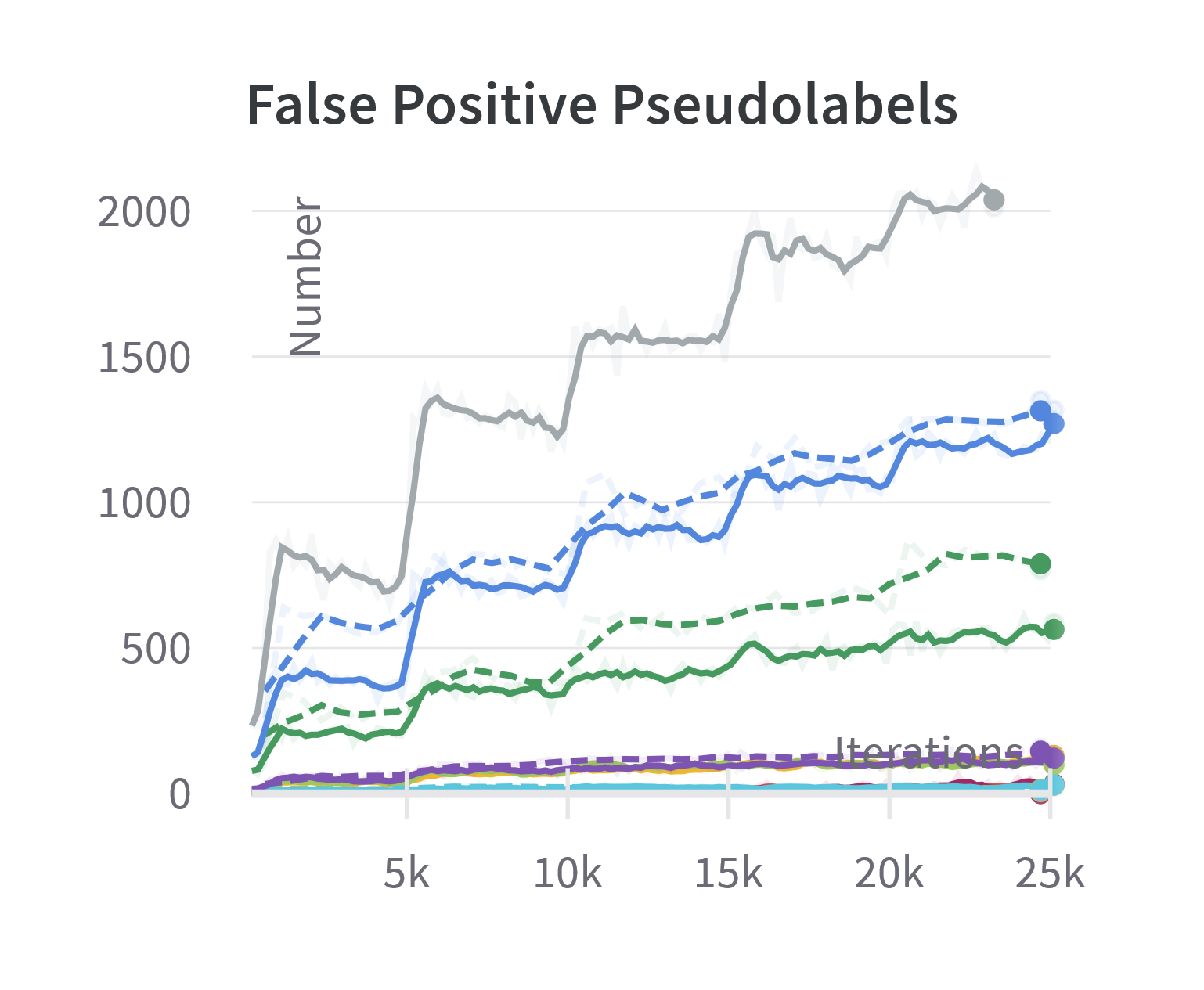}\label{fig:icm_fpp}}
{\includegraphics[trim={110 110 110 110},clip, width=0.32\textwidth]{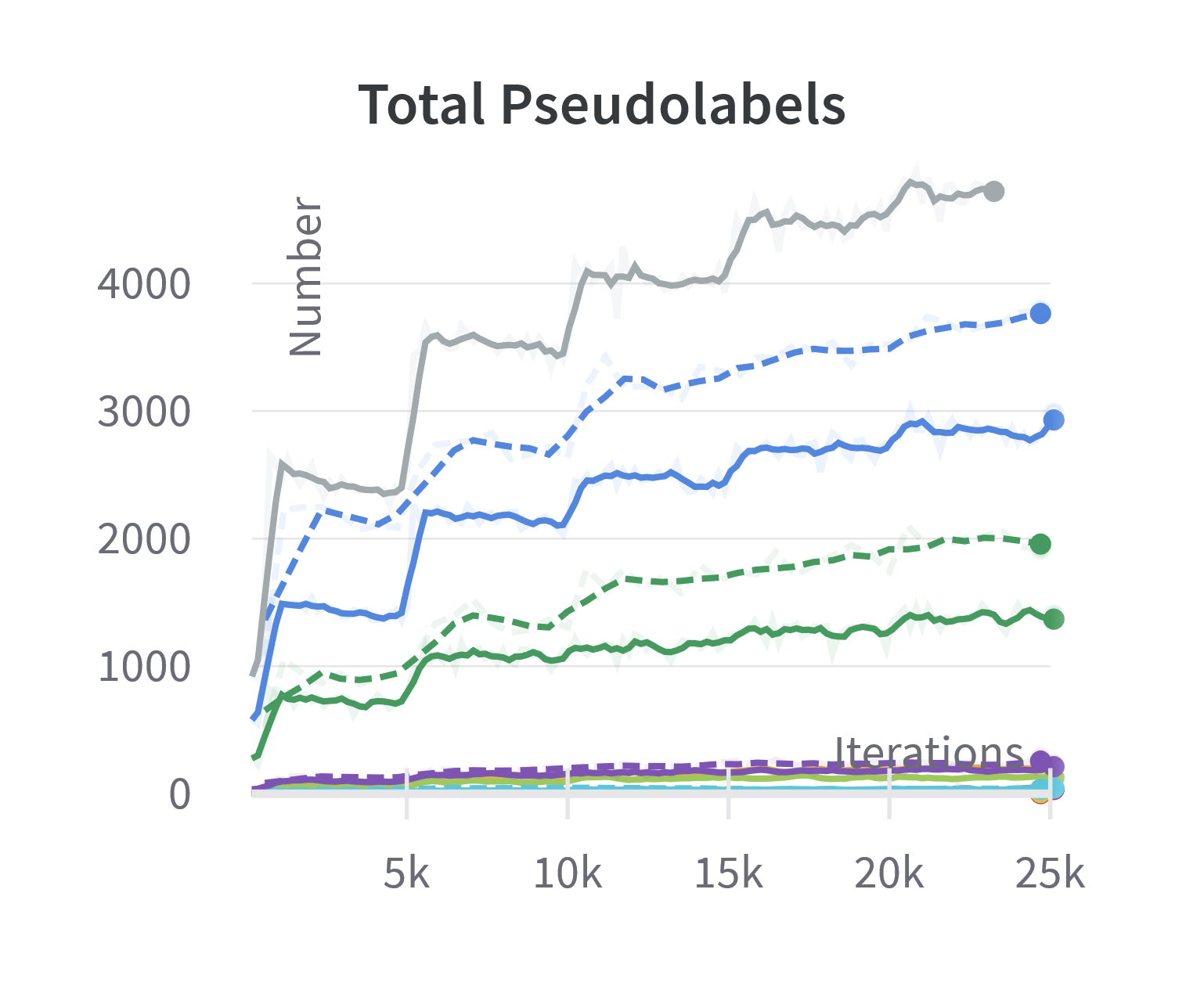}\label{fig:icm_total}}
{\includegraphics[trim={110 110 110 110},clip, width=0.32\textwidth]{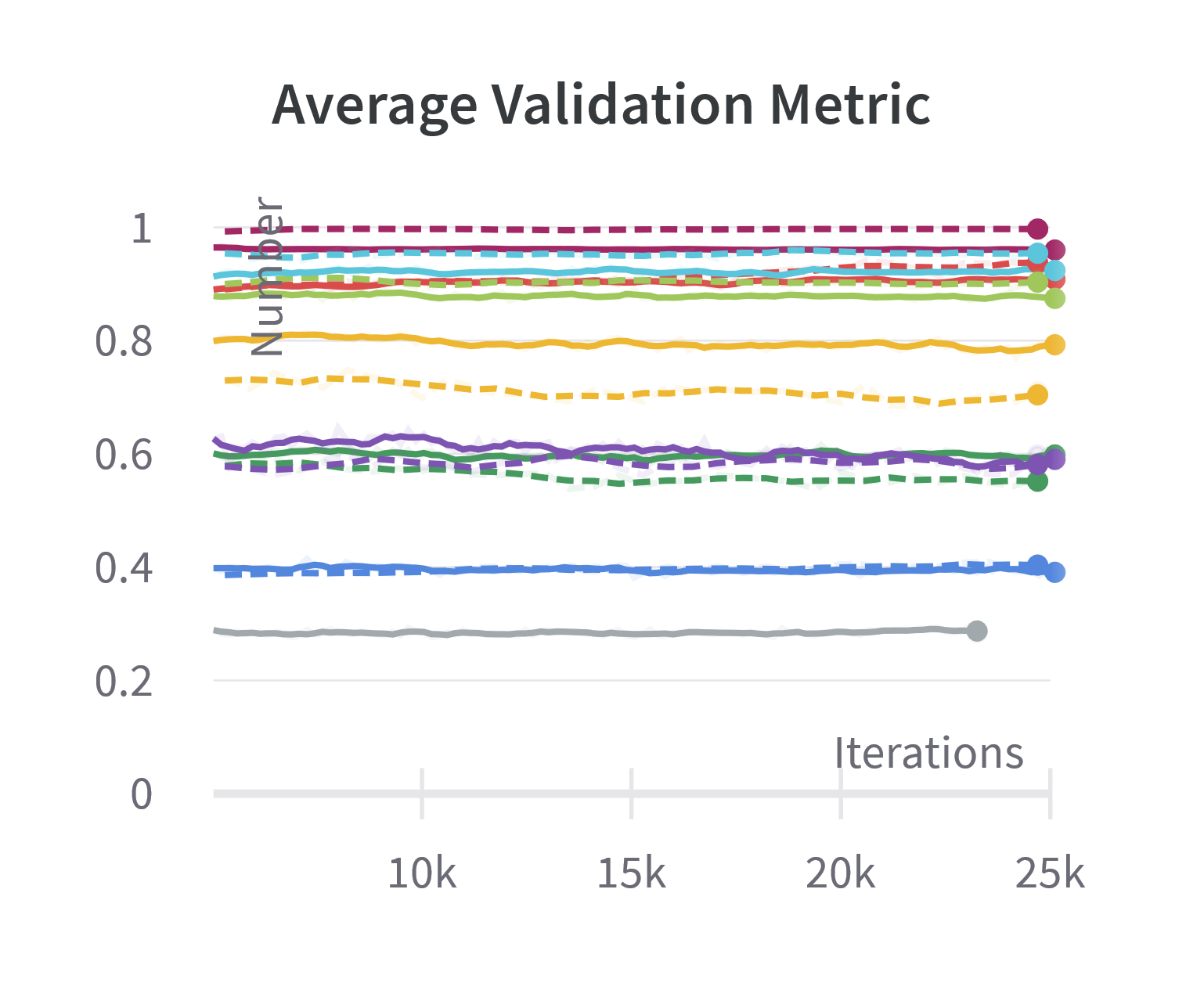}\label{fig:icm_avm}}
{\includegraphics[trim={110 110 110 110},clip, width=0.32\textwidth]{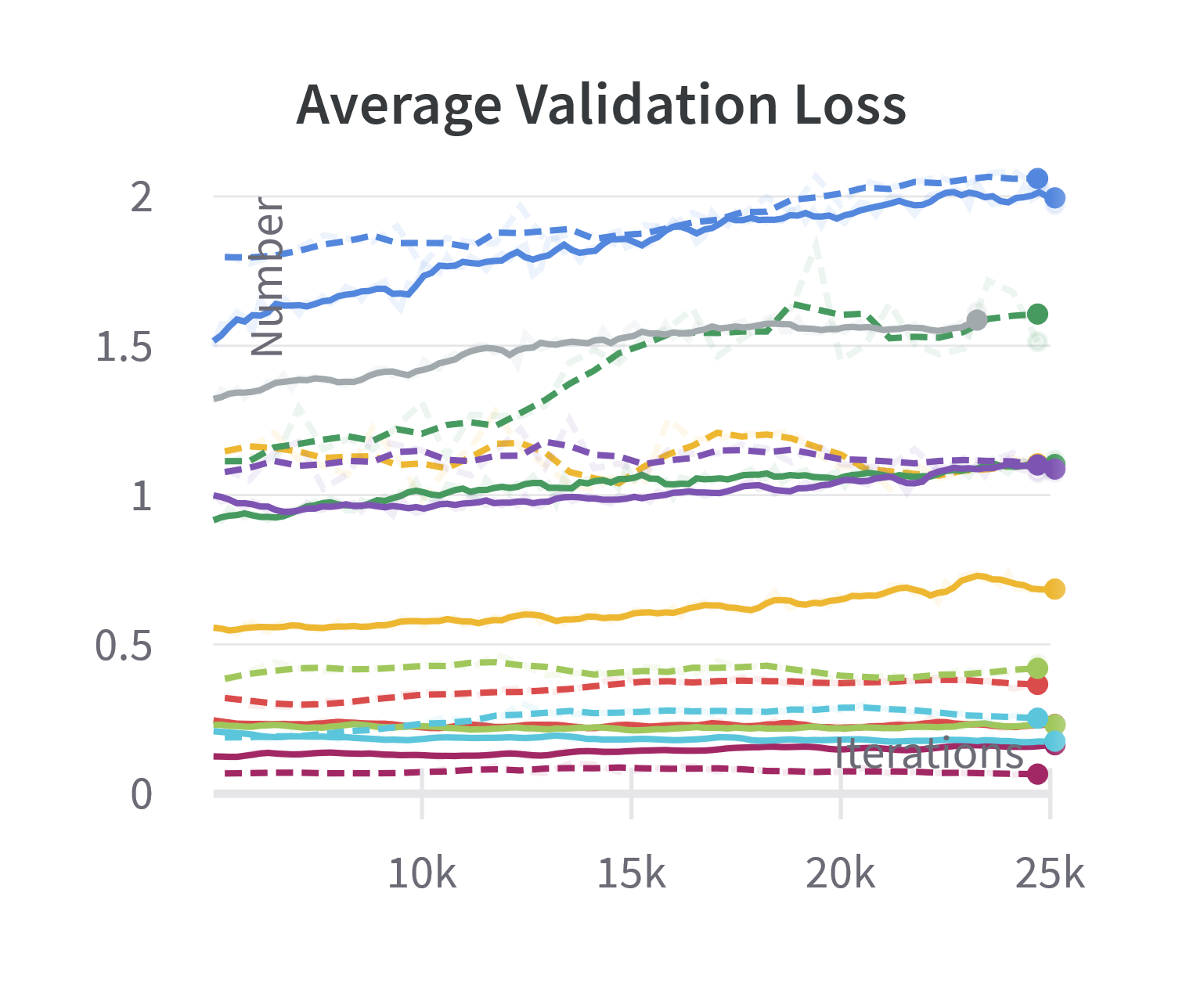}\label{fig:icm_avl}}
{\includegraphics[trim={110 110 110 110},clip, width=0.32\textwidth]{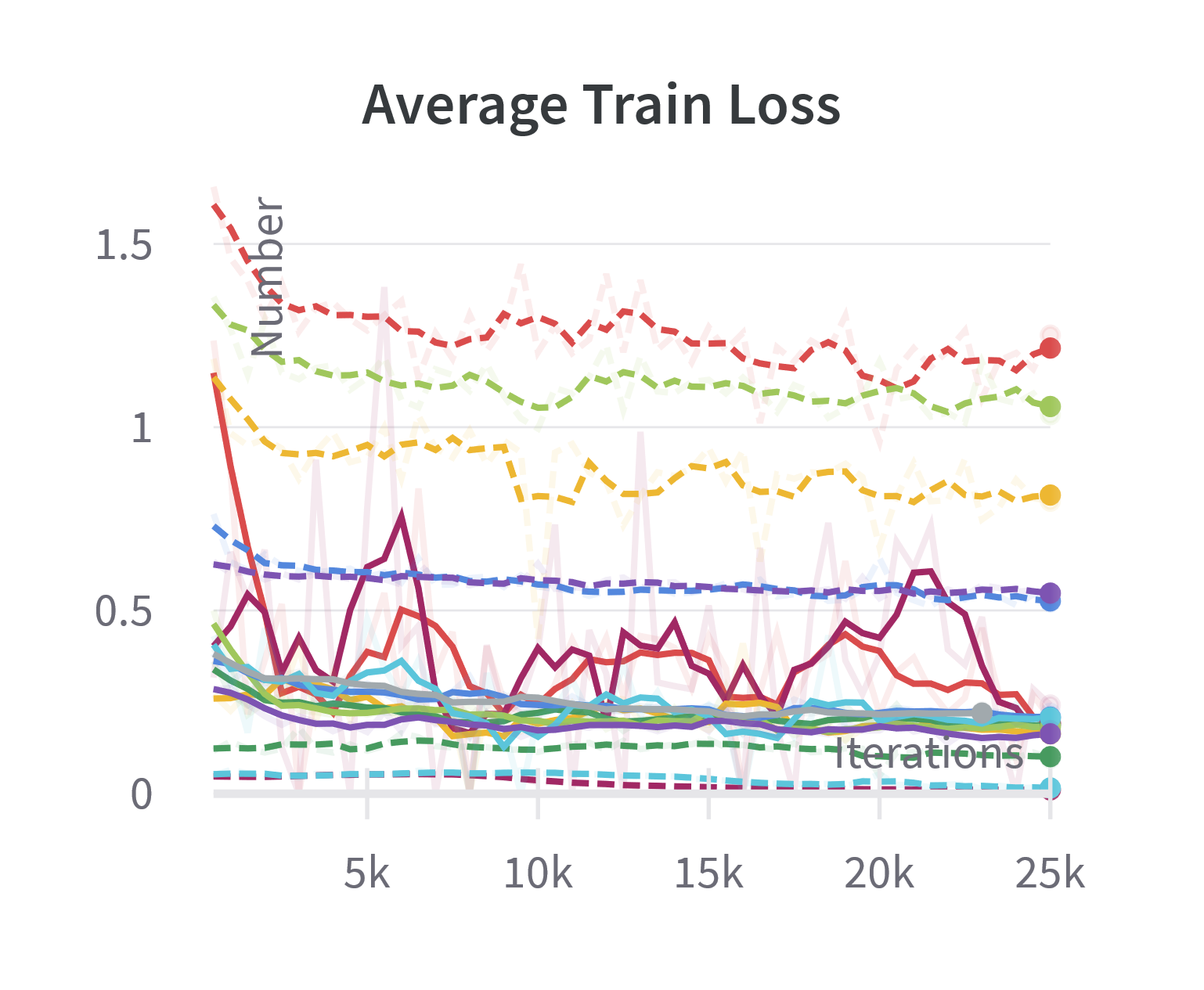}\label{fig:icm_atl}}
{\includegraphics[trim={150 0 150 1400},clip, width=0.99\textwidth]{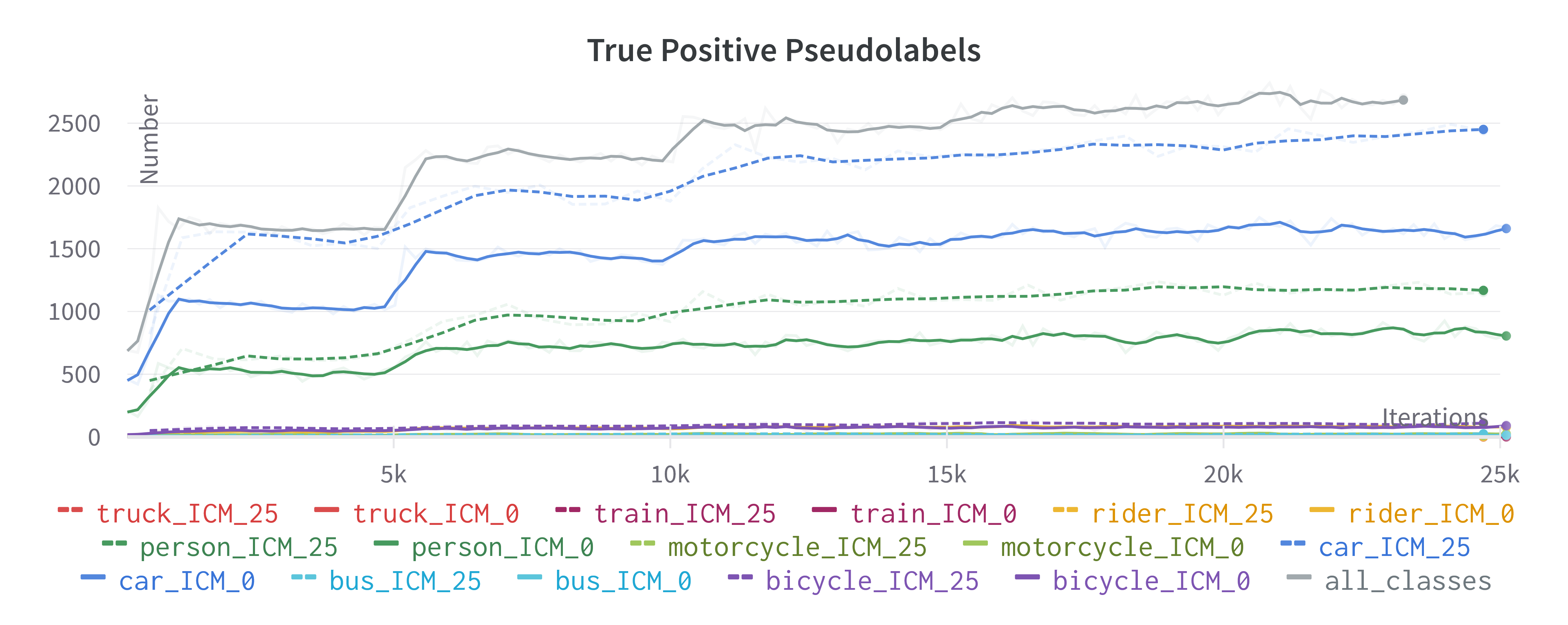}\label{fig:icm_legend}}
\caption{Training metrics ICM method for different levels of source data mixing}
\label{fig:icm}
\end{figure}

As seen in figure \ref{fig:ccm}, the BASE method outperformed the CCM method during self-training. The BASE models had higher true positive pseudo-labels and gave higher average validation metric score, compared to CCM for all three categories. Additionally, both BASE and CCM models outperformed the initial {all\_classes} self-train model. However, the problem of rising false positive pseudo-labels and average validation loss still persisted in all cases. Figure \ref{fig:icm} shows training metrics for the ICM method where each class is individually trained. In the figure the solid lines refer to runs where no source data is mixed during self-training, while dotted lines refer to runs where 25\% of source data is mixed during self-training to reduce catastrophic forgetting, which is discussed in detail in the next section. It can be seen in the figure that the average validation metric score for individual classes is higher compared to the initial {all\_classes} method.

These three approaches provided substantial advantages in terms of efficiency, since the ability to train these smaller specialized models in parallel significantly reduced training time. The \textit{all\_classes} self-train model took approx. 48 hours to train 25k iterations, while the BASE and CCM methods took approx. 24 hours to train, and the ICM method took approx. 12-14 hours. They also offered adaptability to evolving datasets or tasks. In the event of dataset modifications or new tasks, it is more straightforward to adapt or retrain individual models for specific categories, as opposed to retraining a single model from scratch. The next section contains more comparisons of BASE and CCM methods with the aim to reduce catastrophic forgetting.

\section{Reducing Catastrophic Forgetting}

In deep learning, catastrophic forgetting is a notable challenge and occurs when a model forgets previously learned information while acquiring new knowledge \cite{kirkpatrick2017overcoming} \cite{french1999catastrophic}. This issue emerges during fine-tuning model with new data because adapting to the new data distribution can lead to the erasure of previously learned patterns. This is often a result of gradient updates during back-propagation overwriting previously acquired weights, resulting in knowledge loss. Initially, the instance model used source dataset for baseline training and target data for self-training. However, this approach  could disrupt the representations learned in the baseline phase, causing the model to lose valuable features acquired earlier having a negative impact on self-training. 

\begin{figure}[H]
\centering
{\includegraphics[trim={110 110 110 110},clip, width=0.32\textwidth]{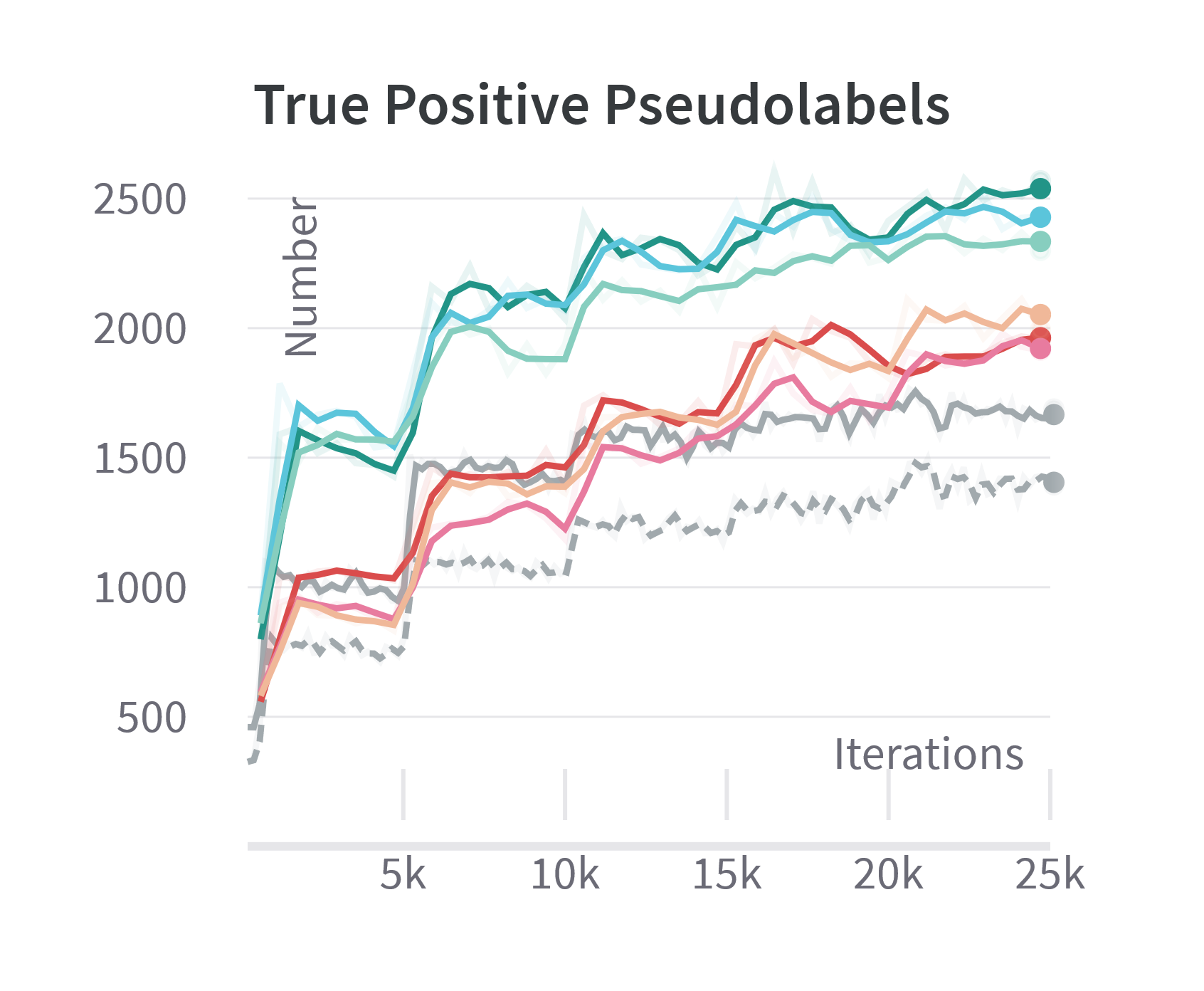}\label{fig:cata_tpp}}
{\includegraphics[trim={110 110 110 110},clip, width=0.32\textwidth]{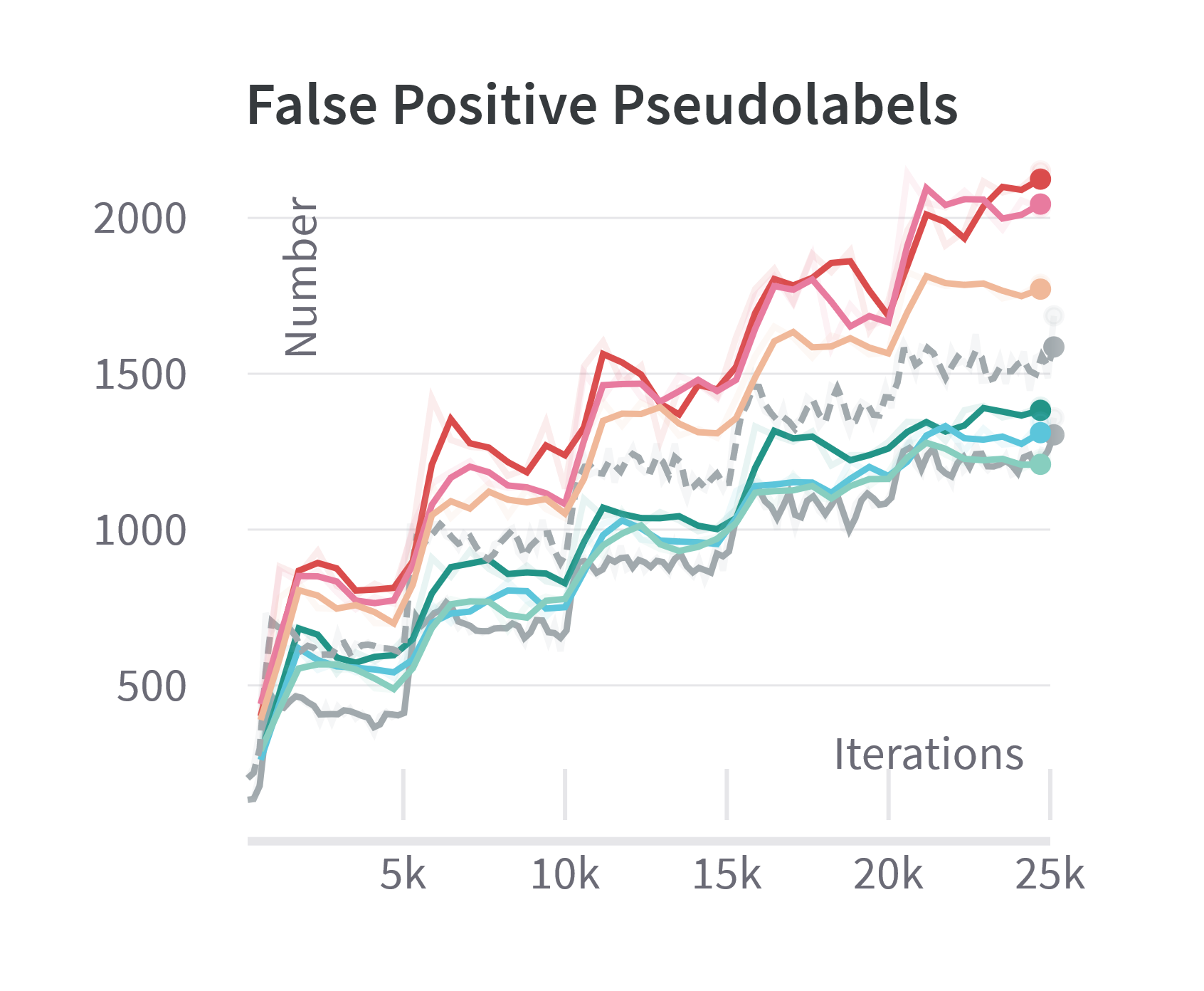}\label{fig:cata_fpp}}
{\includegraphics[trim={110 110 110 110},clip, width=0.32\textwidth]{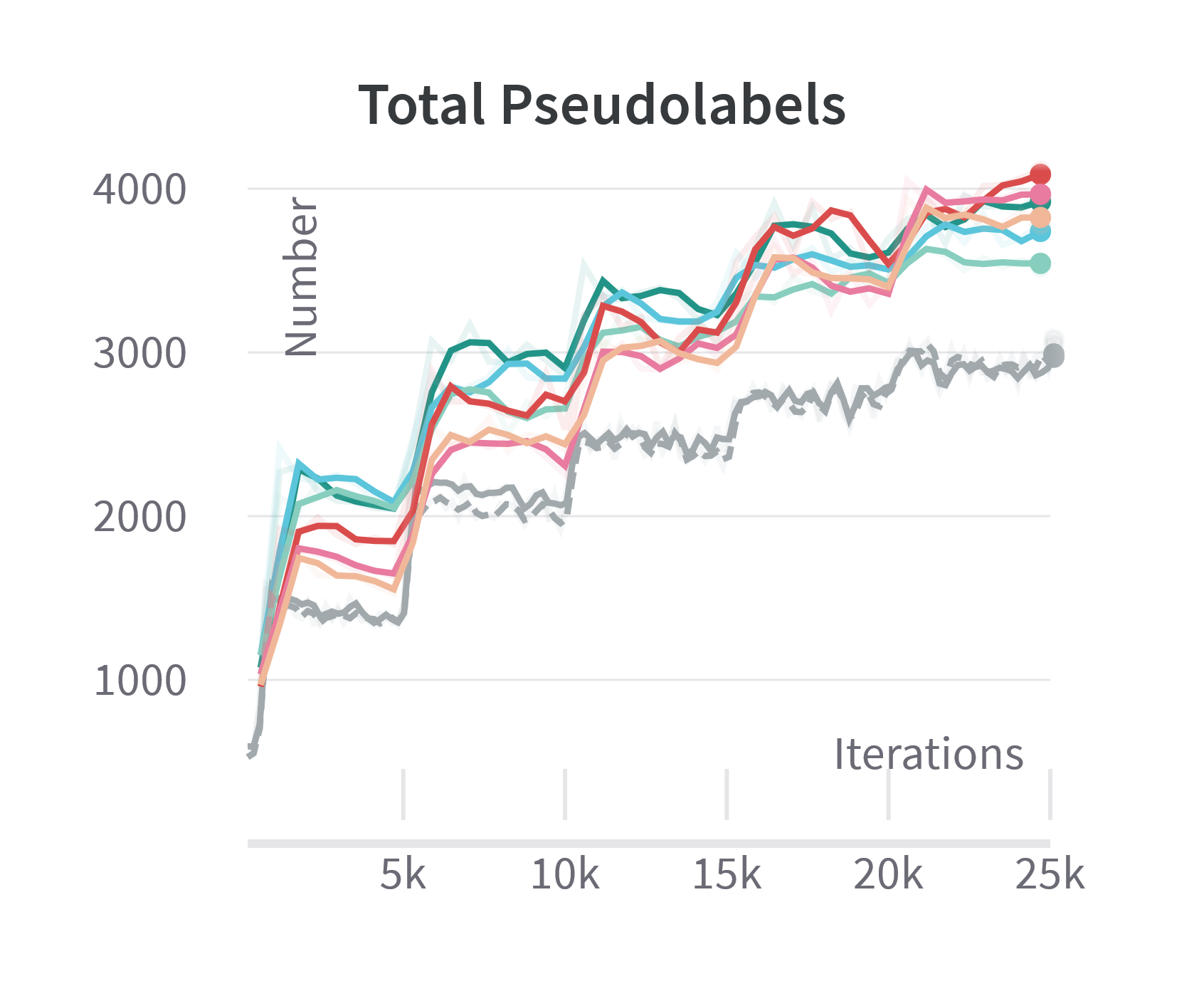}\label{fig:cata_total}}
{\includegraphics[trim={110 110 110 110},clip, width=0.32\textwidth]{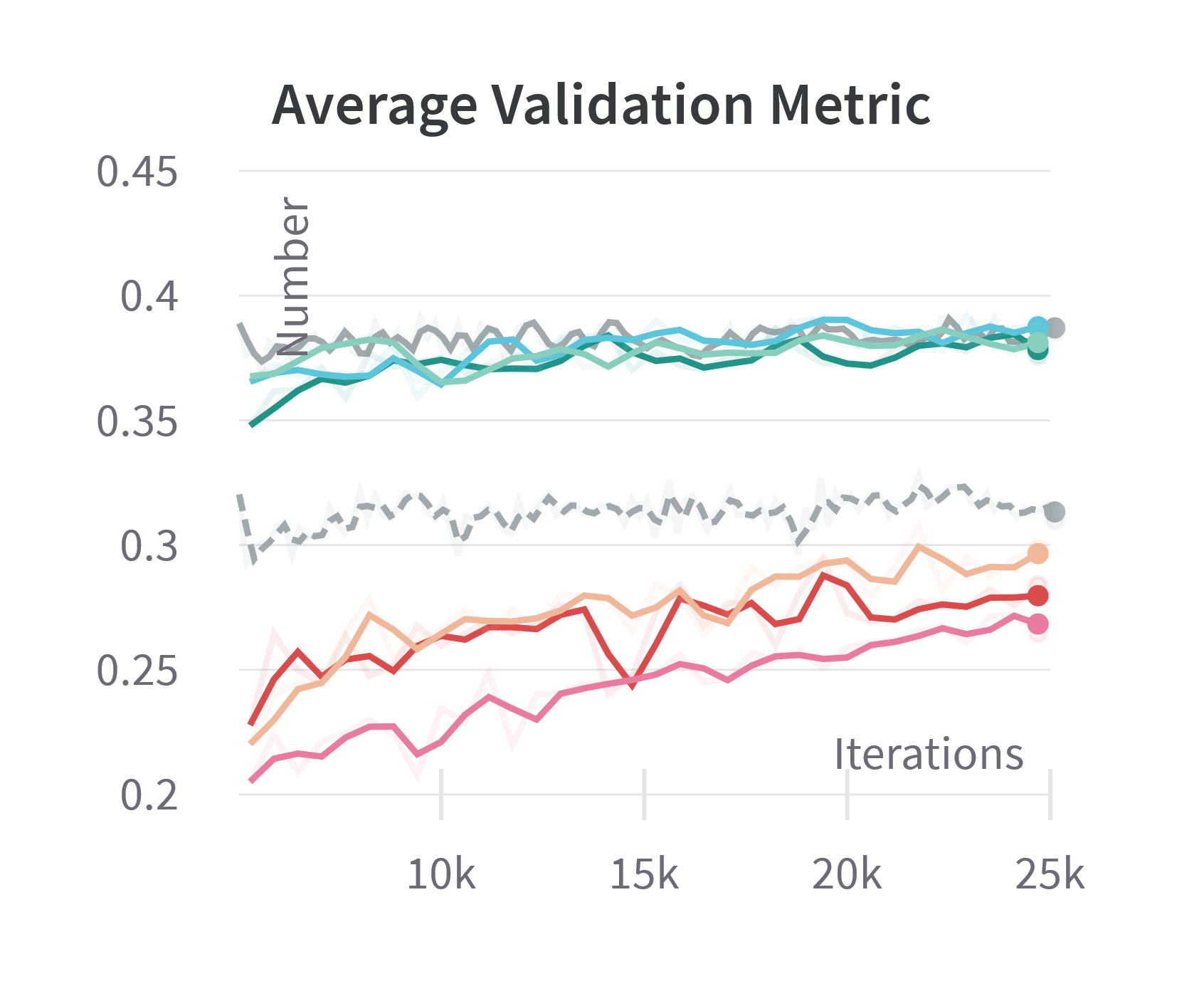}\label{fig:cata_avm}}
{\includegraphics[trim={110 110 110 110},clip, width=0.32\textwidth]{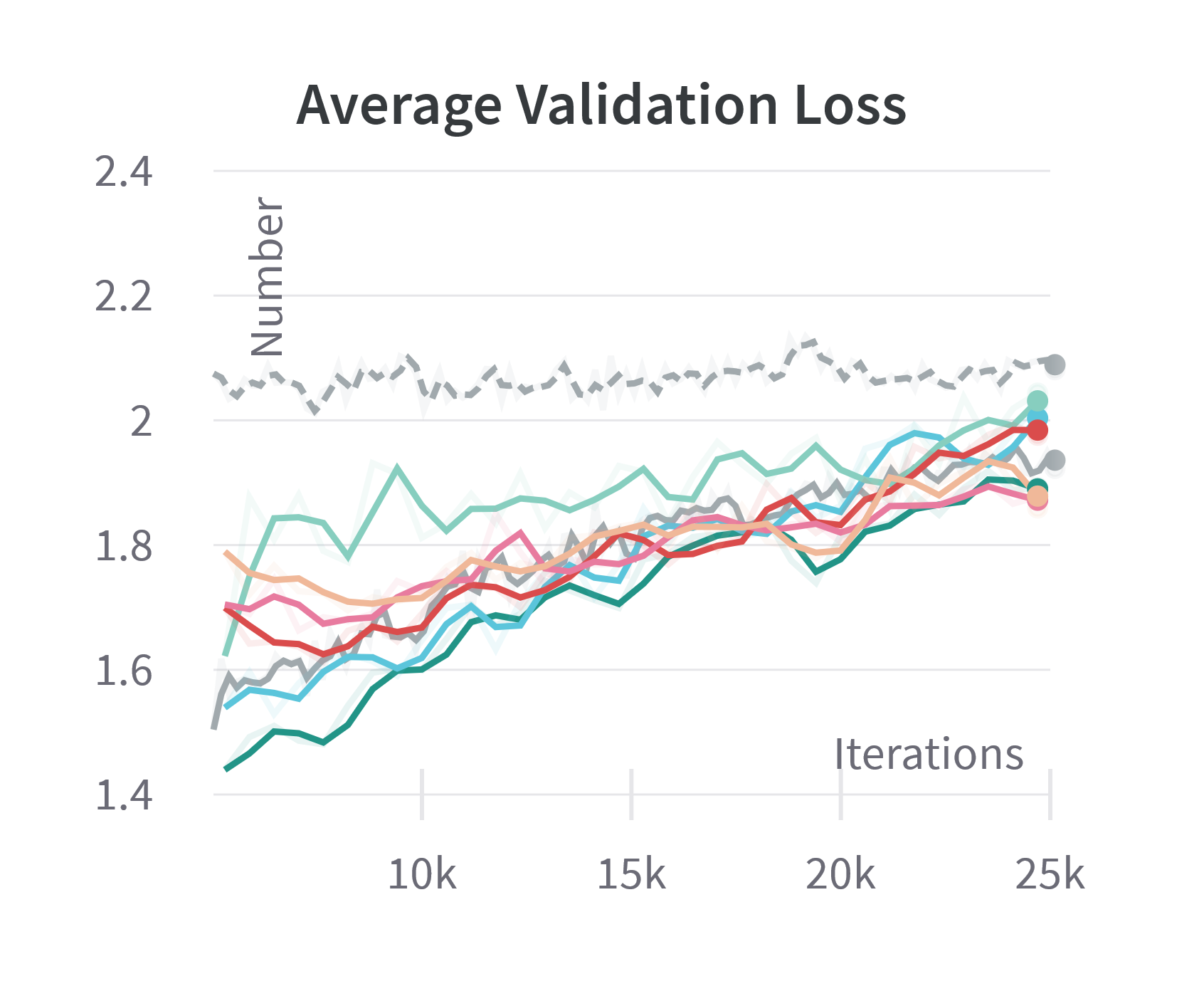}\label{fig:cata_avl}}
{\includegraphics[trim={110 110 110 110},clip, width=0.32\textwidth]{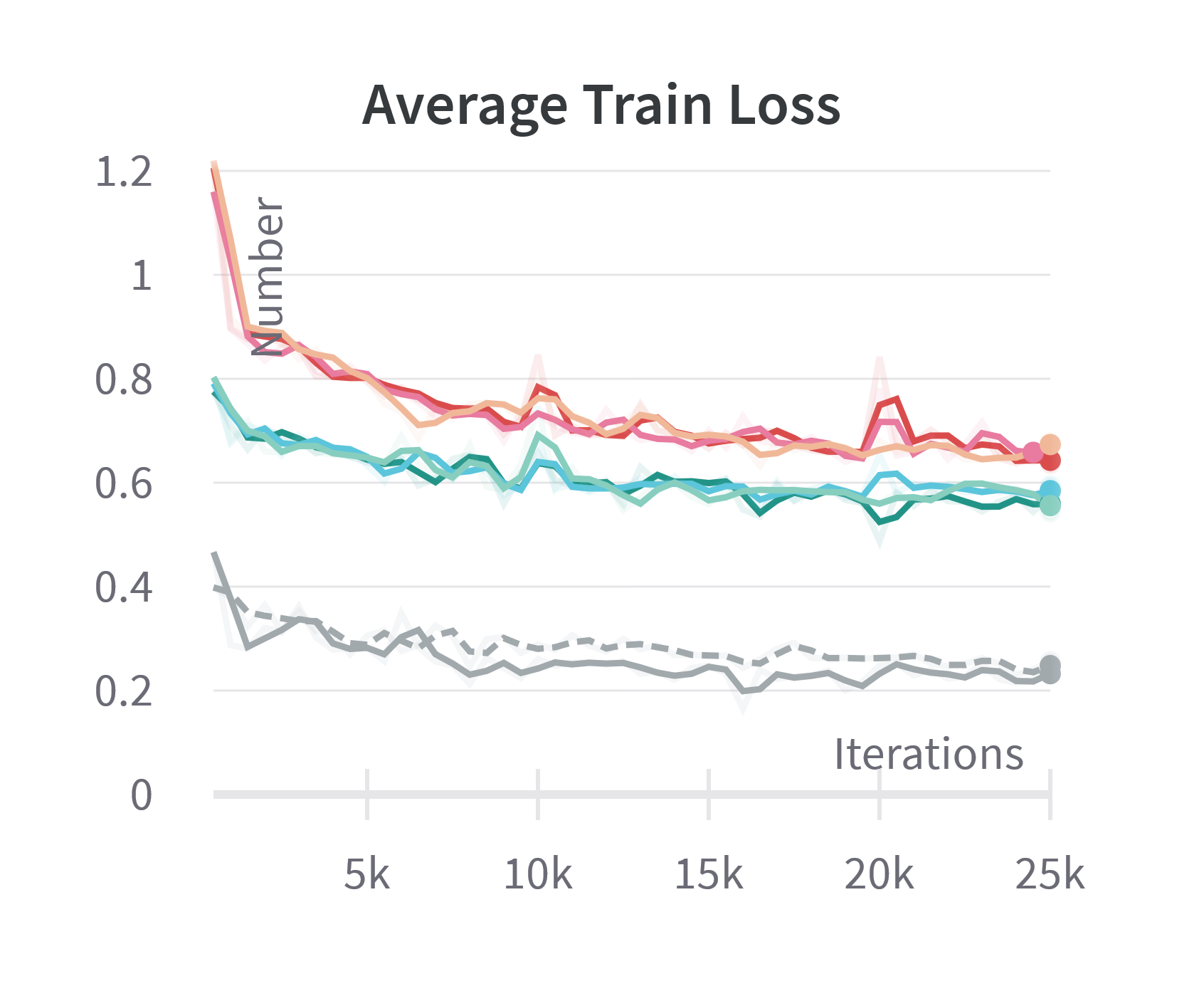}\label{fig:cata_atl}}
{\includegraphics[trim={200 0 200 1700},clip, width=0.85\textwidth]{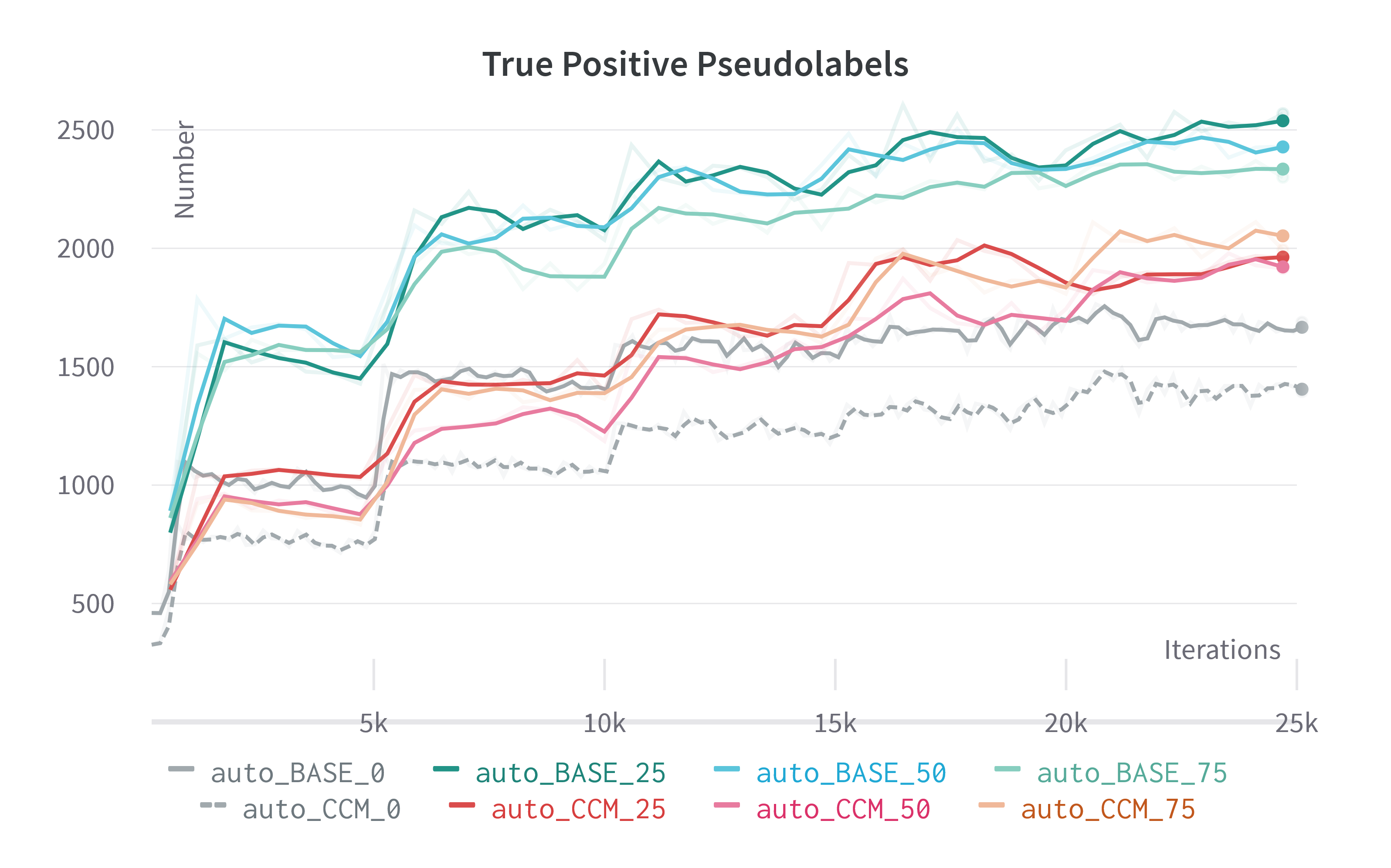}\label{fig:cata_legend}}
\caption{Training metrics of BASE and CCM methods for different levels of source data mixing}
\label{fig:cata}
\end{figure}

To address this issue, source dataset is introduced into self-training phase with the target dataset in each batch. Four levels of source data mixing are explored: 0$\%$, 25$\%$, 50$\%$, 75$\%$. Figure \ref{fig:cata} shows the training metrics for BASE and CCM methods with three levels of source data mixing during self-training. Figure \ref{fig:icm} in the previous section showed training metrics for ICM method with 0$\%$ and 25$\%$ of source data mixing. It is clear that mixing just 25$\%$ of source data is enough to help the model to generate true positive pseudo-labels at a higher rate. The average validation metric for the BASE method got slightly better compared to not mixing the source data, however the CCM method struggled. To calculate the final instance segmentation results the BASE method with 0$\%$ and 25$\%$ source data, CCM method with 0$\%$ and 75$\%$ source data, and ICM method with 0$\%$, 25$\%$ source data are picked. 

A proposed idea is to test different scaling transformations to optimize augmentations and reduce the domain gap, particularly given the diverse field of views in the Synthia and Cityscapes datasets. This included cropping smaller areas of the Synthia images to account for tiny object instances, however it did not help since the actual pixel-level area of the instances are not big enough.

\section{Guided Mask Improvement} \label{impro}

The proposed approach for guided instance pseudo-label generation involved using predicted semantic masks from the initially trained semantic model. Semantic segmentation provides a strong spatial prior for the location of object instances by segmenting the image into semantic masks indicating the regions where instances of that class may exist. These masks enforce the pseudo-label generation process during self-training. The search for object instances can be restricted to areas inside the semantic mask, rather than operating on the full image. This eliminates background clutter and allows for significant computational efficiency gains. The mask defines a clear boundary within which to sample random anchors for clustering algorithms, which also allows calculation of cluster means and assignments to be performed only using embeddings that fall inside the masked region. This lowers the load on the unlabelled push loss as well, since the anchor points will rarely land on zero labelled pixels. The inclusion of semantic masks allows filtering of noise and unstable segments and small isolated mask regions that do not generate associated instances are likely false positives and can be removed, which makes pseudo-label generation more robust.

\begin{figure}[h]
\centering
\captionsetup[subfloat]{labelformat=empty}
\subfloat[Instance ground-truth]{\includegraphics[trim={120 50 50 20},clip, width=0.33\textwidth]{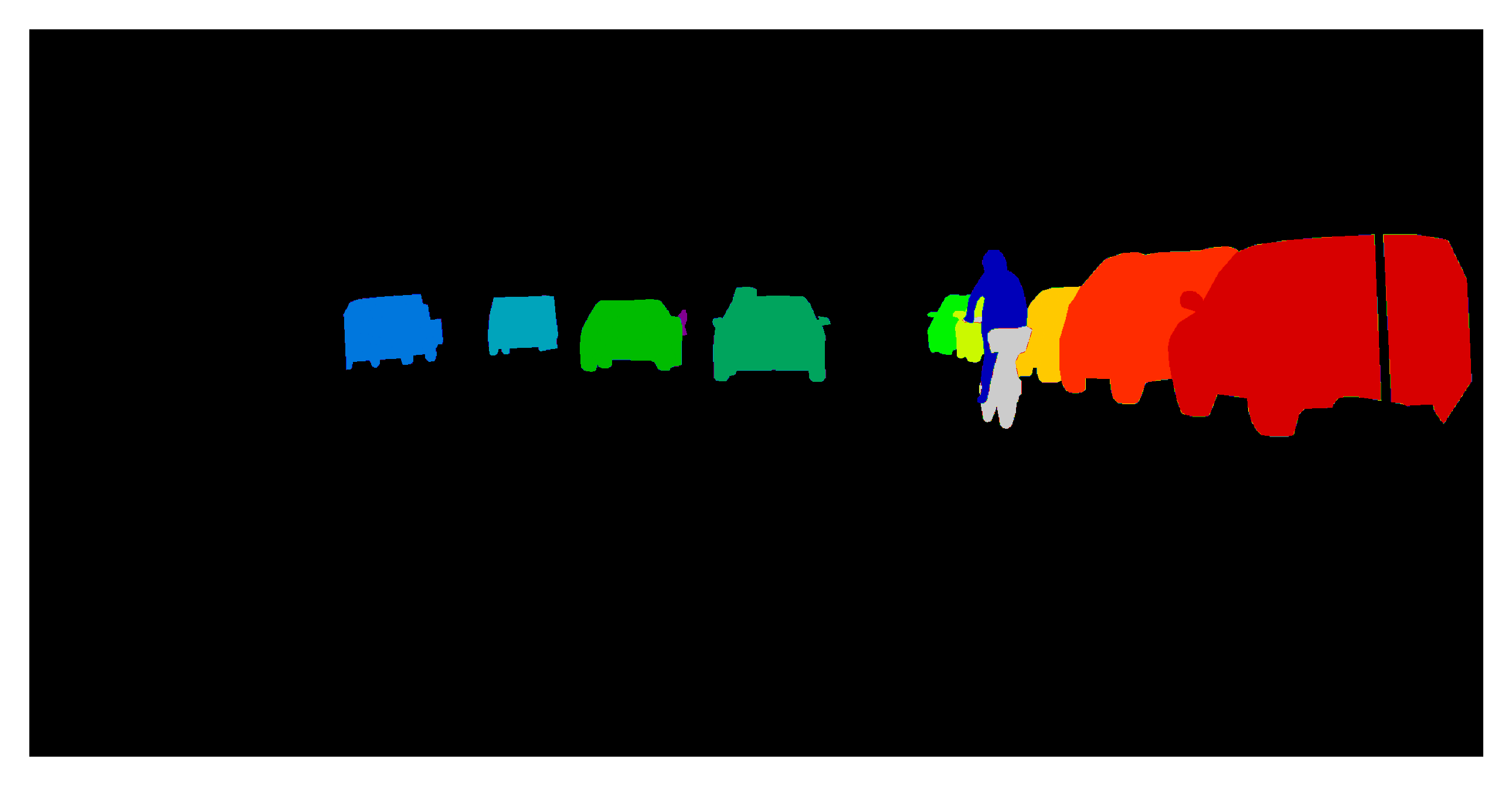}\label{fig:stuff_anchor3}}
\hfill
\subfloat[Predicted semantic mask]{\includegraphics[trim={120 50 50 20},clip, width=0.33\textwidth]{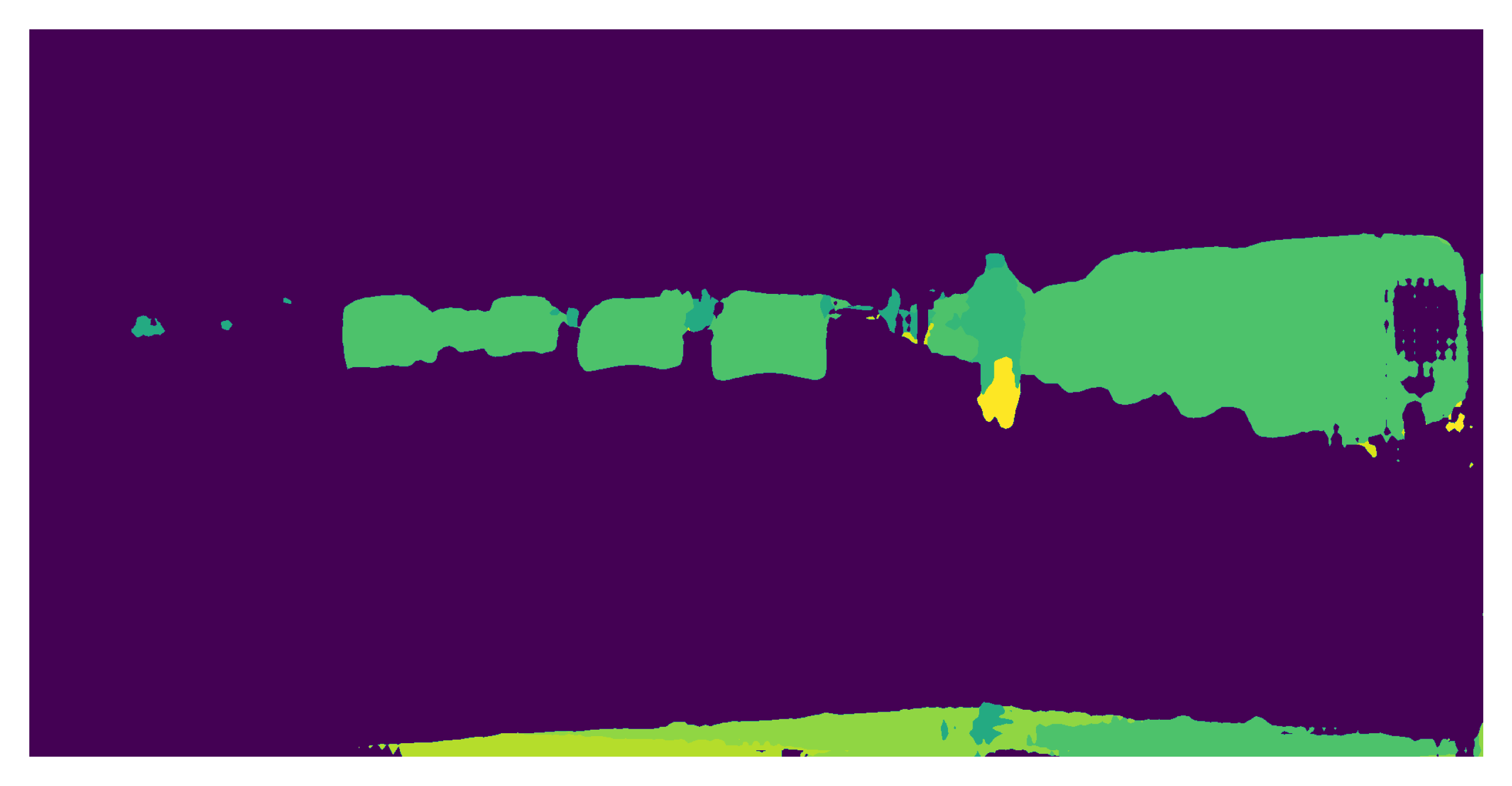}\label{fig:stuff_anchor1}}
\hfill
\subfloat[Cluster embedding]{\includegraphics[trim={120 50 50 20},clip, width=0.33\textwidth]{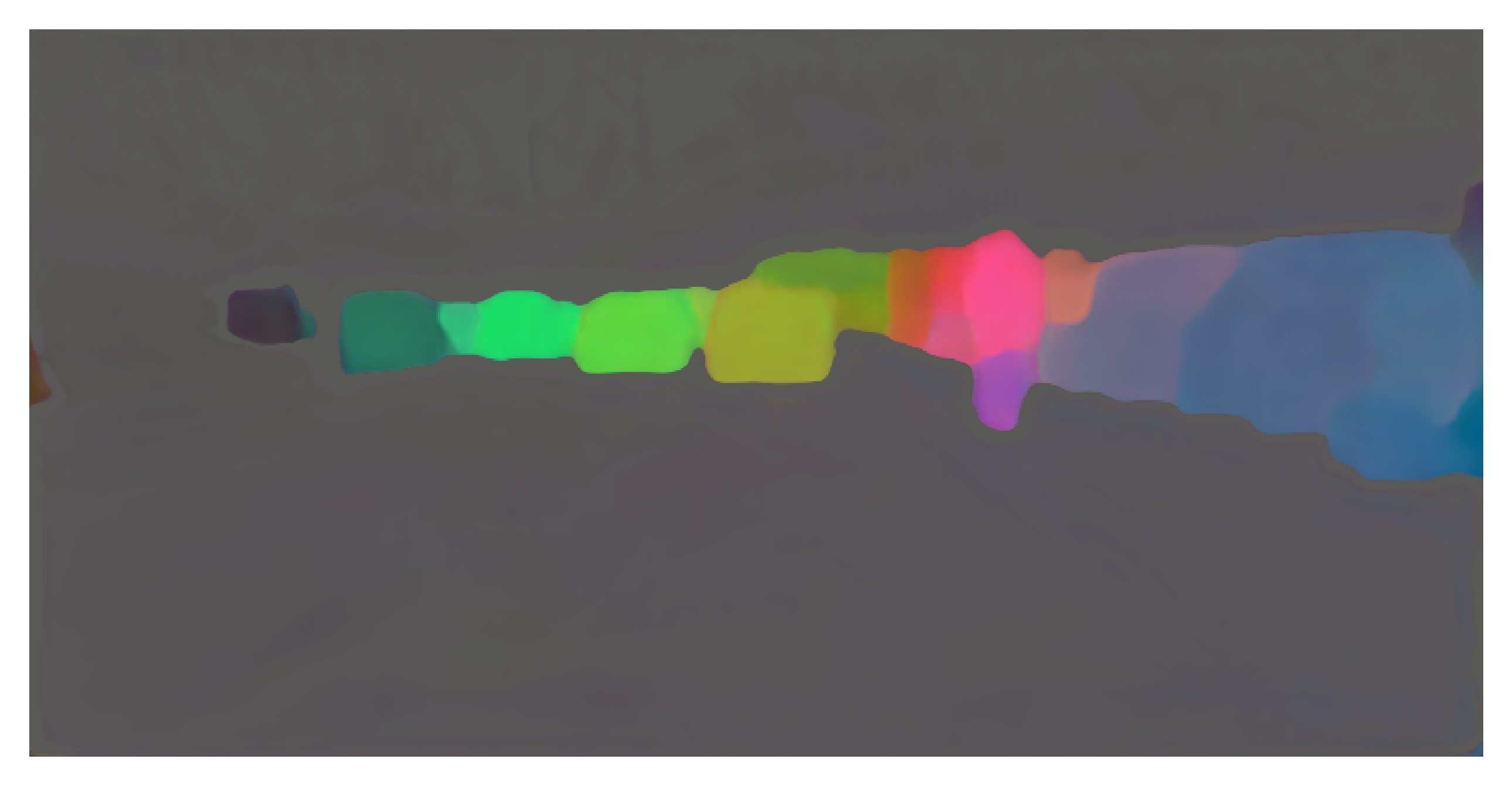}\label{fig:stuff_anchor2}}
\caption{Example of guiding instance clustering using predicted semantic masks}
\label{fig:stuff_anchor}
\end{figure}

\begin{figure}[h]
\captionsetup[subfloat]{labelformat=empty}
\centering
\subfloat[Inital semantic mask]{\includegraphics[trim={100 50 100 20},clip, width=0.33\textwidth]{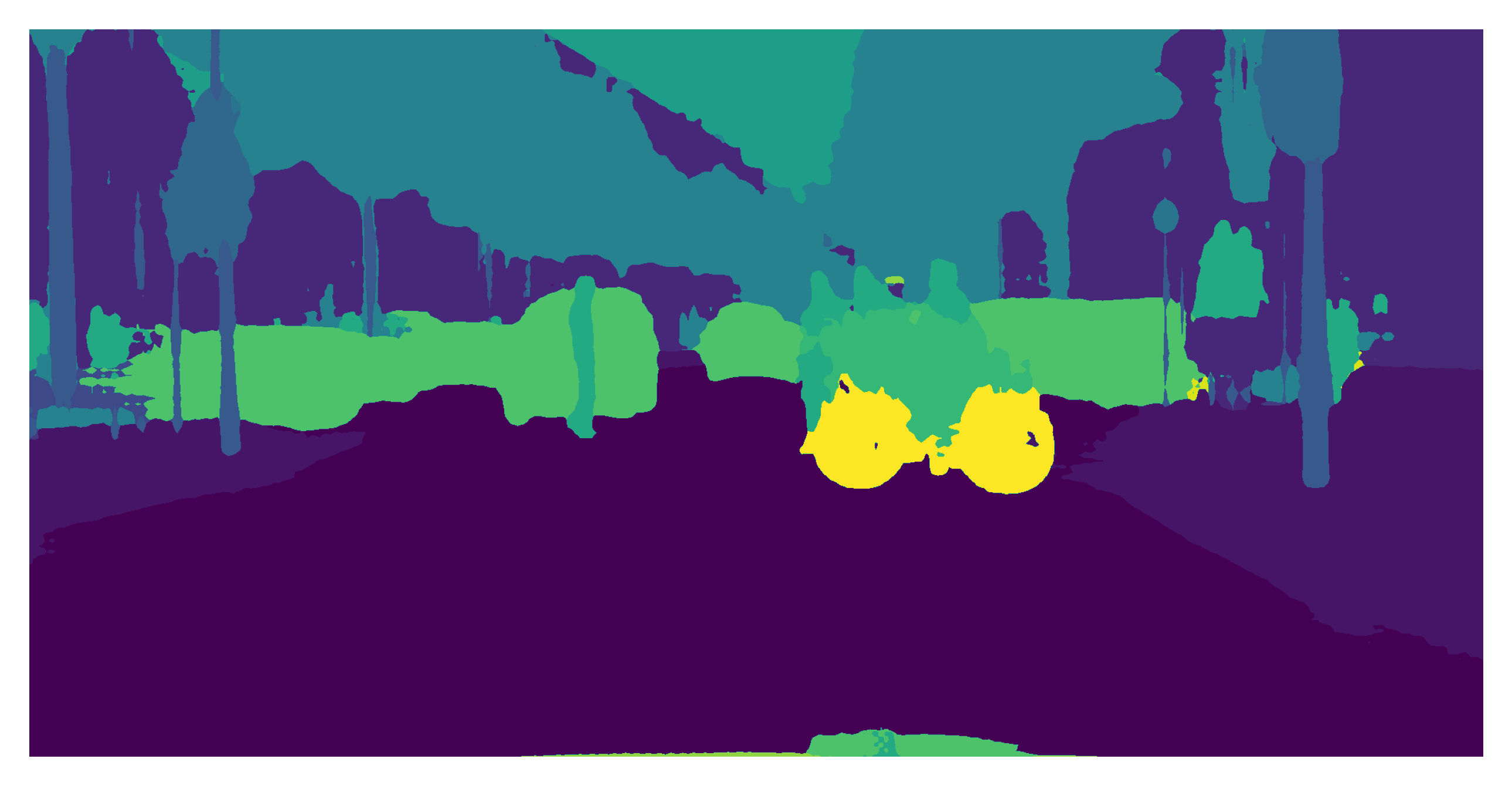}\label{fig:imp1}}
\hfill
\subfloat[Instance mask]{\includegraphics[trim={100 50 100 20},clip, width=0.33\textwidth]{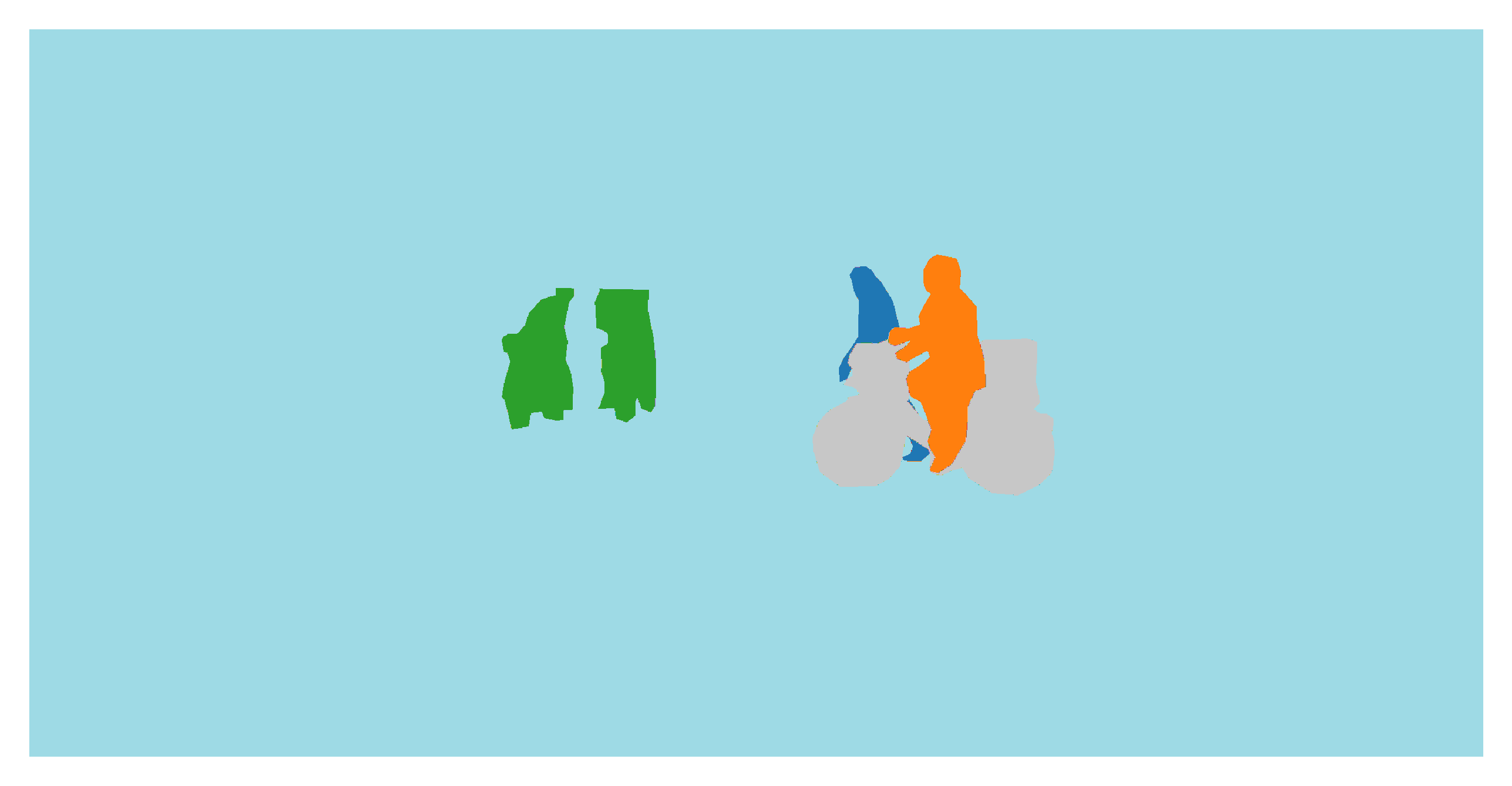}\label{fig:imp2}}
\hfill
\subfloat[Merged semantic mask]{\includegraphics[trim={100 50 100 20},clip, width=0.33\textwidth]{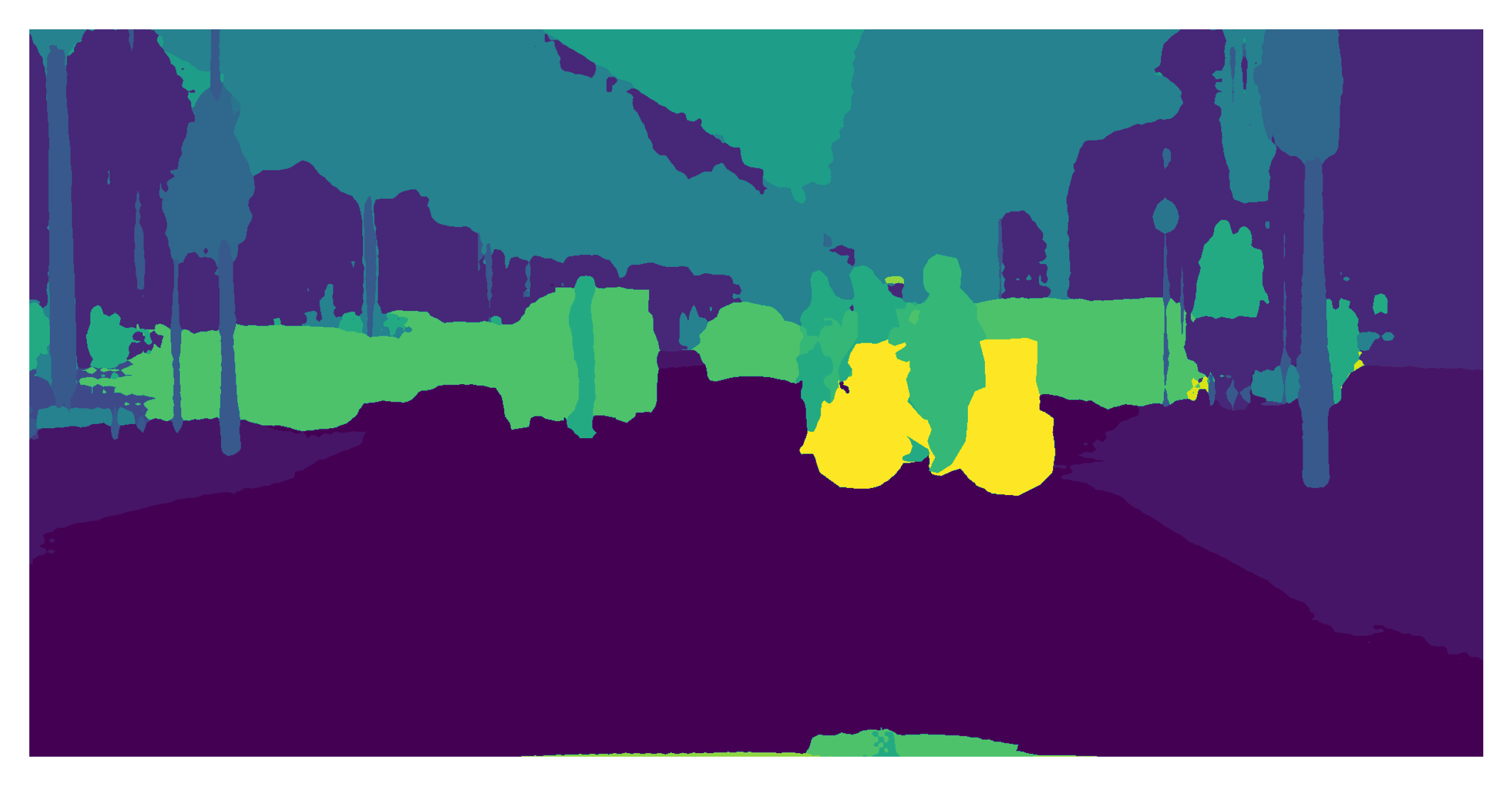}\label{fig:imp3}}
\caption{Example of improving semantic masks by merging instance masks}
\label{fig:merge}
\end{figure}

To further enhance semantic masks the proposed approach involves incorporation of instance masks generated by the instance model, which are superimposed onto semantic pseudo-label masks during self-training on target dataset. Figure \ref{fig:merge} shows example of creating merged masks by overlaying ground-truth instance mask over predicted semantic mask, in practice the predicted instance mask will be merged on to the pseudo-labels generated by the momentum network after the multi-scale fusion step. To validate this approach, tests are conducted using ground-truth instance masks to assess its viability and compare with the predicted instance masks generated from the instance self-training model. In this process, the batch of instance masks is subjected to the same augmentation and normalization transformations as the semantic masks. Each individual object's instance mask is then assigned a label ID that corresponds to the relevant semantic category, ensuring alignment. 

\begin{figure}[h]
\captionsetup[subfloat]{labelformat=empty}
\centering
\subfloat[Ground-truth semantic]{\includegraphics[trim={10 10 10 10},clip, width=0.245\textwidth]{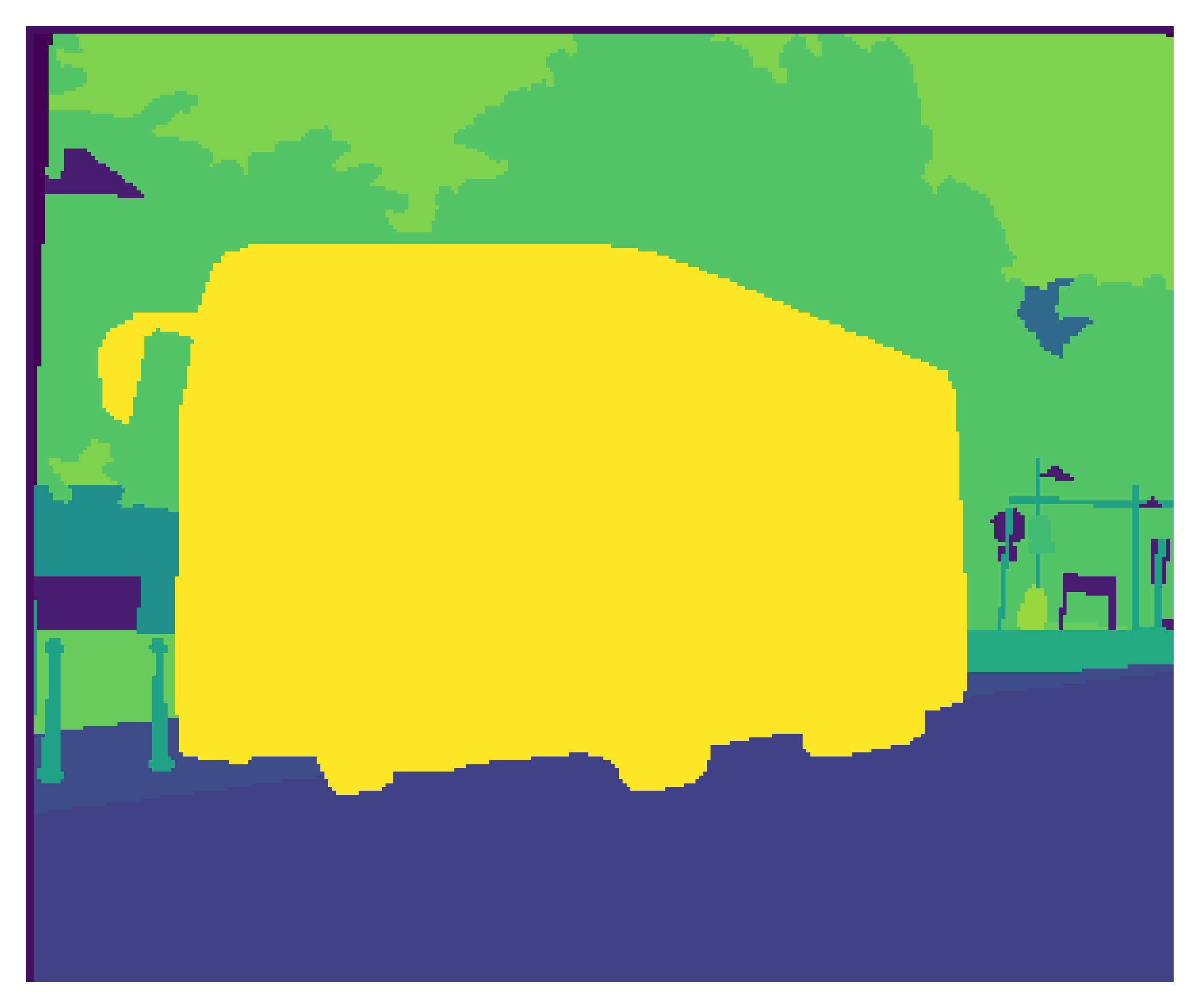}\label{fig:imp11}}
\hfill
\subfloat[Predicted semantic]{\includegraphics[trim={10 10 10 10},clip, width=0.245\textwidth]{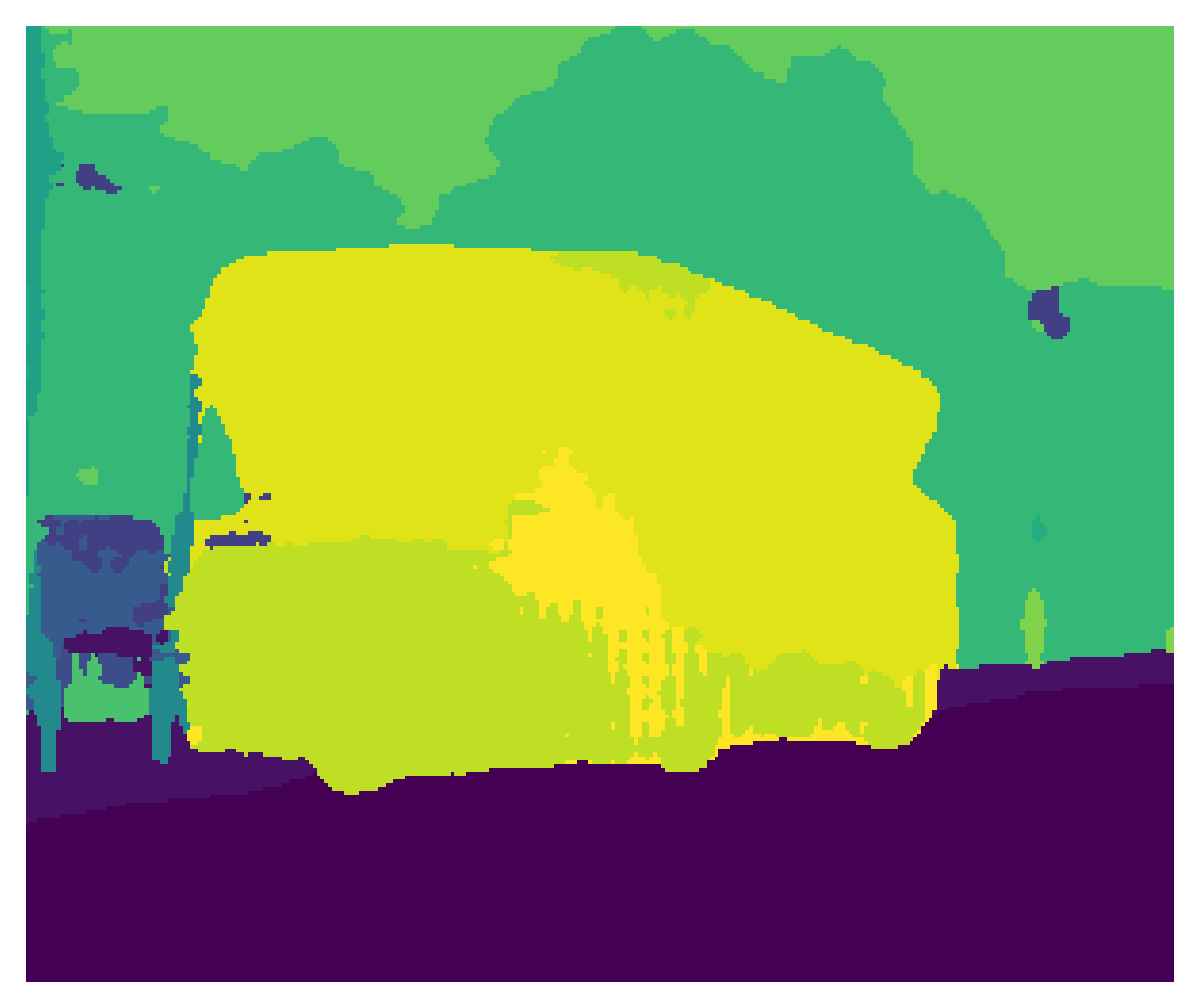}\label{fig:imp22}}
\hfill
\subfloat[Predicted instance]{\includegraphics[trim={10 10 10 10},clip, width=0.245\textwidth]{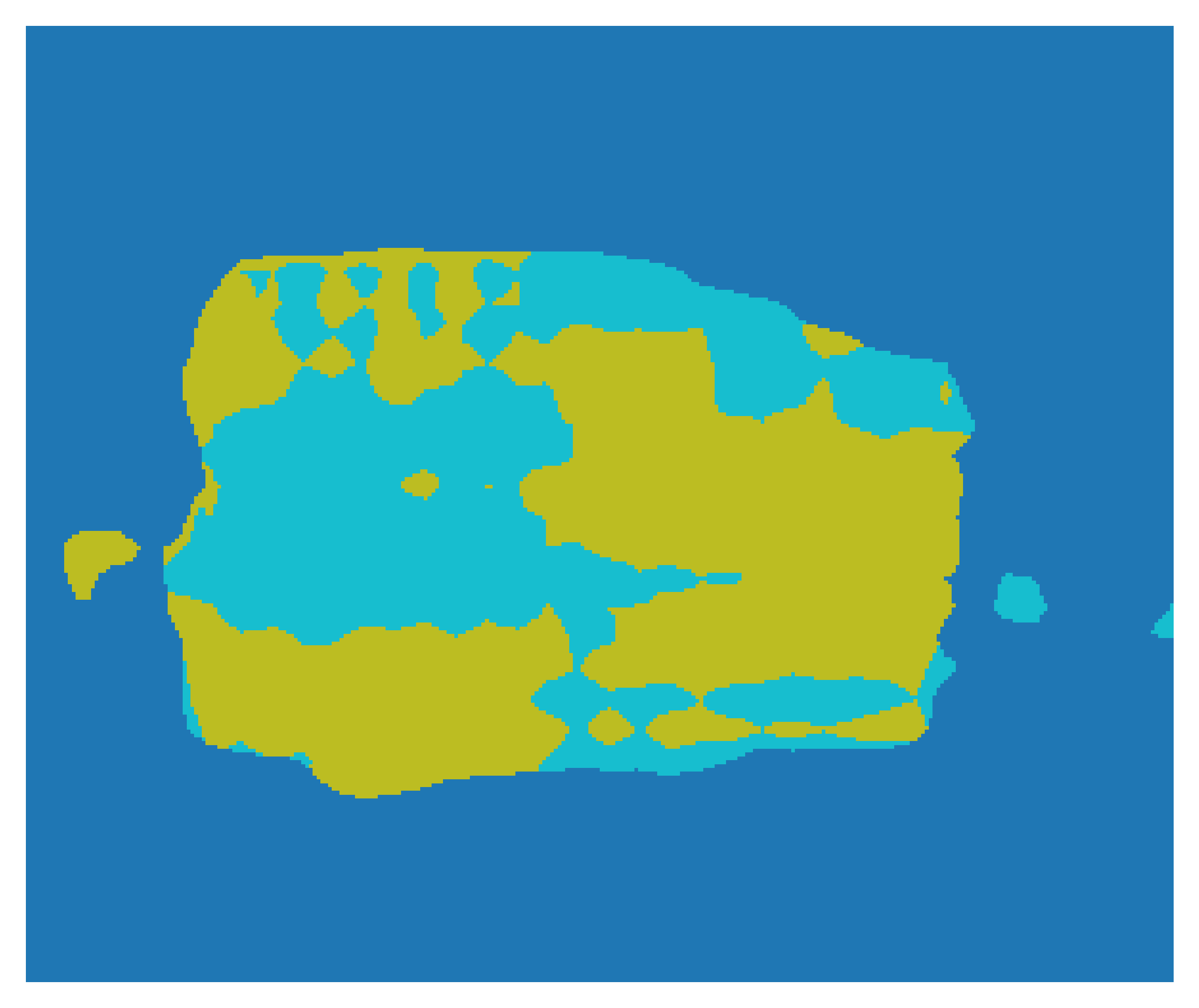}\label{fig:imp33}}
\hfill
\subfloat[Improved instance]{\includegraphics[trim={10 10 10 10},clip, width=0.245\textwidth]{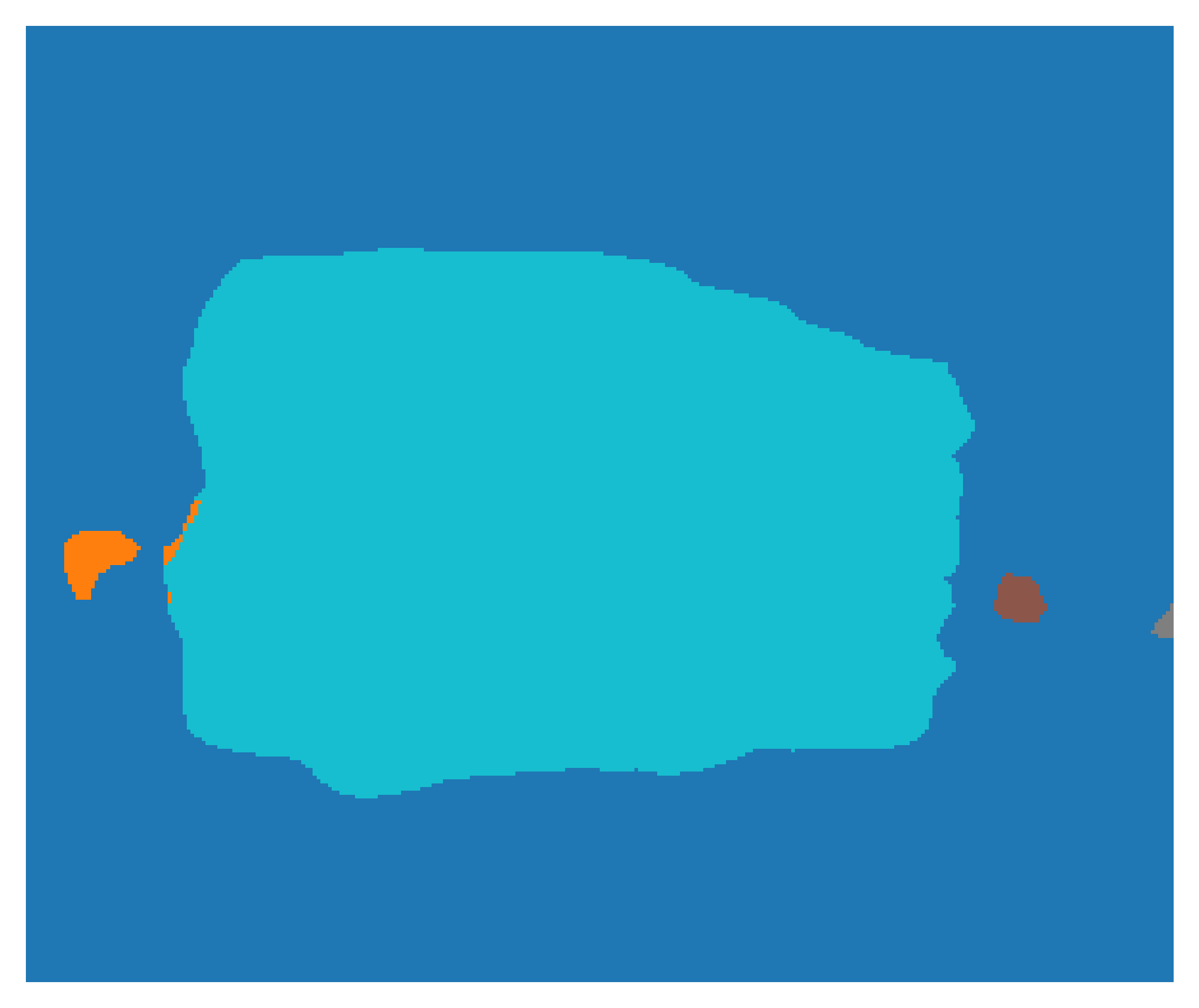}\label{fig:imp44}}
\caption{Example of improving instance masks by using semantic pixel bincount}
\label{fig:merge2}
\end{figure}

For ground-truth instance masks, random instances are selected within each image every iteration, while for the predicted instance masks, all masks are selected in every image. The predicted instance masks contained individual objects which had a unique label. These unique labels did not include information about the category of classes. To ensure correct alignment of these predicted instance masks with the semantic masks, pixel-wise bin-count is calculated to count the occurrences of each semantic ID for a particular object. The semantic ID with the highest count for each pixel is assigned as the new instance label for the object. This method improved the instance masks dramatically for objects like bus and truck where they are split into two or three parts due to poor semantic masks, illustrated in figure \ref{fig:merge2}. These improved instance masks are then overlaid onto the pseudo-label masks during semantic model training, and the resulting merged mask forms the basis for calculating the network's target loss.

\chapter{Unfinished Implementations}

This chapter contains ideas that are not implemented in time to successfully run experiments on. A basic overview of the proposed ideas are presented without going into implementation details.

\section{Objectness Classifier}

The persitent issue of increased false positives can be due to erroneous clusters that diverge from the primary object instance. To address this problem, a proposed method involves training an additional classifier on the source dataset. This classifier's objective is to learn the "objectness" of a mask across all classes through binary classification, determining whether a given mask corresponds to a real object or not. The Faster R-CNN paper by \cite{ren2015faster} introduced region based proposal network to predict instance object boundaries and objectness probability score at each position using a CNN network. However the embedding based method does not require object proposals and hence a simple binary objectness classifier should be enough to filter out false positive pseudo-labels.

\section{Active Learning}

Another proposed idea to further improve pseudo-label generation for the instance segmentation model is to explore active learning. The aim of human-in-loop active learning is to leverage human expertise to guide the model in selecting high-quality pseudo-labels for training. This process involves the integration of a user-interface into the training pipeline, implemented as a plugin in software like Napari \cite{chiu2022napari}. The integration of a Napari is an approach to involve human expertise, where the plugin presents the expert with image pairs: one from the source domain and one from the target domain, both overlaid with pseudo-labels generated by the instance segmentation model. The expert's task is to select the higher quality segmentation image which looks closer to ground truth annotations. The selected image is used to update the model's pseudo-labels for the target domain. This step may help in reduce false positive pseudo-labels and align the model's understanding of instances with the target domain. The process can be iterative, where the model continually refines its pseudo-labels based on human feedback and this refinement loop allows the model to adapt more effectively to the target domain.

\part{Results}
\chapter{Semantic Segmentation}

\begin{figure}[h]
\centering
\captionsetup[subfloat]{labelformat=empty}
\rotatebox[origin=l]{90}{First}\hfill
\subfloat{\includegraphics[trim={10 40 10 60},clip, width=0.97\textwidth]{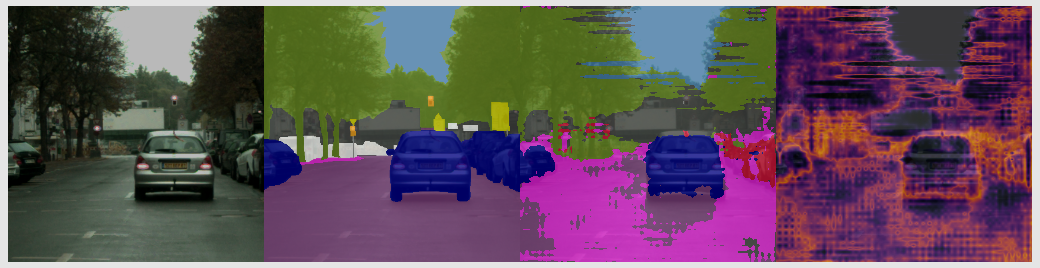}\label{fig:base1}}
\hfill
\rotatebox[origin=l]{90}{Final}\hfill
\newcommand\tab[1][1.5cm]{\hspace*{#1}}
\subfloat[Input Image \tab Ground-truth \tab Prediction \tab Error Map]{\includegraphics[trim={10 45 10 55},clip, width=0.97\textwidth]{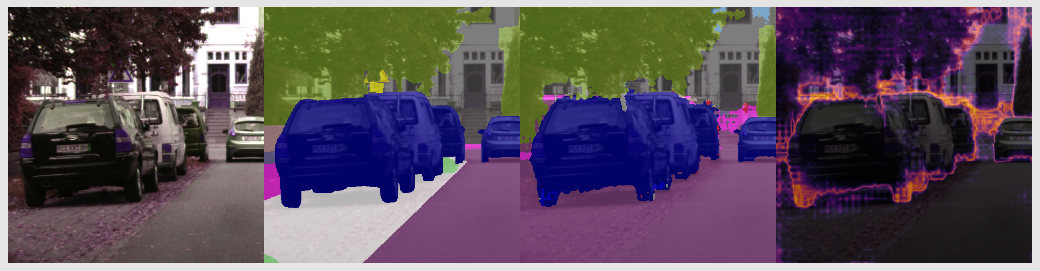}\label{fig:basel2}}
\caption{First vs final epoch segmentation results of baseline semantic model}
\label{fig:sem_base}
\end{figure}

The semantic segmentation task involved classifying each pixel in the image into a specific category or class. The baseline semantic models are trained with full supervision on the source dataset, while the self-train semantic model relied on pseudo-label generation and knowledge distillation to learn features of the target dataset. Figure \ref{fig:sem_base} shows the difference between predicted segmentation for first and final epoch of the baseline semantic model. These predictions are made on Cityscapes validation dataset. The final epoch image results reflect the performance of the baseline semantic model for pure domain adaptation from source dataset with 0 shot training on the target dataset.

\begin{figure}[h]
\centering
\captionsetup[subfloat]{labelformat=empty}
\rotatebox[origin=l]{90}{First}\hfill
\subfloat{\includegraphics[trim={50 10 50 10},clip, width=0.97\textwidth]{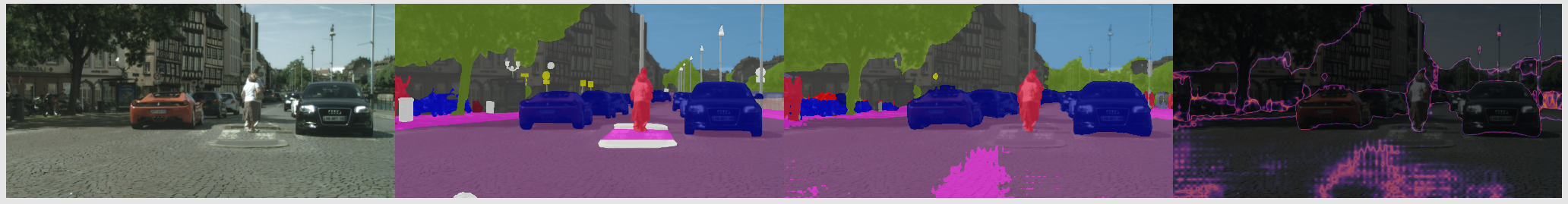}\label{fig:ssl1}}
\hfill
\rotatebox[origin=l]{90}{Final}\hfill
\newcommand\tab[1][1.5cm]{\hspace*{#1}}
\subfloat[Input Image \tab Ground-truth \tab Prediction \tab Error Map]{\includegraphics[trim={50 10 50 10},clip, width=0.97\textwidth]{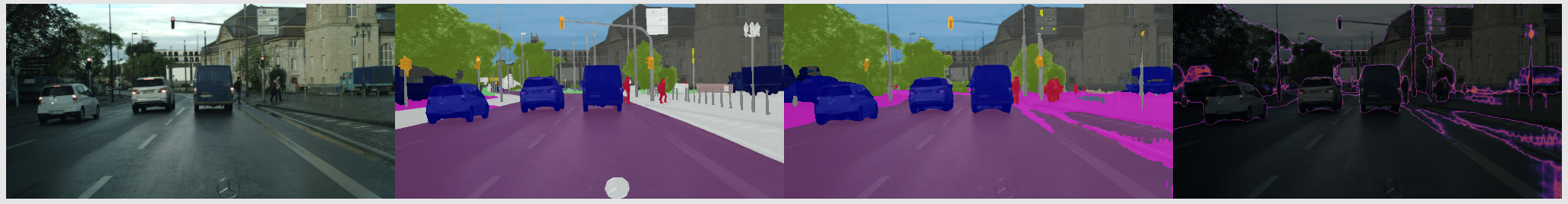}\label{fig:ssl2}}
\caption{First vs final epoch segmentation results of self-train semantic model}
\label{fig:sem_ssl}
\end{figure}

\begin{figure}[h]
\centering
\captionsetup[subfloat]{labelformat=empty}
\includegraphics[trim={0 0 0 0},clip, width=0.495\textwidth]{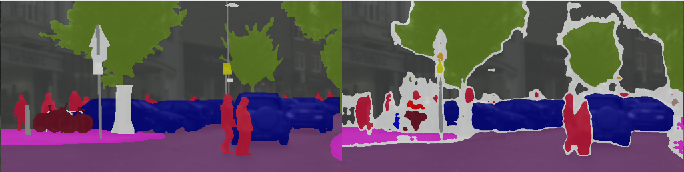}\label{fig:pl1}
\hfill
\includegraphics[trim={0 0 0 0},clip, width=0.495\textwidth]{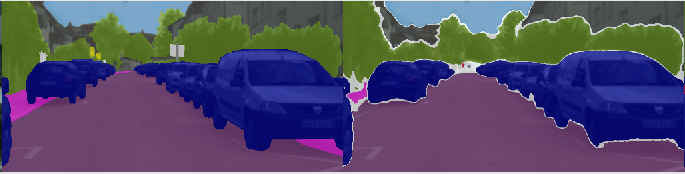}\label{fig:pl2}
\hfill
\newcommand\tab[1][1.5cm]{\hspace*{#1}}
\subfloat[Ground-truth \tab Pseudo-label]{\includegraphics[trim={0 0 0 0},clip, width=0.495\textwidth]{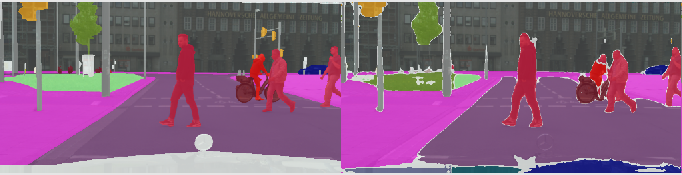}\label{fig:pl3}}
\hfill
\subfloat[Ground-truth \tab Pseudo-label]{\includegraphics[trim={0 0 0 0},clip, width=0.495\textwidth]{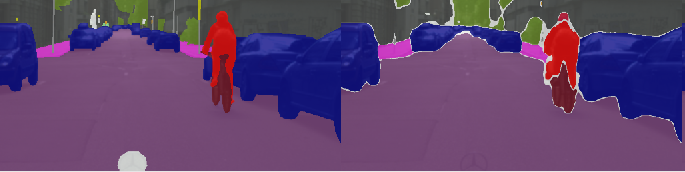}\label{fig:pl4}}
\caption{Ground-truth vs pseudo-label mask for self-train semantic model}
\label{fig:gt_pl}
\end{figure}

Figure \ref{fig:sem_ssl} shows the difference between predicted segmentation for first and final epoch of the self-train semantic model, and figure \ref{fig:gt_pl} shows the comparison between ground-truth semantic masks and pseudo-label masks generated by the momentum network. The images are taken from at increasing epochs and it is clearly visible how the pseudo-label generation and mask predictions gets more confident and accurate as training progresses.

\begin{figure}[h]
\centering
\includegraphics[trim={0 0 0 0},clip,width=0.99\textwidth]{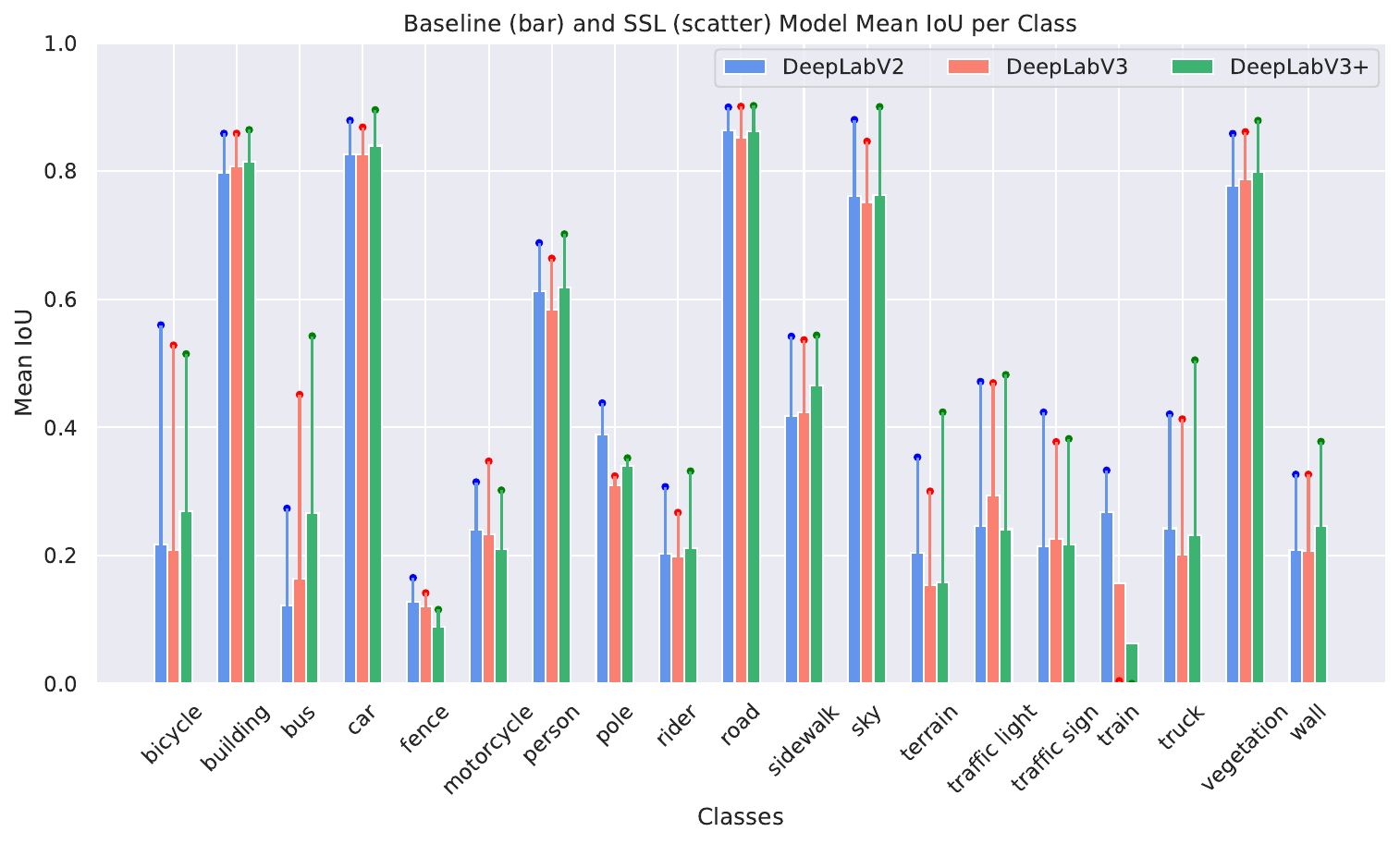}
\caption{MIoU (\%) per class on Cityscapes validation set for baseline and self-train semantic models. Baseline models represented with bars, Self-train models represented with scatter points.}
\label{fig:sem_base_ssl}
\end{figure}

Figure \ref{fig:sem_base_ssl} shows the per class MIoU score of all three architectures used for training the semantic models, evaluated on the Cityscapes validation set. There is a clear increase in performance in the self-train models compared to the baseline. The DeepLabV3+ architecture outperforms the V2 and V3 on majority of classes. However, the performance of V3 and V3+ architectures for the \{\textit{train}\} class is poor. All these models are trained with the Mixed dataset i.e. mixing 948 GTA-V images which contained both \{\textit{truck, train}\} classes to the Synthia dataset, since for these classes semantic annotations are missing in Synthia. The main reason Mixed dataset is used for the experiments is to obtain good semantic masks for the classes which would be used by the instance segmentation model. Althought DeepLabV3+ architecture gave the highest overall performance, it could not segment the \{\textit{train}\} class correctly at all. DeepLabV2 architecture performed only slightly worse compared to V3+ but provided good segmentation for all the important instance classes i.e. \{\textit{car, bus, truck, train, person, rider, motorcycle, bicycle}\}. Hence the DeepLabV2 model is used for training the instance segmentation models. 

Table \ref{tab:class_miou} shows the class-wise MIoU score of all three architectures, along with reported scores from the \cite{araslanov2021self} paper and the reproduced scores, evaluated on the Cityscapes validation set. The "Paper Method" models are trained purely on Synthia with 16 classes, while "Our Method" models are trained on the Mixed dataset with 19 classes. The 'Baseline' row refers to the models pre-trained on source domain only; the 'Self-train' row refers to the models self-trained on both source and target domain data using pseudo-label generation. The '$\mathrm{Improved_{GT}}$' row refers to the model self-trained with the guidance of 30\% of ground-truth instance masks, and '$\mathrm{Improved_{pred}}$' row refers to the model self-trained with predicted instance masks. It shows that both of them improve the scores further giving a higher score than the initial implementation. This training is done on top of the self-trained model checkpoint by giving additional instance mask guidance, and the method is explained in section \ref{impro}. It is seen that using instance mask guidance further boosts the MIoU score by approximately 0.2-0.3\% for the DeepLabV2 network.

\begin{sidewaystable}[h]
\centering
\begingroup
\setlength{\tabcolsep}{2pt}
\resizebox{\columnwidth}{!}{%
\begin{tabular}{l|ccccccccccccccccccc|cc}
\toprule 
\textbf{Paper Method} &road&side&build&wall&fenc&pole&light&sign&veg&terr&sky&pers&ride&car&truck&bus&train&moto&bicy&$\mathrm{MIoU^{16}}$&$\mathrm{MIoU^{19}}$\\
\midrule
\textbf{Araslanov} &&&&&&&&&&&&&&&&&&& \\
Baseline& 63.9& 25.9& 71.0& 11.0& 0.2& 36.9& 7.6& 20.0& 72.9&-& 75.5& 46.7& 16.7& 74.5&-& 15.8&-& 20.8& 21.7& 36.3 & -\\
Self-train& 89.3& 47.2& 85.5& 26.5& 1.3& 43.0& 45.5& 32.0& 87.1&-& 89.3& 63.6& 25.4& 86.9&-& 35.6&-& 30.4& 53.0& 52.6 & -\\
\textbf{Reproduced} &&&&&&&&&&&&&&&&&&& \\
Baseline& 64.9& 29.2& 64.7& 6.8& 0.09& 36.9& 0.8& 17.7& 73.7&-& 72.6& 53.2& 17.1& 71.2&-& 13.9&-& 16.4& 22.3& 35.1 & -\\
Self-train& 79.5& 36.4& 85.3& 25.9& 1.0& 41.8& 47.4& 31.3& 86.8&-& 88.8& 68.3& 25.5& 87.5&-& 37.7&-& 23.2& 52.8& 51.1 & -\\
\midrule
\textbf{Our Method} 
&road&sidew&build&wall&fence&pole&light&sign&veg&terr&sky&pers&ride&car&truck&bus&train&moto&bicy&$\mathrm{MIoU^{16}}$&$\mathrm{MIoU^{19}}$\\
\midrule
\textbf{DeepLabV2} &&&&&&&&&&&&&&&&&&& \\
Baseline &87.0&41.9&80.2&21.0&12.8&38.8&24.5&21.3&77.8&20.3&76.1&61.2&20.3&82.5&24.2&12.2&26.7&24.0&21.6& 43.7 &40.8 \\
Self-train 
&90.7&55.7&85.8&32.7&\textbf{16.5}&\textbf{43.7}&47.1&\textbf{42.3}&85.8&34.9&88.0&68.7&30.7&87.9&42.0&27.3&\textbf{33.2}&31.4&55.9& 55.6 &52.6 \\
\textbf{DeepLabV3} &&&&&&&&&&&&&&&&&&& \\
Baseline &85.6&42.6&80.9&20.7&12.0&31.0&29.3&22.5&78.7&15.4&75.1&58.3&19.8&82.6&20.1&16.4&15.6&23.3&20.8&43.7&39.6 \\
Self-train &\textbf{90.8}&54.9&85.9&32.6&14.1&32.3&46.9&37.3&86.1&30.0&84.6&66.3&26.6&86.8&41.2&45.1&0.4&\textbf{34.7}&52.8& 54.8 &50.0 \\
\textbf{DeepLabV3}+ &&&&&&&&&&&&&&&&&&& \\
Baseline &86.8&47.0&81.7&24.6&8.8&34.0&24.0&21.7&79.9&15.8&76.3&61.8&21.1&83.9&23.1&26.6&6.2&21.4&26.9&45.4&40.6 \\
Self-train &\textbf{90.8}&\textbf{55.7}&\textbf{86.4}&\textbf{37.9}&11.5&35.2&\textbf{48.2}&38.2&\textbf{87.8}&\textbf{42.3}&\textbf{90.0}&\textbf{70.1}&\textbf{33.1}&\textbf{89.5}&\textbf{50.4}&\textbf{54.2}&0.0&30.1&51.4&\textbf{56.8}&\textbf{52.8} \\
\midrule
\textbf{DeepLabV2} &&&&&&&&&&&&&&&&&&& \\
$\mathrm{Improved_{pred}}$ &90.7&55.7&85.8&32.7&16.5&43.7&47.1&42.3&85.8&34.9&88.0&69.2&30.9&\underline{88.3}&42.2&27.3&\underline{35.7}&31.4&56.2&55.72&52.86\\
$\mathrm{Improved_{GT}}$ &90.7&55.7&85.8&32.7&16.5&43.7&47.1&42.3&85.8&34.9&88.0&\underline{69.6}&\underline{31.4}&88.1&\underline{42.4}&27.3&35.4&31.4&\underline{56.8}&55.81&52.91\\
\bottomrule
\end{tabular}%
}
\endgroup
\caption{MIoU (\%) per class on Cityscapes validation for baseline, self-train and improved semantic models}
\label{tab:class_miou}
\end{sidewaystable}

\chapter{Instance Segmentation}

The instance segmentation task involved classifying each pixel into individual objects within an image. The baseline instance models are trained with full supervision on the source dataset while the self-train instance model is divided into three category-centric models which relied on clustering algorithms to generate pseudo-labels to learn features of the target dataset. Table \ref{tab:inst_miou} shows the results for class-wise \textbf{Mean Average Precision} (mAP) \cite{everingham2010pascal} computed on Cityscapes validations set. Here mAP is defined as:
\begin{equation}
\mathrm{mAP} = \frac{1}{N} \sum_{c \in N} \frac{|TP_c|}{(|FP_c|+|TP_c|)},
\end{equation}
where N is number of classes and the IoU threshold value set to 50\%. The mAP scores are calculated for instance baseline models with different levels of consistency loss weight $\delta$ from equation \ref{eq:contrastive}, and model trained with 0.1 consistency loss weight is used for all further results. Note that these results are calculated using Cityscapes' ground-truth semantic masks to perform guided clustering of the network embeddings.

\begin{table}[h]
\centering
\begingroup
\begin{tabular}{c|cccccccc|c}
\toprule
\textbf{$\delta$ value }  & car  & bus & truck & train & pers & ride & bicy & moto & $\mathrm{mAP^8}$ \\ \midrule
0.0 &30.8 &38.4 &53.5 &33.1 &\textbf{58.3} &73.7 &\textbf{65.0} &75.4 &53.5 \\ 
0.01&30.4 &31.6 &51.3 &21.1 &57.0 &74.6 &64.6 &\textbf{77.7} & 51.0 \\ 
0.1 &\textbf{31.9} &\textbf{47.2} &\textbf{56.7} &\textbf{38.5} &51.8 &\textbf{74.9} &58.8 &77.4 &\textbf{54.7} \\ 
1.0 &26.8 &29.5 &42.2 &24.0 &48.3 &61.0 &50.0 &69.0 & 43.8 \\ 
\bottomrule
\end{tabular}%
\endgroup
\caption{mAP (\%) per class on Cityscapes validation for baseline instance models with different levels of consistency loss weight $\delta$. With ground-truth semantic mask guided clustering.}
\label{tab:inst_miou}
\end{table}

Table \ref{tab:inst_miou_ssl} shows the class-wise mAP score for baseline and self-train instance segmentation models using predicted semantic masks to cluster the embeddings, without any ground-truth guidance. In this table the baseline model is compared to various self-train models trained with different training frameworks i.e. ICM, BASE, CCM; and the numbers 0 and 25 refer to the percentage of source data used to prevent catastrophic forgetting. It is clear that the ICM method outperforms BASE and CCM by a big margin. The additional MIX* method uses the best model from each method during inference to obtain the best set of mAP scores. As per current knowledge, no known study has reported scores for Synthia to Cityscapes domain adaptive self-trained instance segmentation task.

\begin{table}[H]
\centering
\begingroup
\resizebox{\columnwidth}{!}{%
\begin{tabular}{l|cccccccc|c}
\toprule
\textbf{Method}  & car  & bus & truck & train & pers & rider & bicy & moto & $\mathrm{mAP^8}$ \\ \midrule
Baseline & 12.55 & 9.50 & 9.96 & 0.61 & 6.06 & 7.05 & 7.98 & 3.04 & 7.09 \\ 
Self-train$\mathrm{_{ALL_0}}$  & 16.34 & 3.73 & 4.42 & 1.58 & 8.25 & 5.56 & 5.75 & 2.87 & 6.06 \\ 
Self-train$\mathrm{_{CCM_0}}$  & 13.36 & 3.50 & 8.19 & 1.58 & 9.24 & 6.90 & 8.59 & 2.88 & 6.78 \\ 
Self-train$\mathrm{_{CCM_{75}}}$ & 12.83 & 4.10 & 5.24 & 1.58 & 6.30 & 4.63 & 7.95 & 3.24 & 5.73 \\ 
Self-train$\mathrm{_{BASE_0}}$ & 17.92 & 5.92 & 8.12 & 1.58 & 9.62 & \textbf{8.07} & \textbf{10.57} & \textbf{4.53} & 8.29 \\ 
Self-train$\mathrm{_{BASE_{25}}}$ & 18.03 & 5.80 & 4.95 & 2.22 & 7.96 & 6.21 & 8.72 & 2.83 & 7.09 \\ 
Self-train$\mathrm{_{ICM_{0}}}$ & 17.82 & 20.81 & 6.39 & 0.74 & \textbf{9.87} & 7.85 & 10.34 & 3.10 & 9.61 \\ 
Self-train$\mathrm{_{ICM_{25}}}$  & \textbf{18.37} & \textbf{27.42} & \textbf{22.05} & \textbf{11.11} & 8.94 & 6.47 & 9.77 & 3.89 & \textbf{13.50} \\ 
\midrule
Self-train$\mathrm{_{MIX}*}$ & \textbf{18.37} & \textbf{27.42} & \textbf{22.05} & \textbf{11.11} & \textbf{9.87} & \textbf{8.07} & \textbf{10.57} & \textbf{4.53} & \textbf{14.00} \\ 
\bottomrule
\end{tabular}}
\endgroup
\caption{mAP (\%) per class on Cityscapes validation for baseline and self-train instance models. With predicted semantic mask guided clustering.}
\label{tab:inst_miou_ssl}
\end{table}

\begin{figure}[H]
\centering
\captionsetup[subfloat]{labelformat=empty}



\subfloat{\includegraphics[trim={70 20 0 10},clip, width=0.33\textwidth]{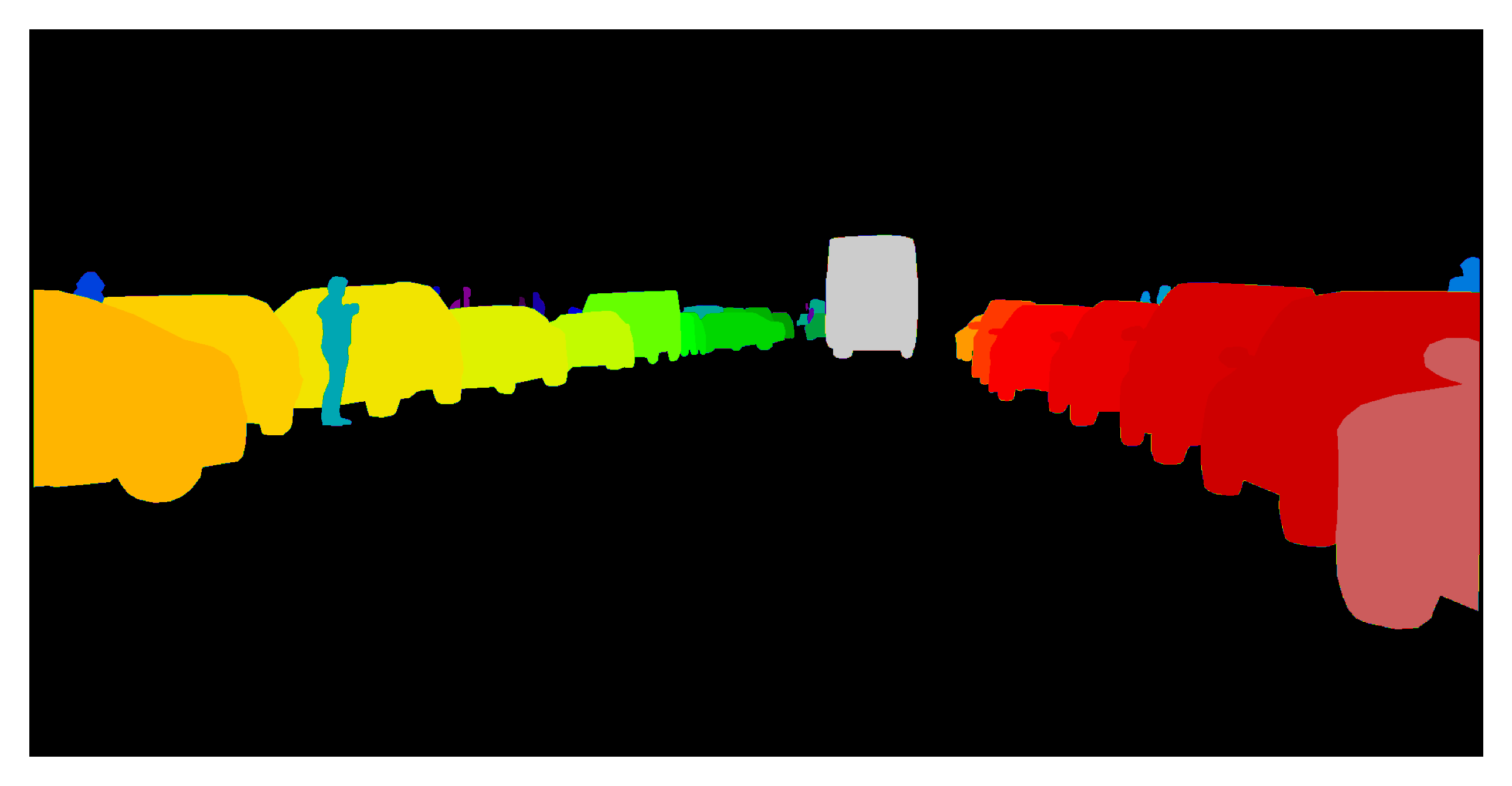}\label{fig:bus_gt}}
\hfill
\subfloat{\includegraphics[trim={70 20 0 10},clip, width=0.33\textwidth]{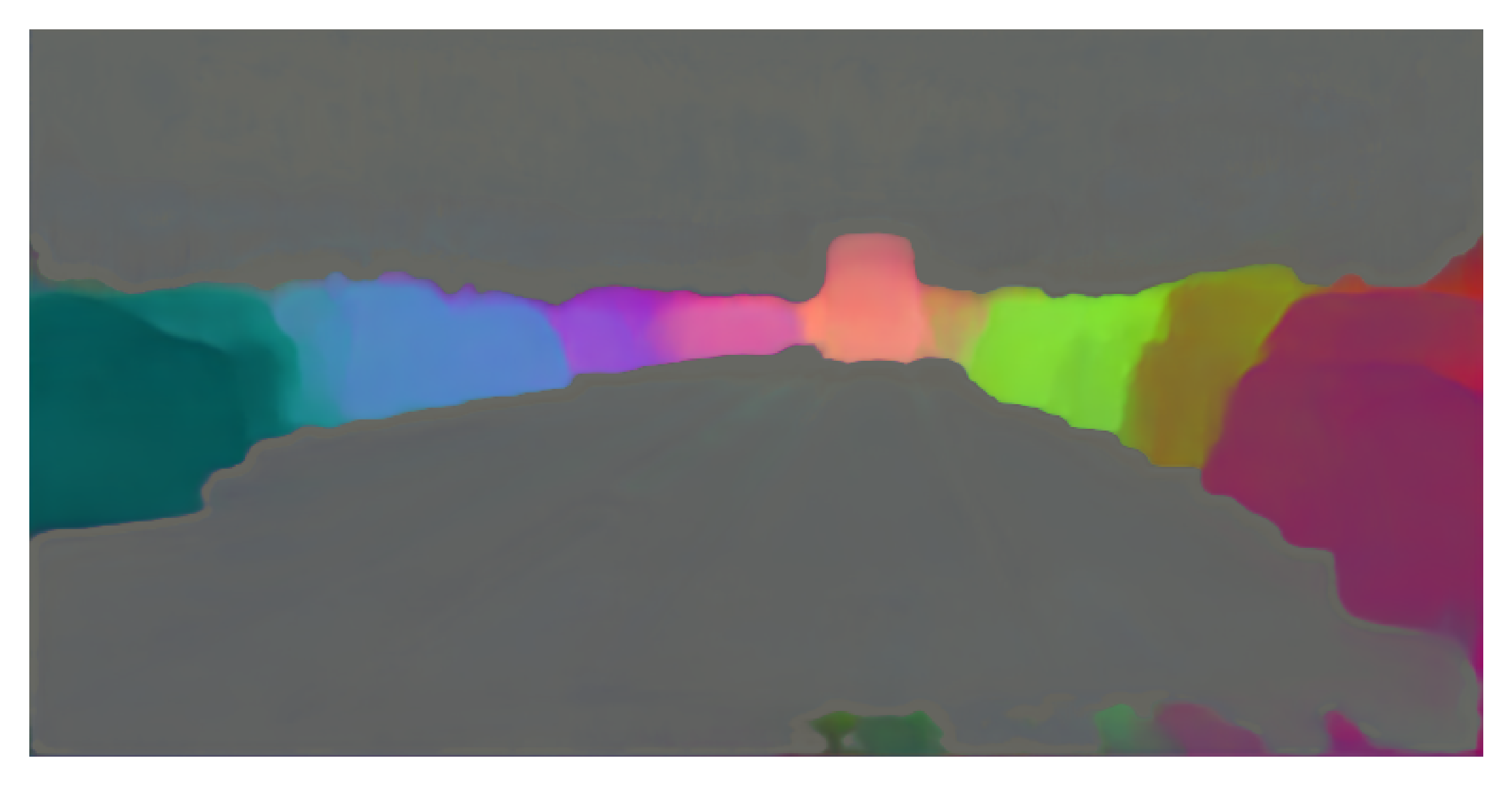}\label{fig:bus_emb}}
\hfill
\subfloat{\includegraphics[trim={70 20 0 10},clip, width=0.33\textwidth]{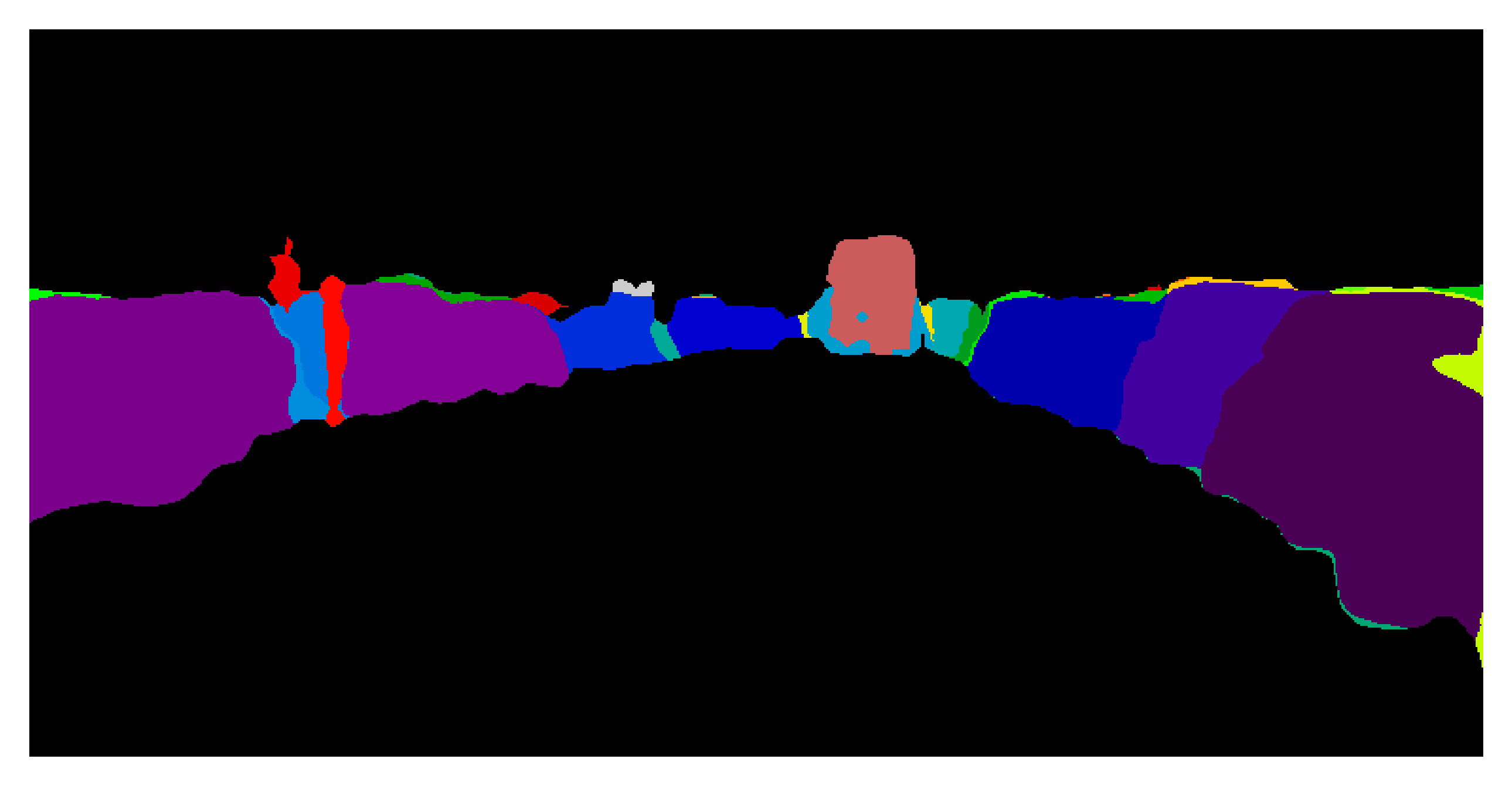}\label{fig:bus_pred}}
\hfill

\subfloat[Ground-truth]{\includegraphics[trim={70 22 0 0},clip, width=0.33\textwidth]{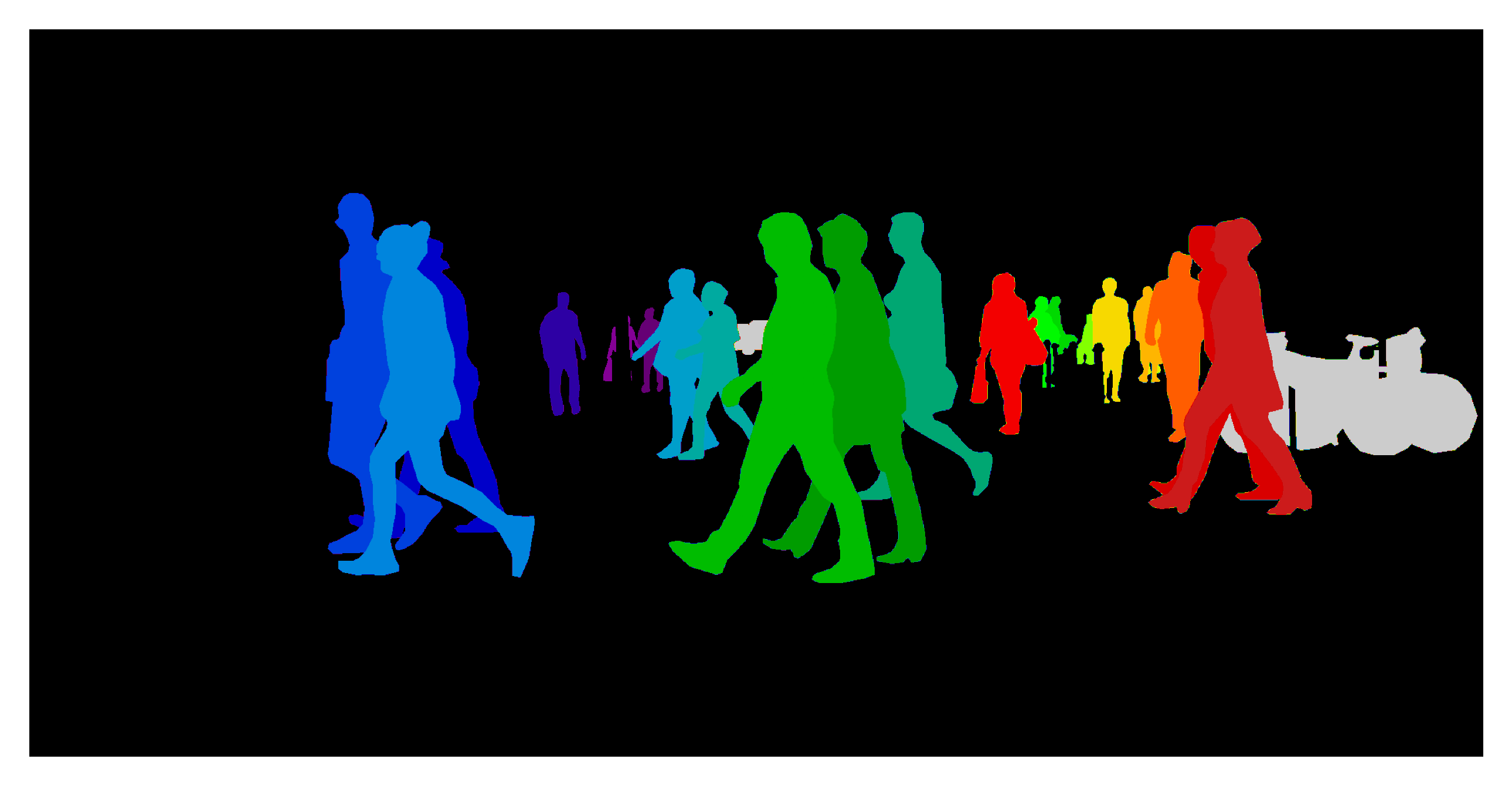}\label{fig:person_gt}}
\hfill
\subfloat[Embeddings]{\includegraphics[trim={70 22 0 0},clip, width=0.33\textwidth]{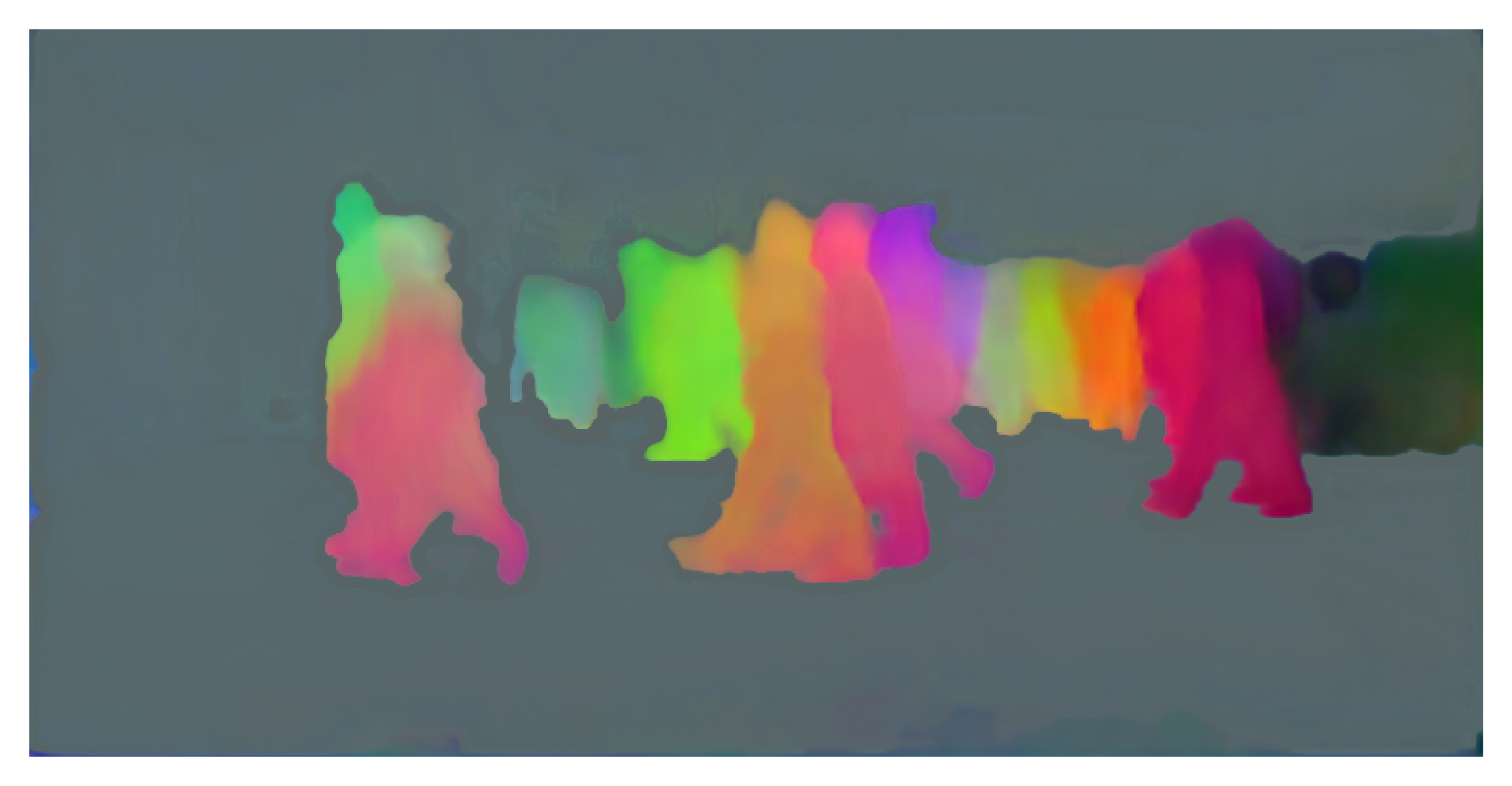}\label{fig:person_emb}}
\hfill
\subfloat[Predictions]{\includegraphics[trim={70 22 0 0},clip, width=0.33\textwidth]{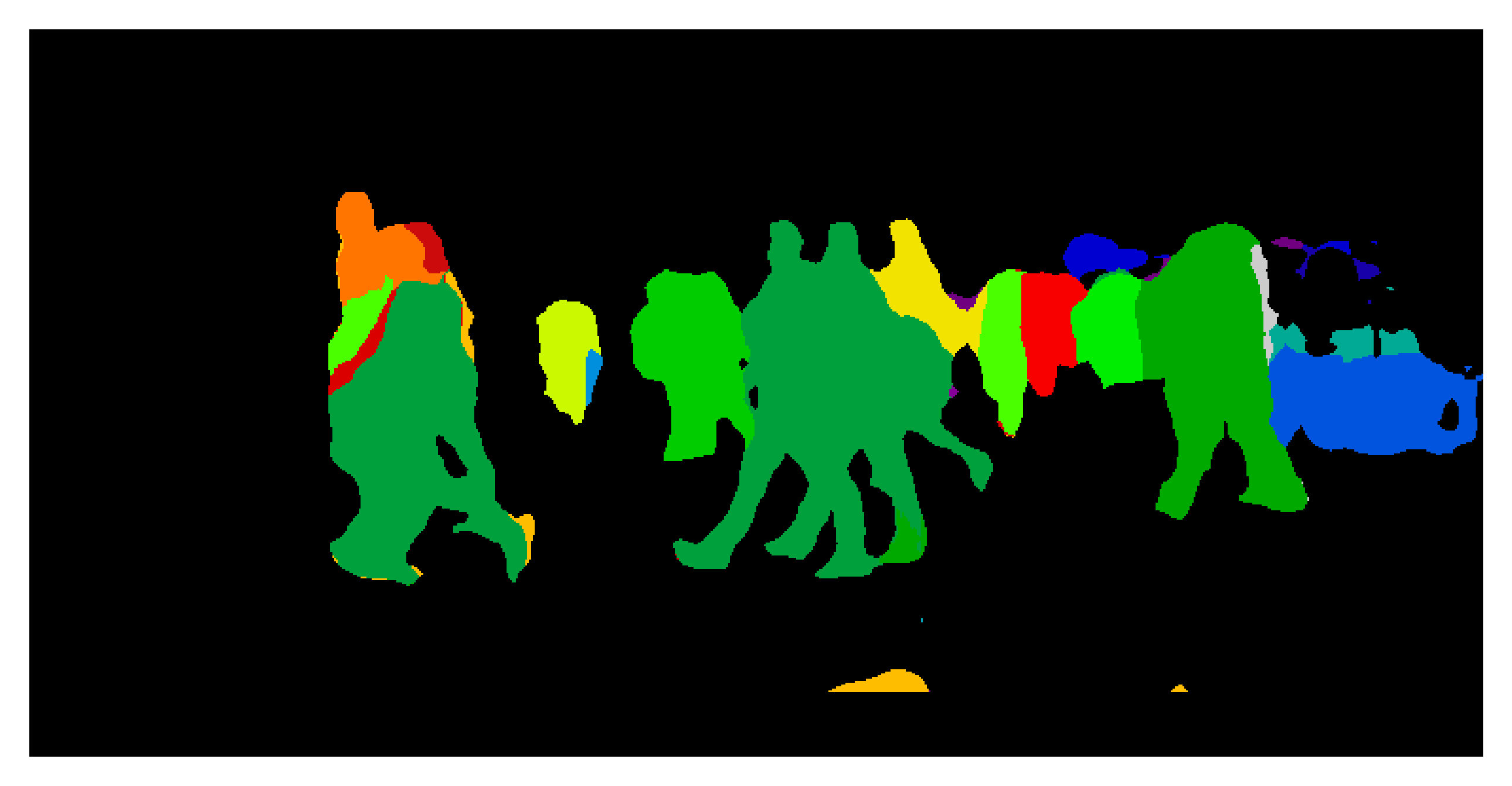}\label{fig:person_pred}}

\caption{Embedding and segmentation masks for Self-train$\mathrm{_{MIX}}$ instance model}
\label{fig:cluster_classes}
\end{figure}

Figure \ref{fig:cluster_classes} shows the PCA projected embeddings and predicted mask outputs of the instance model on different classes. The inference method took the output embeddings from the best model checkpoints for all classes and performed mean-shift clustering and mask merging iteratively for every class to produce the final instance masks. This method ensured that no class is overlapping the other since each anchor point is placed on the respective class' semantic mask to produce the cluster. Other methods are tried where all the embeddings are first averaged or summed across channels along with applying median filters before performing clustering on a single embedding, but it did not improve performance.

\begin{figure}[h]
\captionsetup[subfloat]{labelformat=empty}
\centering
\subfloat[Before]{\includegraphics[trim={10 10 10 10},clip, width=0.245\textwidth]{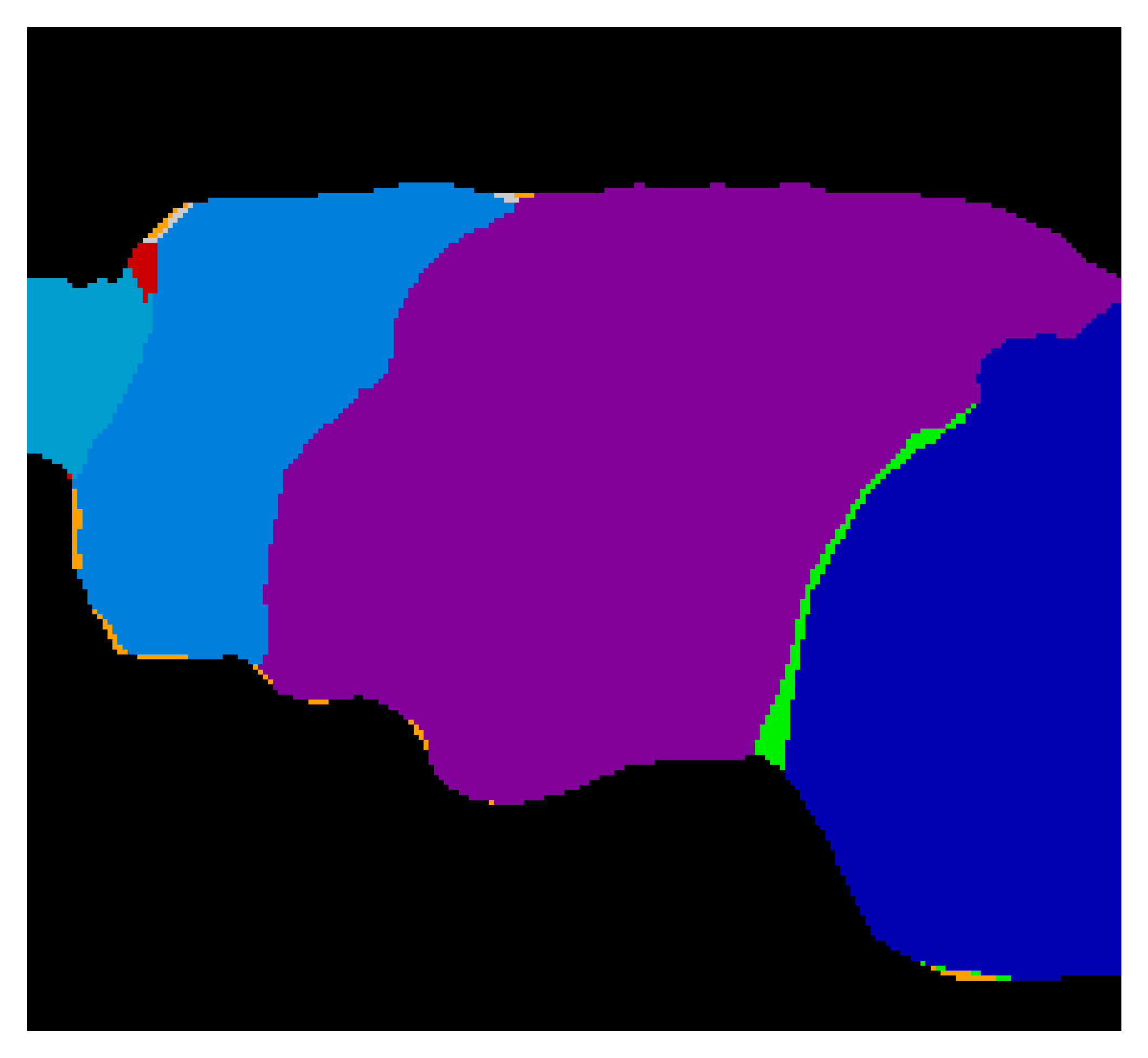}\label{fig:before}}
\hfill
\subfloat[After]{\includegraphics[trim={10 10 10 10},clip, width=0.245\textwidth]{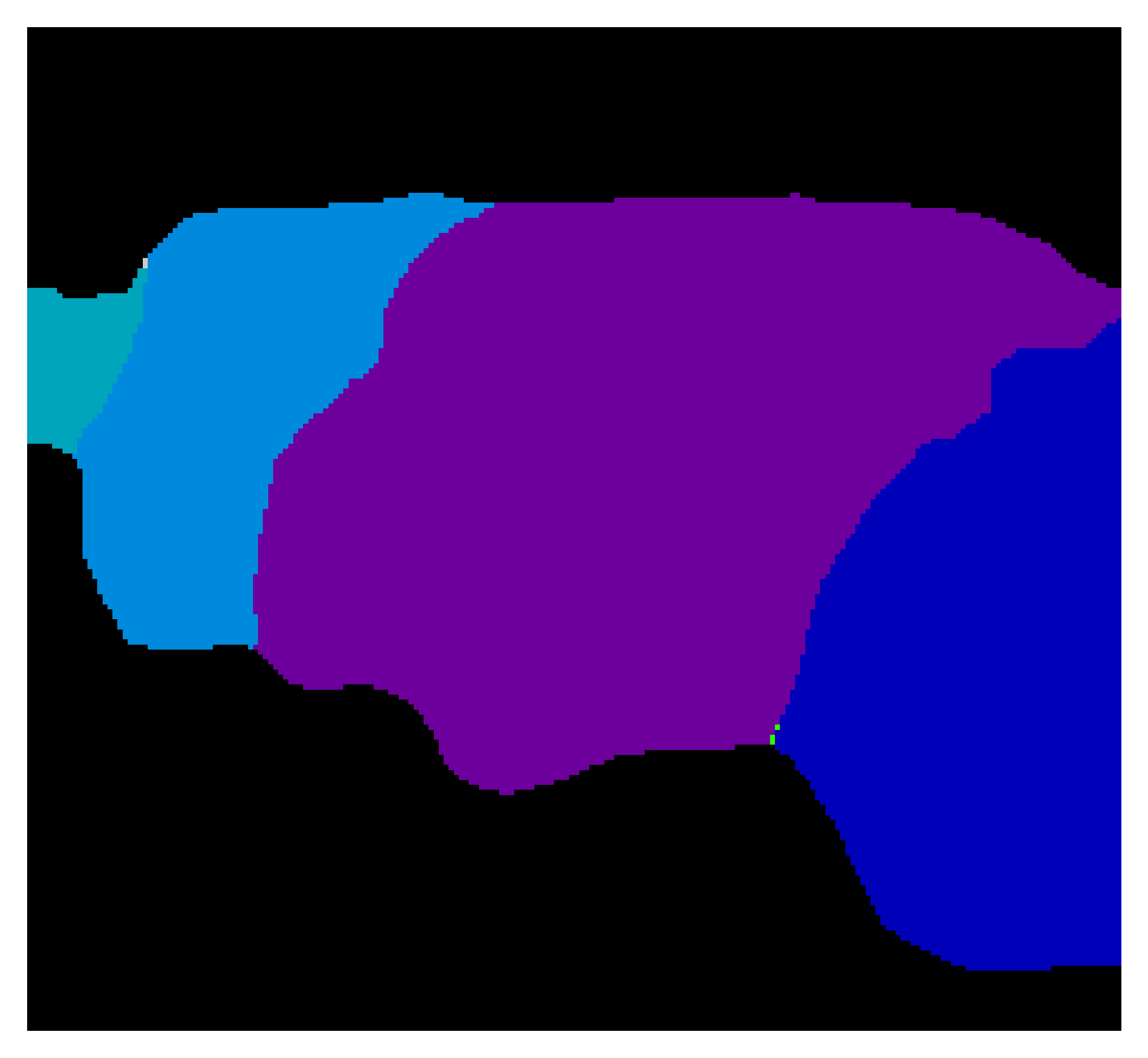}\label{fig:after}}
\hfill
\subfloat[Before]{\includegraphics[trim={52 10 10 10},clip, width=0.245\textwidth]{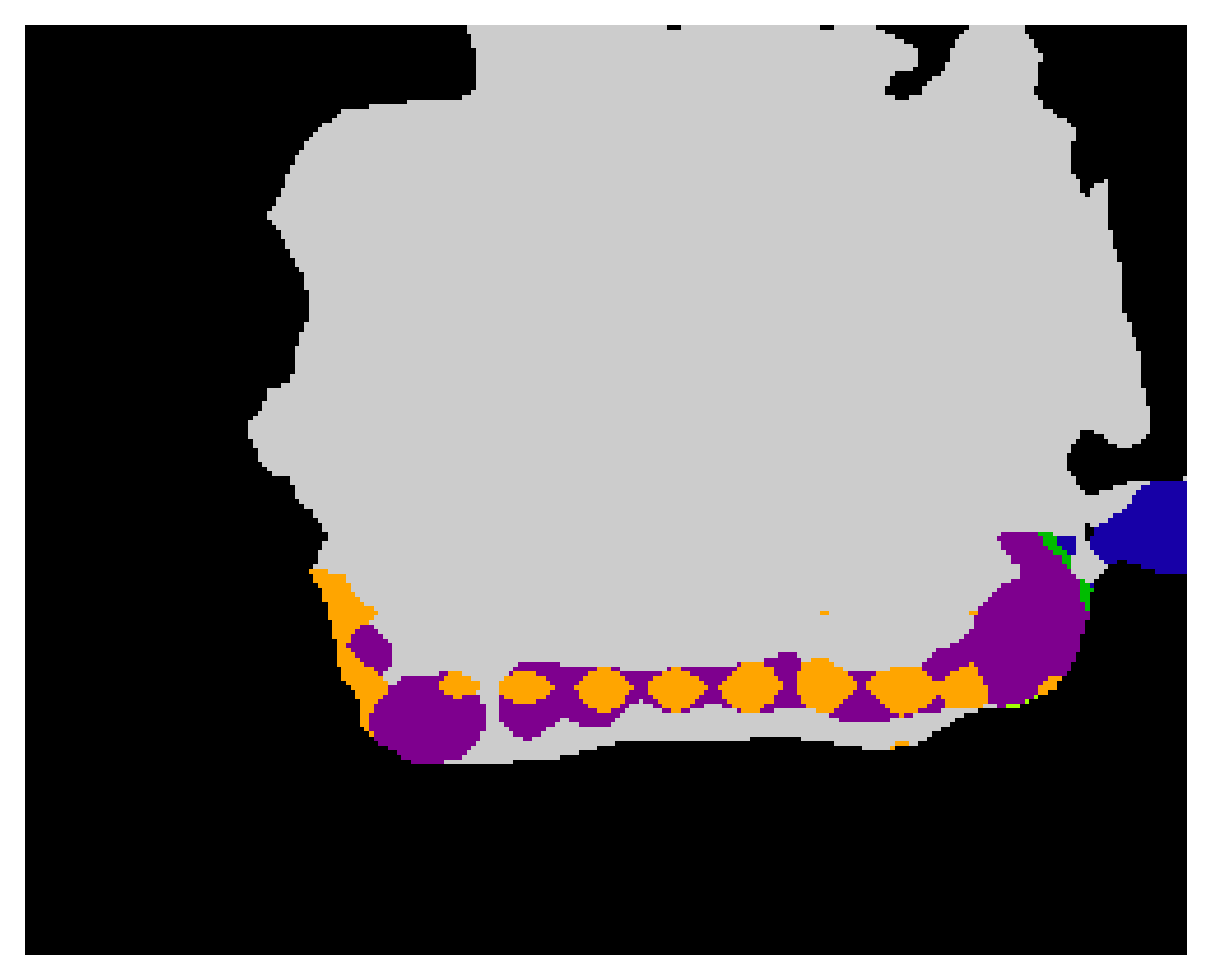}\label{fig:before1}}
\hfill
\subfloat[After]{\includegraphics[trim={52 10 10 10},clip, width=0.245\textwidth]{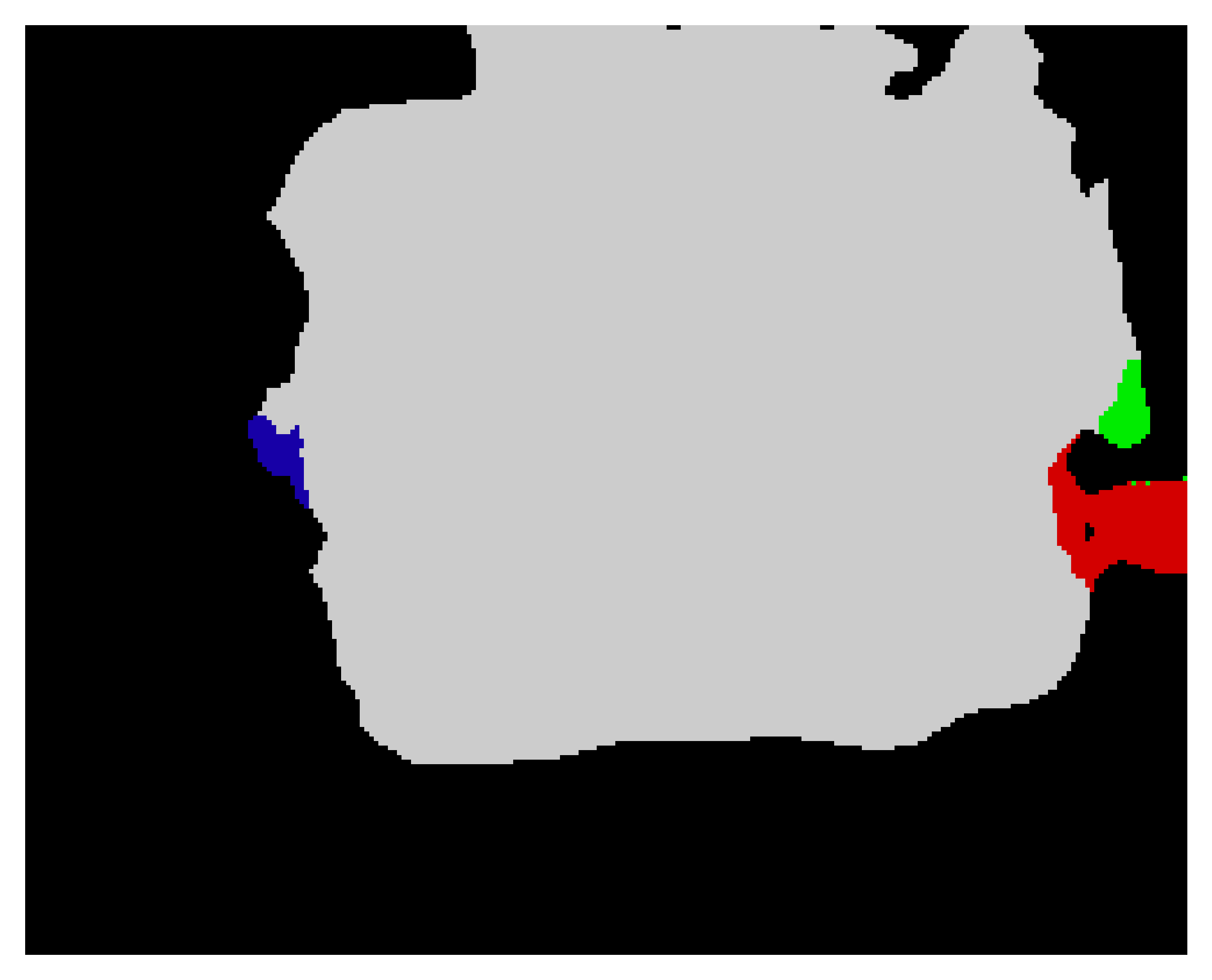}\label{fig:after1}}
\caption{Example of improving instance masks using morphological operation and semantic mask guidance}
\label{fig:post}
\end{figure}

After inspecting the predicted instance masks closely, certain artifacts like noise, object separation, and holes are noticed. These artifacts can be fixed with image post-processing techniques such as erosion and dilation. These methods are morphological operations which can enhance the overall quality and accuracy of images. It involves modifying object boundaries or other artifacts in a binary image, where the operation is performed by applying a structuring element represented as a small matrix or kernel to the image. Erosion eliminates small-scale noise or falsely created regions within the predicted mask, while also reducing the regions between closely adjacent objects. Dilation allows expansion of object boundaries within a binary image using a structuring element, which bridges small gaps within the predicted instance masks ensuring continuity and completeness in the segmented regions. These operations reduce the number of false positive segmentations. Figure \ref{fig:post} illustrates how applying morphological operations reduces erroneous segmentations. Table \ref{tab:inst_miou_final} shows the class-wise mAP score for instance segmentation using different self-train DeepLab networks mentioned in the subscript during inference along with image post-processing improvements.

\begin{table}[H]
\centering
\begingroup
\resizebox{\columnwidth}{!}{%
\begin{tabular}{l|cccccccc|cc}
\toprule
\textbf{Method}  & car  & bus & truck & train & pers & rider & bicy & moto & $\mathrm{mAP^7}$ & $\mathrm{mAP^8}$ \\ \midrule
Self-train$\mathrm{_{V2}}$ & 23.10 & 32.72 & 36.29 & \textbf{11.11} & \textbf{13.16} & 10.21 & \textbf{12.27} & 6.59 & 20.77 & \textbf{18.18} \\ 
Self-train$\mathrm{_{V3}}$ & 23.50 & 50.00 & 46.42 & -     & 9.18  & 7.55  & 9.18  & 7.52 & 21.90 & - \\ 
Self-train$\mathrm{_{V3+}}$& \textbf{23.76} & \textbf{53.44} & \textbf{58.87} & -     & 10.94 & \textbf{12.23} & 6.45 & \textbf{14.93} & \textbf{25.80} & - \\ 
\bottomrule
\end{tabular}}
\endgroup
\caption{mAP (\%) per class on Cityscapes validation for self-train instance models with image post-processing. Using different DeepLab models for predicted semantic mask guided clustering.}
\label{tab:inst_miou_final}
\end{table}

\chapter{Panoptic Segmentation}

The panoptic segmentation task unifies both semantic segmentation, which assigns class labels to each pixel, and instance segmentation, which identifies individual object instances, in order to provide a complete understanding of the visual scene. Panoptic segmentation is evaluated using \textbf{Panoptic Quality} (PQ) \cite{kirillov2019panoptic} which is a comprehensive metric tailored specifically to panoptic segmentation. PQ is the de facto standard for evaluating panoptic segmentation and accounts for both semantic and instance aspects of the task, effectively measuring the quality of class predictions and instance delineations simultaneously. This metric considers intersection-over-union IoU, true positives TP, false positives FP, and false negatives FN, making it informative for assessing the overall performance of a panoptic segmentation model. For each category, PQ is computed as the multiplication of the corresponding \textbf{Segmentation Quality} (SQ) term and and \textbf{Recognition Quality} (RQ) term:

\begin{equation}
\mathrm{PQ}=\underbrace{\frac{\sum_{(p, g) \in T P} \operatorname{IoU}(p, g)}{|T P|}}_{\text {segmentation quality (SQ) }} \times \underbrace{\frac{|T P|}{|T P|+\frac{1}{2}|F P|+\frac{1}{2}|F N|}}_{\text {recognition quality (RQ) }},
\end{equation}

where $g$ is the ground truth segment and $p$ is the matched prediction. One drawback of the PQ metric is its tendency to excessively penalize errors associated with stuff classes, which lack a well-defined instance structure. This issue arises because the metric does not differentiate between stuff and thing classes, consistently applying the criterion of a true positive based on an IoU threshold of 0.5. Therefore, it treats all pixels within an image categorized as stuff as a singular, large instance, which means low pixel level classes like pole and fence would always be under the threshold. To overcome this issue, \cite{porzi2019seamless} suggests a modification to the PQ metric that maintains its application to things classes while introducing a revised metric for stuff classes. In this approach, for a given class $c$, $S_c$ are ground truth segments, and $\hat{S}_c$ are predicted segments. Notably, each image can have at most one ground truth segment and one predicted segment for a given stuff class. For $\mathcal{M}_c=\left\{(p,g) \in \mathcal{S}_c \times \mathcal{\hat{S}}_c\right.$ : $\operatorname{IoU}(p,g)>0\}$ being set of matching segments, $\mathrm{PQ}^{+}$ becomes the final version of the new panoptic metric which averages $\mathrm{PQ}^{\dagger}$ over all classes:

\begin{equation}
\mathrm{PQ}^{+}=\frac{1}{N_{\text {classes }}} \sum_{c \in \mathcal{Y}} \mathrm{PQ}^{\dagger} = \begin{cases}\frac{1}{\left|\mathcal{S}_c\right|} \sum_{(p,g) \in \mathcal{M}_c} \operatorname{IoU}(p,g), & \text { if } c \text { is stuff class } \\ \mathrm{PQ}, & \text { otherwise. }\end{cases}
\end{equation}

\begin{figure}[h]
\centering
\captionsetup[subfloat]{labelformat=empty}

\subfloat[Input]{\includegraphics[trim={10 10 10 10},clip, width=0.33\textwidth]{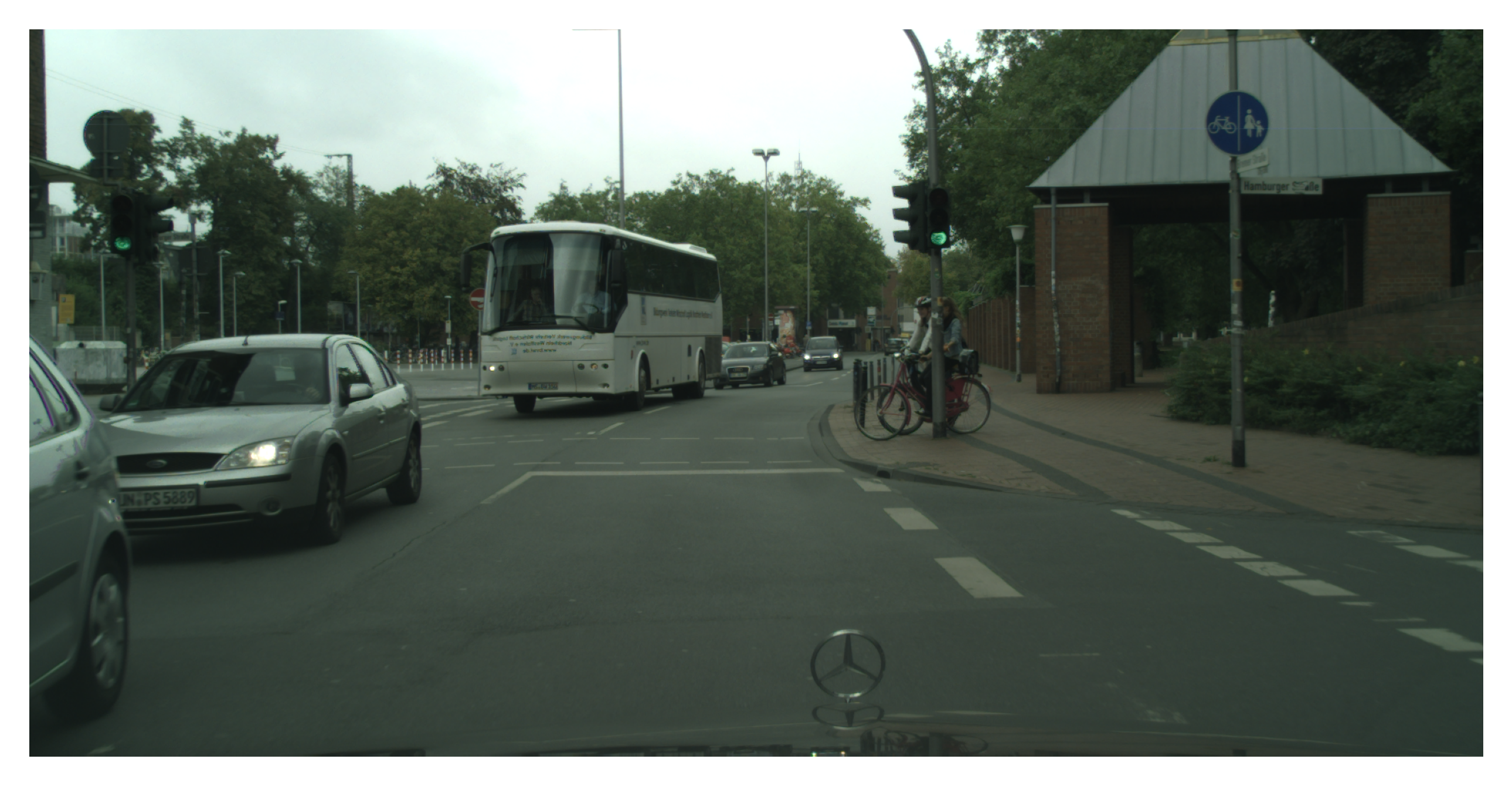}\label{fig:input7}}
\hfill
\subfloat[Ground-truth]{\includegraphics[trim={10 10 10 10},clip, width=0.33\textwidth]{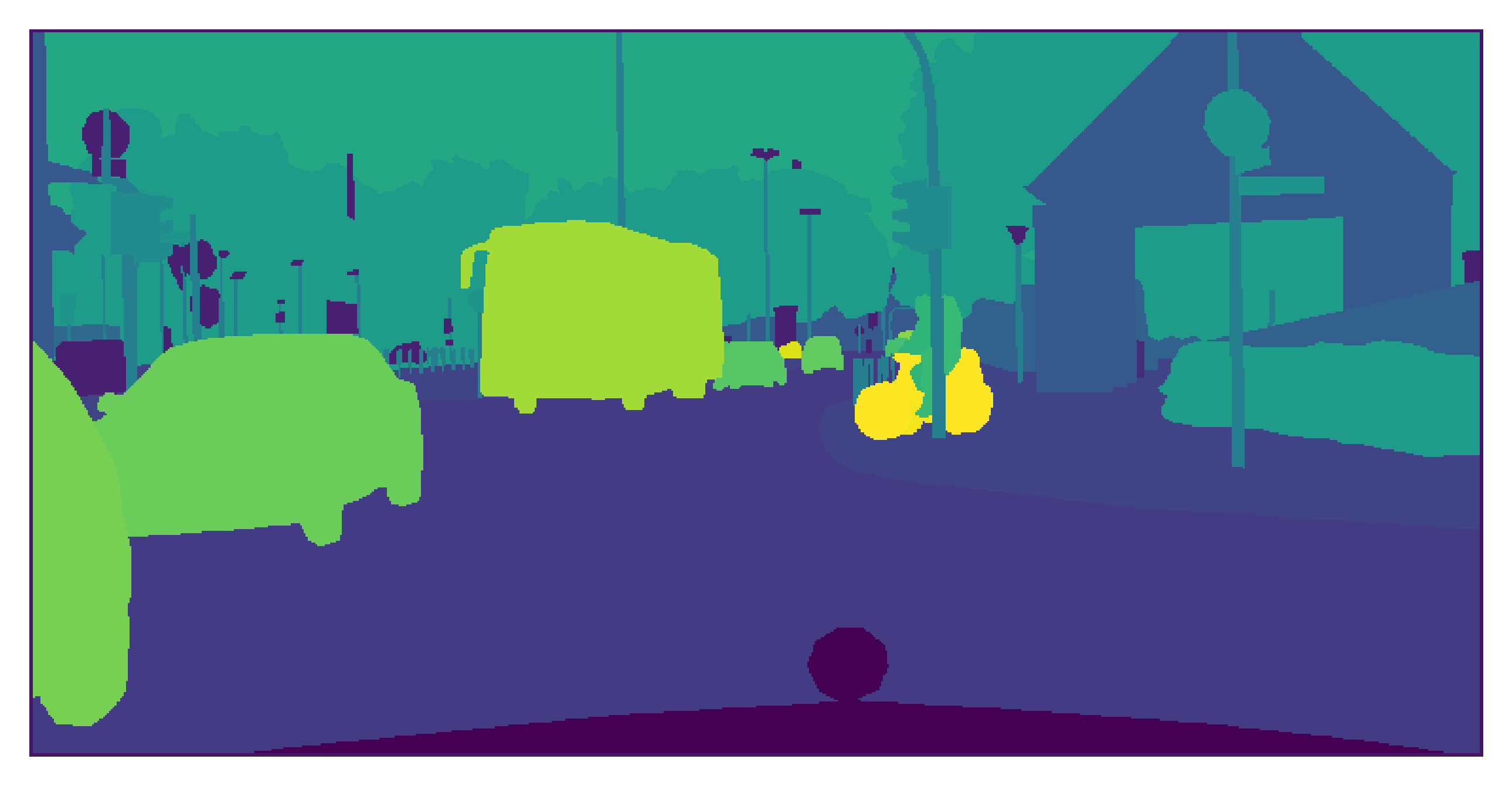}\label{fig:gt7}}
\hfill
\subfloat[Predictions]{\includegraphics[trim={10 10 10 10},clip, width=0.33\textwidth]{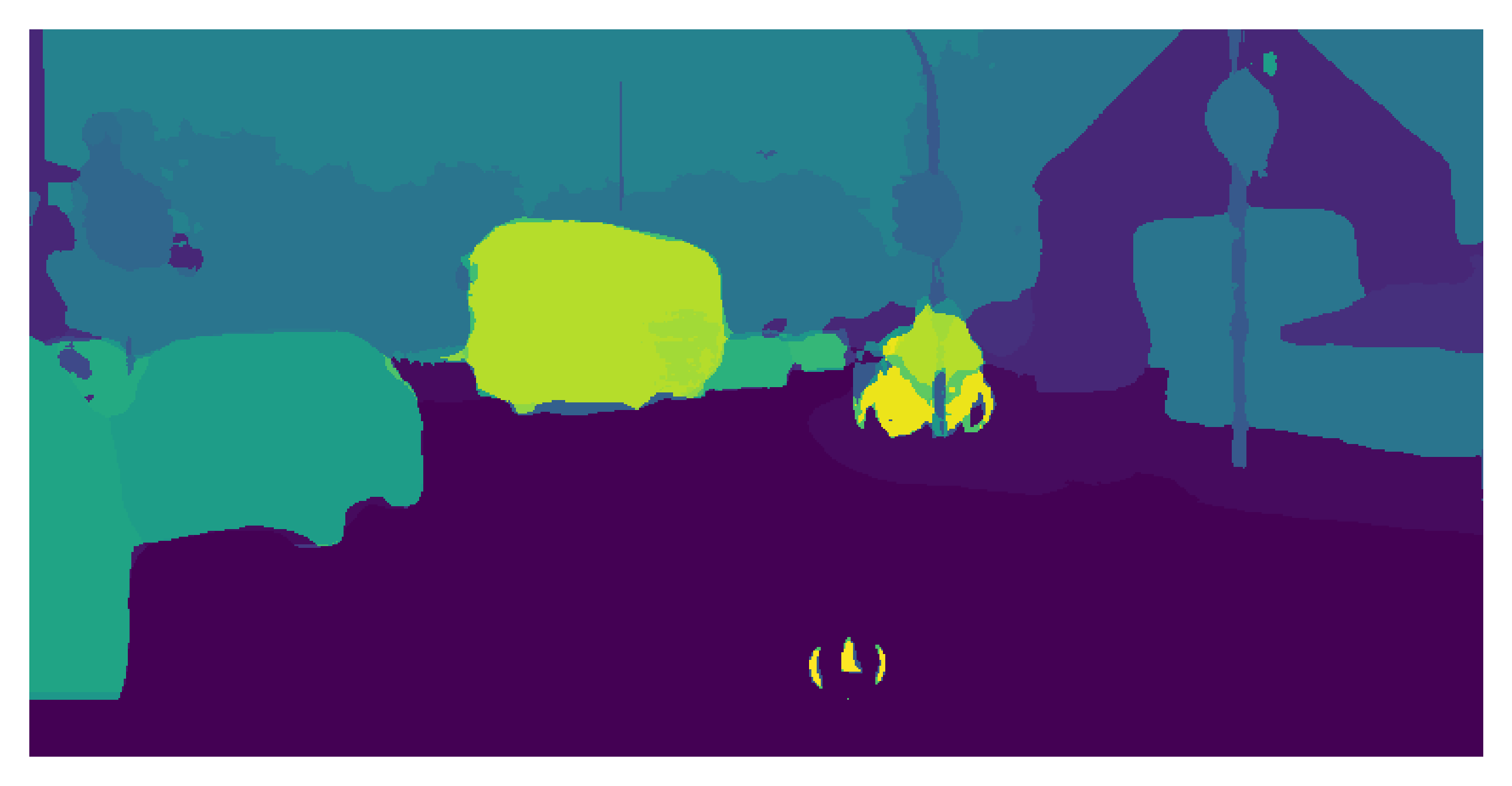}\label{fig:pred7}}
\hfill


\caption{Input, ground-truth and panoptic segmentation predictions}
\label{fig:pan_pics}
\end{figure}

Table \ref{tab:pq} reports the SQ, RQ, PQ and PQ+ scores for different methods from the literature compared to our method. The 'Paper Method' models are trained on 16 Synthia classes, while 'Our Method' models are trained on 19 classes from the mixed dataset explained in section \ref{mix}, and validated on Cityscapes. The panoptic scores for methods PSN \cite{kirillov2019panoptic}, FDA \cite{yang2020fda}, CRST \cite{zou2019confidence}, AdvEnt \cite{vu2019advent}, CVRN \cite{huang2021cross}, UniDA \cite{zhang2023unidaformer}, and EDAPS \cite{saha2023edaps} were obtained from the UniDA and EDAPS paper directly and were not reproduced. 'Our Method' baseline model used both semantic and instance baseline models trained on the mixed source dataset, the self-train models were trained with different DeepLab netoworks mentioned as subscript for both semantic and instance models. It can be noticed that the scores for 'Our Method' are just lower than the state-of-the-art EDAPS method, which can be due to excessive number of false positives present in our instance segmentation model or the presence of bad semantic masks for instance objects. The segmentation quality metric outperforms the EDAPS method but not the recognition quality. However it is a good sanity check to see that the self-train models outperform the baseline model for our method so there seems to be no indication of a bug present in the metric calculation step. Figure \ref{fig:pan_pics} shows the panoptic segmentation prediction results compared to the ground-truth. These panoptic segmentation masks are generated using the self-train DeepLabV3+ networks for both semantic and instance models.

\begin{table}[h]
\centering
\resizebox{\columnwidth}{!}{%
\begin{tabular}{l|cccc|cccc}
\toprule
\textbf{Paper Method} & $\mathrm{SQ^{16}}$ & $\mathrm{RQ^{16}}$ & $\mathrm{PQ^{16}}$ & $\mathrm{PQ^{16}+}$ & $\mathrm{SQ^{19}}$ & $\mathrm{RQ^{19}}$ & $\mathrm{PQ^{19}}$ &$\mathrm{PQ^{19}+}$ \\ \midrule
PSN & 59.0 & 27.8 & 20.1 & - & - & - & - & - \\
FDA & 65.0 & 35.5 & 26.6 & - & - & - & - & - \\
CRST & 60.3 & 35.6 & 27.1 & - & - & - & - & - \\
AdvEnt & 65.6 & 36.3 & 28.1 & - & - & - & - & - \\
CVRN & 66.6 & 40.9 & 32.1 & - & - & - & - & - \\
UniDA & 66.9 & 44.3 & 34.2 & - & - & - & - & - \\
EDAPS & 72.7 & \textbf{53.6} & \textbf{41.2} & - & - & - & - & - \\ \midrule
\textbf{Our Method} & $\mathrm{SQ^{16}}$ & $\mathrm{RQ^{16}}$ & $\mathrm{PQ^{16}}$ & $\mathrm{PQ^{16}+}$ & $\mathrm{SQ^{19}}$ & $\mathrm{RQ^{19}}$ & $\mathrm{PQ^{19}}$ &$\mathrm{PQ^{19}+}$ \\ \midrule
Baseline & 75.03 & 26.35 & 21.13 & 22.76 & 74.93 & 24.76 & 19.83 & 22.28 \\
Self-train$\mathrm{_{V2}}$ & 79.97 & 44.18 & 36.95 & 37.73 & 79.55 & 41.33 & 34.44 & 36.36 \\
Self-train$\mathrm{_{V3}}$ & 78.81 & 43.61 & 36.24 & 38.40 & 78.36 & 40.89 & 33.85 & 36.83 \\
Self-train$\mathrm{_{V3+}}$ & \textbf{80.31} & 45.90 & 38.75 & \textbf{40.57} & \textbf{79.86} & \textbf{43.15} & \textbf{36.27} & \textbf{38.99} \\
\bottomrule
\end{tabular}
}
\caption{Mean SQ, RQ, PQ and PQ+ (\%) on Cityscapes validation set for self-train panoptic segmentation compared to other methods}
\label{tab:pq}
\end{table}

\part{Outlook}

\chapter{Conclusion:}

This project explores developing a framework for self-supervised panoptic segmentation, with the aim to perform segmentation without relying on target data annotations by using techniques like pseudo-labelling and self-training. The semantic model uses predictions from a pre-trained source model to threshold high-confidence pixels and iteratively produce pseudo-labels for the self-trained target model. The instance model utilizes predicted semantic masks to guide embedding clustering to produce pseudo-labels for self-trained instance segmentation. Additionally, morphological post-processing further helps improve mask quality. This project produced the first self-supervised instance segmentation scores for Synthia to Cityscapes domain adaptation task. Several experiments are performed to optimize the training process, like mixing synthetic datasets, comparing clustering algorithms, implementing category-specific models, reducing catastrophic forgetting, and guided improvement of segmentation masks. The self-supervised semantic segmentation scores were improved from the state-of-the-art method \cite{araslanov2021self} by using iterative refinement process, where 30\% of ground-truth instance masks and predicted instance mask from the instance model guided the semantic pseudo-labels to produce more robust segmentations. The main advantage of having this multi-branch approach is to encourage complementary information to flow between semantic and instance models which allows each branch to be optimized for its specific task. It can also be easily fine-tuned as the task and dataset grows with time. This project also implemented the first fully-convolutional embedding-based method for self-supervised panoptic segmentation, compared to the previously implemented region-proposal transformer-based methods. However, the final panoptic quality scores could not overtake the state-of-the-art EDAPS method \cite{saha2023edaps}, and our method sits at second place compared to the current available methods. The proposed pipeline needs further work to implement sophisticated methods to produce more accurate segmentation masks.

\section{Limitation:}

The primary drawback of the proposed method lies in its vulnerability within a critical part of the instance segmentation model. Specifically, when the model is presented with inaccurate semantic segmentation masks, the following clustering process experiences significant disruption, which leads to the generation of erroneous false positive pseudo-labels. This whole process negatively impacts the overall average precision score of the instance model and, consequently diminishes the quality of the panoptic results. This is confirmed by looking at the instance segmentation average precision scores produced using ground-truth segmentation masks, which are much higher than scores produced using predicted semantic masks.

\section{Future Work}

Future work will include implementing the objectness classifier to filter poor segments and incorporate active learning framework to use human feedback for guiding the model to reduce false positive pseudo-labels from emerging in the instance segmentation model. If there was more time, a multi-round training method could be tried where the predicted instance masks could be merged with the semantic masks during self-training to improve the semantic segmentation score. In turn these improved predicted semantic masks could better guide the instance model to improve the instance masks. It would be interesting to see how many rounds of this refinement loop can be implemented before the score reaches saturation point.

\part{Appendix}

\begin{appendix}
\chapter{Lists}
\listoffigures
\listoftables
\bibliography{references}{}
\citestyle{egu}
\bibliographystyle{plainnat}
\end{appendix}
\end{document}